\documentclass[journal]{IEEEtran}

\usepackage[table,xcdraw]{xcolor}
\usepackage{hyperref}
\usepackage{graphicx}
\usepackage{svg}
\usepackage{amsmath,amsfonts,amssymb,amsthm}
\usepackage{fdsymbol}
\usepackage{mathtools}
\usepackage{commath}
\usepackage{enumitem}
\usepackage{booktabs}
\usepackage{pifont}
\usepackage{dblfloatfix}
\usepackage{subcaption}
\usepackage{floatrow}
\usepackage{tikz}
\usetikzlibrary{arrows.meta}
\usetikzlibrary{math}
\usetikzlibrary{positioning}
\usetikzlibrary{calc}
\usetikzlibrary{backgrounds} 
\usetikzlibrary{shapes.geometric, arrows}
\usepackage{xfrac}
\usepackage{multirow}
\usepackage{placeins}
\usepackage[nodayofweek]{datetime}
\newdateformat{mydate}{\THEDAY~\shortmonthname~\THEYEAR}

\makeatletter
\newcommand\notsotiny{\@setfontsize\notsotiny\@vipt\@viipt}
\makeatother

\newcommand{\available}{\star}

\newcommand{\graytext}[1]{\textcolor{gray}{{#1}}}

\newcommand{\initial}[1]{}

\def\etal{\textit{et\penalty50\ al.}}

\newcommand{\subsubsectionspace}[1]{
    \vspace*{2mm}
    \subsubsection{#1}
    \phantom{.}\\[1mm]
}

\newcommand{\TakeAway}{
    \vspace{1mm}
    \textbf{Takeaway:}
}

\renewcommand{\IEEEbibitemsep}{1pt plus 0.5pt}
\makeatletter
\IEEEtriggercmd{\reset@font\normalfont\fontsize{7.15pt}{7.2pt}\selectfont}
\makeatother
\IEEEtriggeratref{1}

\IEEEoverridecommandlockouts
\begin{document}
\title{\huge Few-Shot Object Detection: A Comprehensive Survey}
\author{Mona Köhler, Markus Eisenbach, and Horst-Michael Gross%
\thanks{Authors are with Neuroinformatics and Cognitive Robotics Lab, Ilmenau University of Technology, 98693 Ilmenau, Germany. {\tt\small mona.koehler@tu-ilmenau.de}
This work has received funding from the Carl-Zeiss-Stiftung as part of the project E4SM.}
}%

\markboth{Submitted to IEEE Transactions on Neural Networks and Learning Systems}%
{Köhler \MakeLowercase{\textit{et al.}}: Few-Shot Object Detection: A Survey}

\maketitle

\begin{abstract}
Humans are able to learn to recognize new objects even from a few examples.
In contrast, training deep-learning-based object detectors requires huge amounts of annotated data.
To avoid the need to acquire and annotate these huge amounts of data, few-shot object detection aims to learn from few object instances of new categories in the target domain.
In this survey, we provide an overview of the state of the art in few-shot object detection.
We categorize approaches according to their training scheme and architectural layout.
For each type of approaches, we describe the general realization as well as concepts to improve the performance on novel categories.
Whenever appropriate, we give short takeaways regarding these concepts in order to highlight the best ideas.
Eventually, we introduce commonly used datasets and their evaluation protocols and analyze reported benchmark results.
As a result, we emphasize common challenges in evaluation and identify the most promising current trends in this emerging field of few-shot object detection.
\end{abstract}

\begin{IEEEkeywords}
Object Detection, Few-Shot Learning, Survey, Meta Learning, Transfer Learning
\end{IEEEkeywords}

\IEEEpeerreviewmaketitle

\renewcommand*{\figureautorefname}{Fig.}
\renewcommand*{\tableautorefname}{Tab.}
\renewcommand*{\equationautorefname}{Eq.}
\renewcommand*{\sectionautorefname}{Sec.}
\renewcommand*{\subsectionautorefname}{Sec.}

\newcommand{\Aggregator}{\mathcal{A}}
\newcommand{\concat}{\mathrm{cat}}

\newcommand{\featBackbone}{\mathrm{f}^{\mathcal{B}}}
\newcommand{\featRPN}{\mathrm{f}^{RPN}}
\newcommand{\RoI}{\mathcal{R}}
\newcommand{\featRoI}{\mathrm{f}^{\RoI}}
\newcommand{\featA}{\mathrm{f}_{\mathrm{agg}}}
\newcommand{\featQ}{\mathrm{f}^\mathcal{Q}}
\newcommand{\featS}{\mathrm{f}^\mathcal{S}}
\newcommand{\featSc}{\mathrm{f}^{\mathcal{S},c}}
\newcommand{\featScnum[1]}{\mathrm{f}^{\mathcal{S},c_#1}}
\newcommand{\ImQ}{I^\mathcal{Q}}
\newcommand{\ImS}{I^\mathcal{S}}
\newcommand{\ImSc}{I^{\mathcal{S},c}}
\newcommand{\BranchQ}{\mathcal{Q}}
\newcommand{\BranchS}{\mathcal{S}}
\newcommand{\RoIc}{\mathcal{R}^c}
\newcommand{\RoIAgg}{\mathcal{R}^\mathcal{A}}
\newcommand{\RoIAggc}{\mathcal{R}^{\mathcal{A}, c}}
\newcommand{\Cbase}{\mathcal{C}_{base}}
\newcommand{\Cnovel}{\mathcal{C}_{novel}}
\newcommand{\Dbase}{\mathcal{D}_{base}}
\newcommand{\Dnovel}{\mathcal{D}_{novel}}
\newcommand{\ModelInit}{\mathcal{M}_{init}}
\newcommand{\ModelBase}{\mathcal{M}_{base}}
\newcommand{\ModelFinal}{\mathcal{M}_{final}}
\newcommand{\Backbone}{\mathcal{B}}
\newcommand{\featBackboneQ}{\mathrm{f}^{\mathcal{Q},\mathcal{B}}}
\newcommand{\featBackboneS}{\mathrm{f}^{\mathcal{S},\mathcal{B}}}

\section{Introduction}
In the last decade, object detection has tremendously improved through deep learning
\cite{ObjectDetectionReview-TNNLS2019, ObjectDetectionSurvey-IJCV2020}.
However, deep-learning-based approaches typically require vast amounts of training data.
Therefore, it is difficult to apply them to real-world scenarios involving novel objects that are not present in common object detection datasets.
Annotating large amounts of images for object detection
is costly and tiresome.
In some cases -- such as medical applications~\cite{medical-katzmann-neurocomputing2021} or the detection of rare species~\cite{DetectRareSpecies-ConsBio2021} -- it is even impossible to acquire plenty of images.
Moreover, in contrast to typical deep-learning-based approaches, humans are able to learn new concepts with little data even at early age~\cite{ObjectNameLearningHumans-Psycho-2002, HumanObjectNaming-Development-2005, HumansWordMeaning-phd-2009}.
When children are shown new objects, they are able to recognize these objects even if they have seen them only once to a few times.

\begin{figure}[t]
    \centering
    \includegraphics[width=\linewidth]{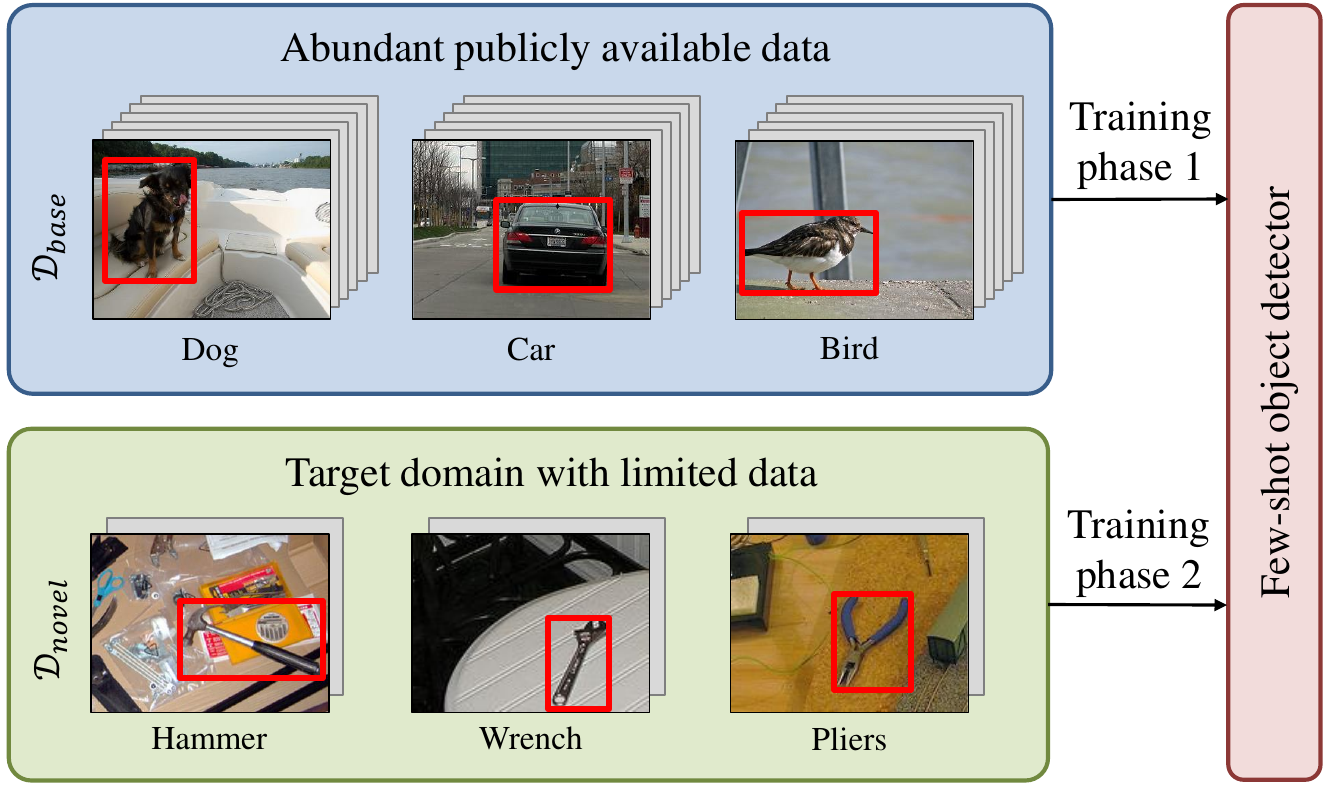}
    \caption{
    General idea: By first training on a base dataset with abundant annotated bounding boxes, it is possible to apply few-shot object detectors to settings with only few annotated object instances, such as mechanical tools.
    }
    \label{fig:eyecatcher}
\end{figure}

Therefore, a promising research area in this direction is few-shot object detection (FSOD).
FSOD aims at detecting novel objects with only few annotated instances after pretraining in the first phase on abundant publicly available data, as shown in \autoref{fig:eyecatcher}.
Consequently, it alleviates the burden of annotating large amounts of data in the target domain.

In this survey, we aim to provide an overview of state-of-the-art FSOD approaches for new researchers in this emerging research field.
First, we define the problem of FSOD.
Afterwards, we categorize current approaches and highlight similarities as well as differences.
Subsequently, we introduce commonly used datasets and provide benchmark results.
Finally, we emphasize common challenges in evaluation and identify promising research directions to guide future research.

\section{Problem Definition}

Few-shot object detection aims at detecting novel objects with only few annotated instances.
Formally, the training dataset $\mathcal{D} = \Dbase \cup \Dnovel$ is separated into two datasets $\Dbase$ and $\Dnovel$ containing non-overlapping sets of base categories $\Cbase$ and novel categories $\Cnovel$, with \mbox{$\Cbase \cap \Cnovel = \varnothing$}.
Each tuple \mbox{$(I_i, \hat{y}_{o_1}, \ldots, \hat{y}_{o_M}) \in \mathcal{D}$} consists of an image \mbox{$I_i = \{ o_1, \ldots, o_M\}$} containing $M$ objects \mbox{$o_1, \ldots, o_M$} and their corresponding labels \mbox{$\hat{y}_{o_i} = \{c_{o_i}, b_{o_i}\}$}, including the category $c_{o_i}$ and the bounding box \mbox{$b_{o_i} = \{x_{o_i}, y_{o_i}, w_{o_i}, h_{o_i}\}$} with coordinates $(x_{o_i}, y_{o_i})$, width $w_{o_i}$, and height $h_{o_i}$.
For the base categories $\Cbase$ abundant training data are available in the base dataset $\Dbase$. 
In contrast, the novel dataset $\Dnovel$ contains only few annotated object instances for each novel category in $\Cnovel$.
For the task of K-shot object detection, there are exactly $K$ annotated object instances available for each category in $\Cnovel$.
Therefore the number of annotated novel object instances $|\{o_j \in I_i \forall I_i \in \Dnovel\}| = K \cdot |\Cnovel|$ is relatively small.
Note that the number of annotated object instances does not necessarily correspond to the number of images, as one image may contain multiple instances.
The most difficult case for FSOD is one-shot object detection, where $K=1$.
N-way object detection denotes a detector that is designed to detect object instances from $N$ novel categories, where $N \leq | \Cnovel|$.
Few-shot object detection is therefore often referred to as \mbox{N-way K-shot} detection.

Training an object detector only on $\Dnovel$ quickly leads to overfitting and poor generalization due to limited training data~\cite{LSTD-aaai2018, MetaRCNN-iccv2019}.
However, training on the highly imbalanced combined data $\mathcal{D} = \Dnovel \cup \Dbase$ generally results in a detector that is heavily biased towards the base categories and, therefore, unable to correctly detect instances from novel categories~\cite{MetaRCNN-iccv2019}.
Therefore, current research focuses on novel approaches for few-shot object detection.
Typically, the initial detector model $\ModelInit$ equipped with a backbone pretrained on classification data is first trained on $\Dbase$, resulting in the base model $\ModelBase$.
Most approaches then train $\ModelBase$ on data $\mathcal{D}_{finetune} \subseteq \mathcal{D}$ including novel categories $\Cnovel$, resulting in the final model $\ModelFinal$:
\begin{equation}
\footnotesize
    \ModelInit \xrightarrow{\Dbase} \ModelBase \xrightarrow{\mathcal{D}_{finetune}} \ModelFinal
\end{equation}
\section{Related Work on Training with Limited Data}
There are some related research areas that also focus on training with limited data.
In the following, we will briefly discuss differences and similarities with FSOD.
\vspace{-3mm}
\subsection{Related Concepts for Learning with Limited Data}
\hspace*{-3.5mm}\textbf{Few-Shot Learning and Classification}:
Before being applied to detection, few-shot learning was first explored for classification tasks~\cite{MatchingNets-NeurIPS2016, PrototypicalNetworks-NeurIPS2017, Survey-Few-Shot-Learning-CSUR2020, Meta-Learning-Survey-TPAMI2021}.
As objects with only few training instances do not need to be localized, classification is clearly easier.
Yet, many ideas can be adopted for few-shot object detection.

\vspace*{1.5mm}%
\hspace*{-3.5mm}\textbf{Semi-Supervised Learning}
is related to few-shot learning in that only few labeled instances of the target categories are available.
However, in contrast to few-shot learning, large amounts of additional unlabeled data are often available that help to learn appropriate representations \cite{Semi-Supervised-Book-2010, semi-Supervised-Survey-ML2020, Semi-Supervised-Detection-WACV2021}.

Thus, when additional unlabeled data are available, methods from semi-supervised learning should be considered to improve the learned representations in few-shot learning approaches.

\vspace*{1.5mm}%
\hspace*{-3.5mm}\textbf{Incremental Learning}:
Typical deep-learning approaches suffer from catastrophic forgetting, when the model is trained on new data.
In contrast, incremental learning approaches \cite{incremental-algorithms-applications-esann2016, incremental-end-to-end-eccv2018, incremental-large-scale-cvpr2019} aim to retain the performance on old categories, when new categories are added incrementally.
Some FSOD approaches also incorporate incremental learning techniques.

\vspace{-5mm}
\subsection{Object Detection}
\hspace*{-3.5mm}\textbf{Generic Object Detection}
is the joint task of localizing and classifying object instances of categories the detector was trained on.
Regions of interest are localized by coordinates of bounding boxes and classified into a predefined set of categories.
All other object categories which are not part of the training categories are regarded as background, and the detector is trained to suppress detections of those other categories.
While achieving impressive results, these approaches require loads of annotated object instances per category and typically fail when applied to the few-shot regime.
For researchers new in this field, we refer to a brief summary in the appendix in \autoref{sec:appendix-generic-object-detection} or to comprehensive surveys~\cite{ObjectDetectionReview-TNNLS2019, ObjectDetectionSurvey-IJCV2020} on this topic.

\vspace*{1.5mm}%
\hspace*{-3.5mm}\textbf{Cross-Domain Object Detection}
\cite{cai2019exploring, xu2020cross, chen2020harmonizing} is the task of first training a detector on abundant labeled data and then adapting this detector to a different domain with limited data; a typical example is synthetic-to-real.
However, unlike for FSOD, the categories stay the same accross different domains.

\vspace*{1.5mm}%
\hspace*{-3.5mm}\textbf{Zero-Shot Object Detection}
can be defined similar to few-shot object detection.
However, as an extreme case, the number of annotated object instances is zero ($K{=}0$).
Zero-shot detectors often incorporate semantic word embeddings~\cite{zero-shot-object-detection-eccv2018, ZSOD-accv2018}, i.e., semantically similar categories lead to similar features in the embedding space.
This works for detecting everyday objects which can be easily labeled, but might be problematic when providing a specific label is difficult or when very similar objects need to be distinguished.

\vspace*{1.5mm}%
\hspace*{-3.5mm}\textbf{Weakly Supervised Object Detection}
relaxes the required annotations such that the training data contain only image-level labels, i.e., whether a specific object category is present or absent somewhere in the image \cite{weaklySupervisedSurvey-tpami2021, weaklySupervisedSurvey-arxiv2021}.
These annotations are much easier to obtain and can often be acquired by keyword search.
The challenge for weakly supervised object detectors is detecting all object instances without having any localization information during training. 
Although alleviating the annotation burden, weakly supervised object detectors still require large amounts of images, which might be hard to obtain for detecting rare objects.

\subsection{Learning Techniques for Few-Shot Object Detection}
In addition to the related research areas described above, in the following we will address learning techniques that are widely adopted in few-shot object detection.

\vspace*{1.5mm}%
\hspace*{-3.5mm}\textbf{Transfer Learning}
refers to the re-use of network weights pretrained on a baseline dataset to improve generalization capabilities on a new domain with limited data.
As in few-shot learning and detection, this usually involves novel categories from the target domain.
However, unlike few-shot learning, the number of object instances for novel categories is not necessarily small.
Therefore, techniques for learning from few data need to be incorporated in transfer learning approaches for few-shot object detection.

\vspace*{1.5mm}%
\hspace*{-3.5mm}\textbf{Metric Learning}
aims for learning an embedding in which inputs with similar content are encoded in features that have a small distance to each other in terms of the metric while encoded features from dissimilar inputs are supposed to be far apart~\cite{Deep-Metric-Learning-Survey-2019}.
To learn features with low inner-class and high inter-class $\ell^2$ distances, triplet loss~\cite{TripletLoss-jmlr2009} or its extensions (see overview in~\cite{Aganian-ReID-icann2021}) are often used.
Since this learned feature embedding typically generalizes well, the model can also be applied to encode instances of novel categories, which were unknown during training, and make metric-based decisions without the need for re-training.
In the context of few-shot classification, this means that during inference the model extracts feature embeddings of the few annotated examples of $\Dnovel$ as well as of corresponding test images.
The test image is then assigned to the category of the closest feature embedding of an annotated example.
However, for few-shot detection, concepts for localizing instances in the images need to be integrated.

\vspace*{1.5mm}%
\hspace*{-3.5mm}\textbf{Meta Learning}
approaches learn how to learn in order to generalize for new tasks or new data~\cite{Meta-Learning-Survey-TPAMI2021}.
For few-shot learning that means that these approaches learn how to learn to categorize the given inputs even though the categories are not fixed during training.
These approaches need to learn how the required knowledge about the category is learned most efficiently, so that this category knowledge can also be learned for novel categories with few training examples.

\subsection{Related Surveys}

Although other surveys on FSOD are available \cite{FSOD-Survey-acm2022, FSOD-comparative-review-arxiv2021, FSOD-empirical-study-arxiv2022, Low-Shot-Detection-survey-arxiv2021, FSOD-self-supervised-survey-arxiv2021, FSOD-survey-chinese}, they do not cover as many publications related to FSOD as we do (see appendix, \autoref{tab:surveys-comparison}).
\cite{FSOD-comparative-review-arxiv2021, Low-Shot-Detection-survey-arxiv2021,  FSOD-self-supervised-survey-arxiv2021} are broader surveys, also addressing self-supervised, weakly-supervised, and/or zero-shot learning and do not focus as much on FSOD.
\cite{FSOD-Survey-acm2022, FSOD-comparative-review-arxiv2021} only cover earlier work on FSOD and hence are somewhat outdated since at least some of the currently best performing approaches on common benchmarks are missing.
As~\cite{FSOD-survey-chinese} is not available in English language, it is only accessible for a limited group of researches.
Overall, our survey is most related to~\cite{FSOD-empirical-study-arxiv2022}, as it also elaborates several core concepts and groups approaches according to this concepts.
However, with the visual taxonomy in \autoref{fig:dual-branch-meta-categories}, \autoref{fig:single-branch-meta-categorization}, and \autoref{fig:categroies-transfer-learning} we enable the reader to faster grasp which approaches follow similar concepts and what concepts seem to complement each other well.
We also provide a better guidance on benchmark results by highlighting differences in evaluation protocols and grouping approaches with comparable evaluations.
Furthermore, we provide a much more comprehensive survey by covering nearly twice as many FSOD papers as \cite{FSOD-empirical-study-arxiv2022} did.

\newcommand{\dualbranch}{\ding{162}}
\newcommand{\singlebranch}{$\medblackcircle$}
\newcommand{\transfer}{$\rhd$}

\vspace{3mm}
\section{Categorization of Few-shot object detection approaches}
Approaches for few-shot object detection incorporate novel ideas in order to be able to detect objects with only few training examples.
In general, the abundant training examples for base categories $\Cbase$ are used to leverage knowledge for the novel categories $\Cnovel$ with limited labeled data.

\begin{figure}[!ht]
    \centering
    \resizebox{0.7\textwidth}{!}{%
    \begin{tikzpicture}
    \tikzstyle{rect} = [rectangle, text centered, font=\small, minimum height=0.6cm, draw=black]
    \tikzstyle{arrow} = [thick,->,>=stealth]
    \node (FSOD) [rect] {\textbf{Few-Shot Object Detection}};
    \node (transfer) [rect, below of=FSOD, xshift=1.5cm] {\transfer~Transfer Learning};
    \node (meta) [rect, below of=FSOD, xshift=-1.5cm] {Meta Learning};
    \node (dual-branch) [rect, below of=meta, xshift=-1.3cm] {\dualbranch~Dual-Branch};
    \node (single-branch) [rect, below of=meta, xshift=1.3cm] {\singlebranch~Single-Branch};
    \draw [arrow] (FSOD) -- (meta);
    \draw [arrow] (FSOD) -- (transfer);
    \draw [arrow] (meta) -- (dual-branch);
    \draw [arrow] (meta) -- (single-branch);
    \end{tikzpicture}
    }
    \vspace{-1mm}
    \caption{Categorization of few-shot object detection approaches}
    \label{fig:categorization}
\end{figure}
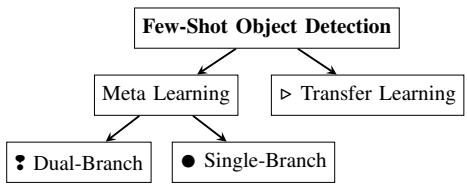

We categorize approaches for FSOD into meta learning and transfer learning approaches as shown in~\autoref{fig:categorization}.
We further divide meta learning approaches into single-branch and dual-branch architectures.
Dual-branch architectures are constituted by a query and a support network branch, i.e., the network processes two inputs (a query and a support image) separately.
Single-branch approaches in general resemble the architecture of generic detectors but reduce the number of learnable parameters when training on novel categories or utilize metric learning.
Yet, also several dual-branch architectures and some transfer learning approaches incorporate ideas from metric learning.
Therefore, to avoid ambiguous categorization, we do not use metric learning as a separate category, as done in early work on FSOD.
Instead, we distinguish by training schemes and architectural aspects, which better reflects the different trends in the current state of the art.

In the appendix (\autoref{fig:timeline}), we show a diagram, that lists all approaches, that we cover in this survey sorted by year of publication and conference or journal, respectively.
We can see that few-shot object detection is a rather young but emerging research field as most approaches have been published only within the last three years.
Most approaches use transfer learning or dual-branch meta learning.

In the following, we first describe dual-branch meta learning approaches in \autoref{sec:meta-learning-dual-branch}.
We start with the general training scheme for meta learning and follow with the typical realization.
In the following subsections, we describe how specific approaches deviate from the general realization.
In \autoref{sec:meta-learning-single-branch} we focus on single-branch meta learning approaches.
Although there is no common realization from which others deviate, we still group approaches to their main ideas.
In \autoref{sec:transfer-learning} we cover transfer learning approaches.
Similarly to dual-branch meta learning approaches, we first describe the general realization and then turn to modifications.

Whenever appropriate, we give short takeaways at the end of the subsections in order to highlight key insights.
Some takeaways also contain citations to link the concept to specific well-performing methods regarding to benchmarks in \autoref{sec:experiments}.

Moreover, we summarize the best performing approaches for each training scheme at the end of the corresponding section.
Finally, in \autoref{sec:comparison} we draw a comparison between meta learning and transfer learning approaches before discussing common datasets and benchmark results in \autoref{sec:experiments}.

\begin{figure*}[t]
    \centering
    \includegraphics[width=\linewidth]{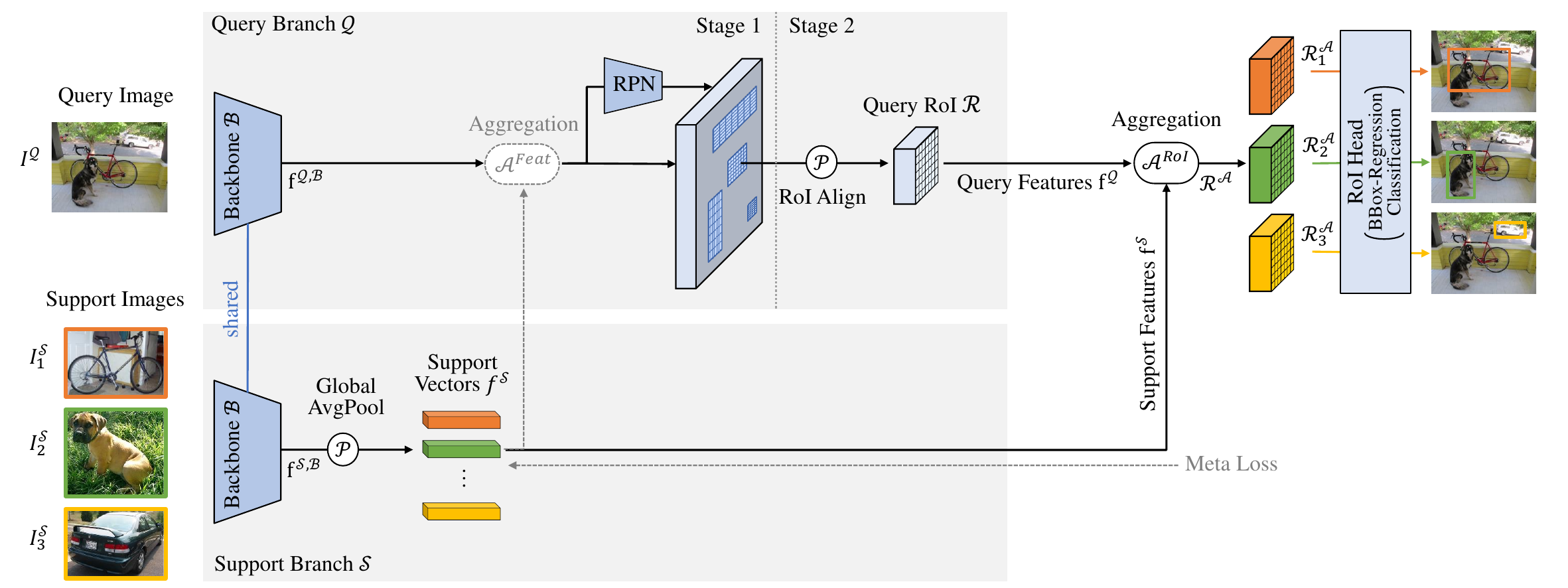}
    \caption{General architecture for dual-branch meta learning based on Faster R-CNN. Query and support images are fed through a shared backbone. The support features are pooled through global averaging and aggregated with the query features. We show here the 1-shot-3-way case without loss of generality.}
    \label{fig:Meta-Learning}
\end{figure*}

\section{Dual-Branch Meta Learning}
\label{sec:meta-learning-dual-branch}
A lot of approaches for few-shot object detection utilize meta learning in order to learn how to generalize for novel categories.
In this section, we first describe the general training scheme for meta learning in~\autoref{sec:meta-learning-training-scheme}.
To realize meta learning, dual-branch approaches use a query and a support branch as we outline in~\autoref{sec:dual-branch-realization}.
In the following subsections, we describe how specific approaches deviate from the general realization.

\subsection{Training Scheme}
\label{sec:meta-learning-training-scheme}
For meta learning, the model is trained in multiple stages.
First of all, the model $\ModelInit$ is trained only on the base dataset $\Dbase$, resulting in $\ModelBase$.
Typically, an episodic training scheme is applied, where each of the $E$ episodes mimics the $N$-way-$K$-shot setting.
This is called meta training.
In each episode $e$ (also known as few-shot task) the model is trained on $K$ training examples of $N$ categories on a random subset $\mathcal{D}_{meta}^e \subset \Dbase, |\mathcal{D}_{meta}^e| = K \cdot N$.
Therefore, the model needs to learn how to discriminate the presented categories in general depending on the input.
Lastly, during meta finetuning the model $\ModelBase$ is trained on the final task, resulting in $\ModelFinal$.

\begin{equation}
\footnotesize
    \ModelInit \underset{e=1 \ldots E}{\overset{\mathcal{D}_{meta}^e \subset \Dbase}{\rightarrow \rightarrow \rightarrow \cdots \rightarrow}} \ModelBase \xrightarrow{\mathcal{D}_{finetune}} \ModelFinal
\end{equation}

If the model is supposed to detect both base and novel categories, it is trained on a balanced set $\mathcal{D}_{finetune} \subset \mathcal{D}$ of $K$ training examples per category, irrespective if it is a base or a novel category.
Otherwise, if we are only interested in the novel categories, the model is trained only on $\mathcal{D}_{finetune} = \Dnovel$.
Note that some approaches do explicitly not finetune on novel categories, but simply apply $\ModelBase$ to novel categories, which is called meta testing.
During meta testing, the model simply predicts novel objects in inference mode, when presented with $K$ annotated examples of $N$ categories.

\subsection{General Realization}
\label{sec:dual-branch-realization}
Dual-branch approaches utilize a two-stream architecture with one query branch $\BranchQ$ and one support branch $\BranchS$ as shown in \autoref{fig:Meta-Learning}.
The input to the query branch $\BranchQ$ is an image $\ImQ$ on which the model should detect object instances, whereas the support branch $\BranchS$ receives the support set \mbox{$\mathcal{D}^\BranchS=\left\{ (\ImS_i, \hat{y}_{o_j}) \right\}_{i=1}^{K \cdot N}$}, with $K$ support images $\ImS_i$ for each of $N$ categories and exactly one designated object $o_j$ and its label $\hat{y}_{o_j}$ per image.
There are three options, how to present the designated object:
First, all training examples are already cropped to the designated object by the ground truth bounding box, as shown in~\autoref{fig:Meta-Learning} bottom.
Second, the full-size image and an additional binary mask, indicating the location of the object, is presented.
Third, as in \cite{SQMG-cpvr2021}, the full-size image can be used and the region with features of the designated object is extracted by RoI Align~\cite{MaskRCNN}.
For all three options, we refer to the presented image as support image $\ImS$.
A support image for a specific category $c$ is denoted as $\ImSc$.

The support branch $\BranchS$ is now supposed to extract relevant features $\featS$ of the support image $\ImS$.
These support features $\featS$ are then aggregated with the features $\featQ$ from the query branch $\BranchQ$, denoted as $\Aggregator(\featQ, \featS)$, in order to guide the detector towards detecting object instances of category $c$ from $\ImSc$ in the query image $\ImQ$.

\begin{figure}[!b]
    \centering
    \input{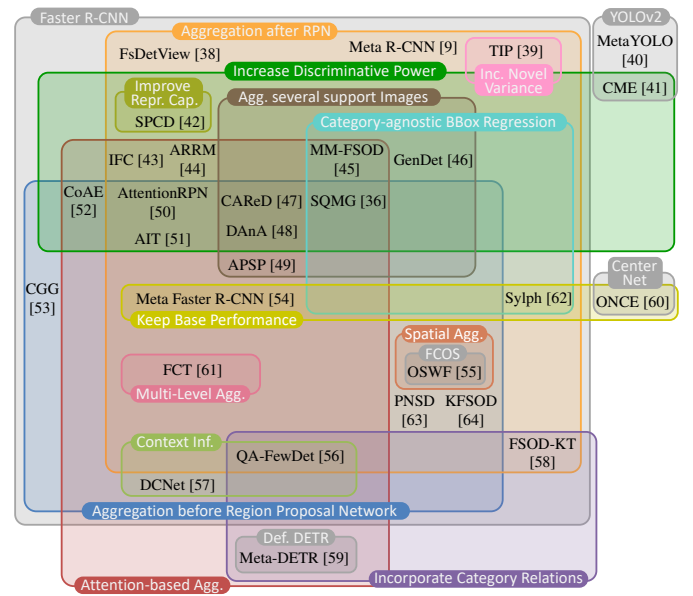}
    \vspace{-3mm}
    \caption{Categorization of dual-branch meta learning approaches. Best viewed in color.}
    \label{fig:dual-branch-meta-categories}
\end{figure}

Note that the following explanation refers to the most basic and widely used architecture for few-shot object detection with meta learning that is depicted in \autoref{fig:Meta-Learning}.
As shown in~\autoref{fig:dual-branch-meta-categories}, the specific approaches may differ in one or multiple points described here and will be explained in detail in the following subsections.

Many approaches build on top of Faster R-CNN~\cite{FasterRCNN-neurips2015} with a ResNet~\cite{ResNet-cvpr2016} backbone.
Often, a siamese backbone is utilized, i.e., the query branch $\BranchQ$ and the support branch $\BranchS$ share their weights.
The backbone features $\featBackboneQ$ of the query branch $\BranchQ$ are further processed by a region proposal network~(RPN) and a RoI align, resulting in the query RoIs $\RoI$.
In the support branch $\BranchS$, the support features from the backbone $\featBackboneS$ are pooled through global averaging, resulting in representative support vectors $\featS$ for each category.
In case of $K > 1$, for each category $c$ the mean of its support vectors is calculated, resulting in one support vector $\featSc$ per category.
These support vectors encode category-specific information which are then used to guide the RoI head in recognizing objects of these categories.
Therefore, query RoIs $\RoI$ and support vectors $\featS$ are aggregated -- shown as $\Aggregator^{RoI}$ in ~\autoref{fig:Meta-Learning} --, in the most simple case by channel-wise multiplication $\Aggregator_{mult}$ as in \autoref{eq:Agg_simple}.

After aggregation, for each of the $N$ categories there are separate RoIs $\RoIAggc$.
Their features are specialized for recognizing objects of the respective category $c$.
These category-specific RoIs $\RoIAggc$ are then fed into a shared RoI head for bounding box regression and binary classification.
Since the aggregated RoIs $\RoIAggc$ already contain category-specific information, the multi-category classification can be replaced by a binary classification that only outputs the information whether the RoI $\RoIAggc$ contains an object of the specific category $c$ or not.
To enforce only one category for each RoI $\RoI$, a softmax layer can be applied afterwards.

Note that the RoI heads for all categories share the same weights.
Therefore, the RoI head must generalize across categories.
With this mechanism, it is theoretically possible to detect objects of novel categories without finetuning on novel categories, but simply meta testing.
This makes meta-learning approaches especially useful for real-world applications, as no further training is required.

During inference, the support features $\featS$ of the few images of $\Dnovel$ can be computed once for all $N$ categories, such that the support branch $\BranchS$ is no longer required.

\subsection{Variants for Aggregation}
The particular dual-branch meta learning approaches differ most in the way the aggregation between query $\featQ$ and support features $\featS$ is implemented.

\subsubsectionspace{Aggregation before the Region Proposal Network}
Typically, the features of the query RoI $\RoI$ are aggregated with the support vectors $\featS$.
However, this requires the region proposal network (RPN) to output at least one RoI for each relevant object.
Otherwise, even the best aggregation method can not help in recognizing the desired object.
However, the RPN is trained only on base categories.
If the novel categories $\Cnovel$ differ a lot from the base categories $\Cbase$, the RPN might fail to output suitable RoIs for recognizing objects of $\Cnovel$.
Therefore, Fan~\etal~\cite{AttentionRPN-cvpr2020} design a so-called AttentionRPN, which effectively aggregates query and support features before the region proposal network.
We denote this by $\Aggregator^{Feat}$in~\autoref{fig:Meta-Learning}.
Specifically, the support features $\featS$ are first average pooled and then aggregated with the query features $\featQ$ by a depth-wise cross correlation.
Afterwards, the RPN is applied onto the enhanced features, resulting in region proposals which are more related to the presented category $c$ of the support image $\ImSc$, and thus improves the recall.
Zhang~\etal~\cite{PNSD-accv2020} (PNSD) build upon this method but replace average pooling with second-order pooling and power normalization~\cite{power-normalization-wacv2019}.
These second-order representations rather function as a detector of features to capture co-occurances than a counter as in average pooling.
This helps to alleviate the harmful variability of features which stem from varying appearances of objects such as color, viewpoint, texture, etc.%

As second-order pooling is limited to linear correlations, in their follow-up work Zhang~\etal~\cite{KFSOD-cvpr2022} (KFSOD) utilize kernelized covariance matrices~\cite{kernelized-covariance-ijcv2021} and Reproducing Kernel Hilbert Space kernels~\cite{kernels-regularization-springer2003} which capture non-linear patterns.
These kernels can factor out spatial order while keeping rich statistics about each region.
Due to this shift-invariance, similar objects that vary in physical location, orientation or viewpoint can be more easily matched.

Furthermore, many others also adopt the idea of AttentionRPN~\cite{AttentionRPN-cvpr2020}, as shown in~\autoref{fig:dual-branch-meta-categories}.
Yet, some use a different aggregation operation, which we will discuss in the following section.

\TakeAway When using Faster R-CNN as detector, an aggregation before the region proposal network leads to better region proposals, and thus less missed detections.

\subsubsectionspace{Aggregation Operation}
In the most simple case, the support vectors $\featS$ and the query features $\featQ$ are multiplied channel-wise:
\begin{equation} \label{eq:Agg_simple}
\footnotesize
\Aggregator_{mult}(\featQ, \featS) = \featQ \odot \featS
\end{equation}
where $\odot$ denotes the Hadamard product.

Moreover, different aggregation operations are explored in the state of the art.
In AttentionRPN~\cite{AttentionRPN-cvpr2020} and GenDet~\cite{GenDet-tnnls2021} support features are convolved / correlated with the query features.
Li~\etal~\cite{OneShotWithoutFinetuning-arxiv2020} (OSWF) use cosine similarity between each element of $\featQ$ and $\featS$, which resembles $\Aggregator_{mult}$ in \autoref{eq:Agg_simple}, but with an additional scaling factor.

Michaelis~\etal~\cite{OSIS-arxiv2018} (OSIS), \cite{ClosingGeneralizationGap-arxiv2020} (CGG) calculate the $\ell^1$-distance at each position and concatenate the resulting similarity features to the query features.
Xiao~\etal~\cite{FSDetView-eecv2020} (FsDetView) use a more complex aggregation operation by combining channel-wise multiplication as in $\Aggregator_{mult}$ with subtraction and query features themselves similar to~\cite{ClosingGeneralizationGap-arxiv2020}:

\begin{equation} \label{eq:Agg_complex}
\footnotesize
\Aggregator(\featQ, \featS) = \left[ \featQ \odot \featS, \featQ - \featS, \featQ \right]
\end{equation}
where $\left[ \cdot, \cdot \right]$ denotes channel-wise concatenation.

Meta Faster R-CNN~\cite{MetaFasterRCNN-aaai2022} builds upon this aggregation,
\thinmuskip=2mu
\medmuskip=3mu plus 2mu minus 3mu
\thickmuskip=4mu plus 5mu minus 2mu
\begin{equation} \label{eq:Agg_MetaFasterRCNN}
\footnotesize
\Aggregator =
\left[ 
    \Phi_{Mult} \left( \featQ \odot \featS \right),
    \Phi_{Sub} \left( \featQ - \featS \right),
    \Phi_{Cat} \left[ \featQ, \featS \right]
\right]
\end{equation}
where $\Phi_{Mult}$ and $\Phi_{Sub}$ and $\Phi_{Cat}$ each denote a small convolutional network with three conv and ReLU layers.

Zhang~\etal~\cite{SQMG-cpvr2021} (SQMG) decided to enhance the query features $\featQ$ by support features $\featS$ with dynamic convolution~\cite{DynamicConvolution-cvpr2020}.
$\featS$ is fed into a kernel generator to generate the weights of the convolution.
Afterwards, the generated weights are convolved with $\featQ$.

\TakeAway The simple channel-wise multiplication of support features $\featS$ and query features $\featQ$ can not fully exploit the information they contain.

\begin{figure}[htb]
  \begin{subfigure}{\linewidth}
  \centering
  \includegraphics[width=\linewidth, trim=0 0mm 0 1mm]{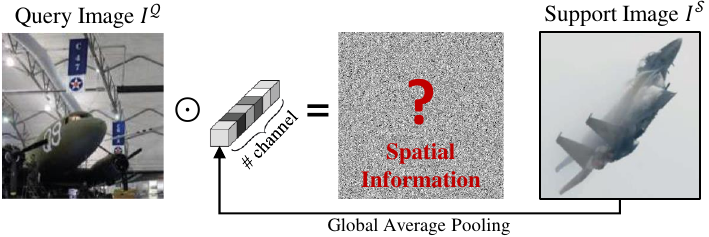}
  \caption{Loss of spatial information due to global average pooling.}
  \label{fig:aggregation-problem-pooling}
  \end{subfigure}\par\bigskip
  \begin{subfigure}{\linewidth}
  \centering
  \includegraphics[width=\linewidth, trim=0 5mm 0 5mm]{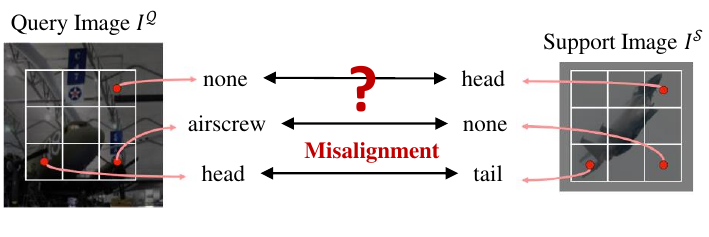}
  \caption{Spatial misalignment due to convolution-based aggregation.}
  \label{fig:aggregation-problem-convolution}
  \end{subfigure}\par\bigskip
  \vspace{-5mm}
  \caption{Common aggregation problems. Images based on \cite{DAnA-tmm2021}.}
  \label{fig:aggregation-problems}
\end{figure}

\subsubsectionspace{Keep Spatial Information for Aggregation}
As opposed to aggregating support features via average pooling, others (see~\autoref{fig:dual-branch-meta-categories})
propose to utilize spatial information.
For the object detection task, objects are located by bounding boxes.
However, not every part of that bounding box is occupied by the object and, therefore, does not contain relevant information about the respective category.
With average pooling however, these irrelevant features are aggregated into the support vector.
Moreover, with global average pooling spatial information is completely lost, as illustrated in~\autoref{fig:aggregation-problem-pooling}.

Therefore, Li~\etal~\cite{OneShotWithoutFinetuning-arxiv2020} (OSWF) first pool support features to the same spatial dimension as the query RoI $\RoI$. Afterwards, these pooled features are concatenated to the query RoI $\RoI$.
Finally, $1{\times}1$ convolutions are used to compare structure-aware local features.

However, Chen~\etal~\cite{DAnA-tmm2021} argue that a convolution of query features $\featQ$ and support features $\featS$ is less suitable, since the objects in query images $\ImQ$ and support images $\ImS$ are generally not aligned in the same way, as shown in~\autoref{fig:aggregation-problem-convolution}.
Therefore, they design an attention-based aggregation as described in the following section.

\TakeAway In order to incorporate the valuable spatial information of a support image $\ImS$, its features should not be simply averaged for aggregation.

\subsubsectionspace{Attention-based Aggregation}
\label{sec:meta-aggregation-attention}

Lately, attention mechanisms could significantly improve performance on many vision tasks~\cite{survey-vision-transformer-tpami2022}.
Thus, it is not surprising, that the aggregation of support and query features also benefits from incorporating attention mechanisms.
These attention mechanisms range from traditional, over non-local~\cite{NonLocal-cvpr2018} to multi-head attention as in transformers~\cite{Transformer-neurips2017}.
We will discuss all of them in the following.

Chen~\etal~\cite{DAnA-tmm2021} (DAnA) aim to incorporate the spatial correlations between query image $\ImQ$ and support image $\ImS$ but also take into account that these images are generally not aligned (see~\autoref{fig:aggregation-problem-convolution}).
Therefore, a dual-awareness attention first highlights relevant semantic features of the respective category on the support features $\featS$ and suppresses background information.
Afterwards, the spatial correlations are incorporated with an attention-based aggregation.
This spatial misalignment is also addressed in Meta Faster R-CNN~\cite{MetaFasterRCNN-aaai2022}.
Using two attention modules, the support and RoI features are first spatially aligned and then the foreground regions are highlighted.

Wang~\etal~\cite{IFC-apin2022} (IFC) first use a self-attention module on top of average-pooled and max-pooled query features to separately mine local semantic and detailed texture information.
Afterwards, with a new feature aggregation mechanism based on a learnable soft-threshold operator~\cite{soft-threshold-shrinkage-trans-ind-inform2019}, redundant information can be shrinked while enhancing feature sensitivity and stability for both novel and base categories.

Huang~\etal~\cite{ARRM-electronimaging2022} (ARRM) aim to achieve a better interaction of support and query features by designing an attention-based affinity relation reasoning module consisting of several convolutions and matrix multiplications of different features.
With an additional global-average-pooling branch also the global semantic context of the support features is integrated.
Using this attention-based module for aggregation, misclassifications can be reduced.

Hsieh~\etal~\cite{CoAE-neurips2019} (CoAE) propose a co-attention method in order to make the query features $\featQ$ attend to the support features $\featS$ and vice versa.
Therefore, two mutual non-local operations~\cite{NonLocal-cvpr2018} are utilized which receive inputs from both $\featQ$ and $\featS$.
This helps the RPN to compute region proposals which are able to better locate objects of the category $c$ from the support image $\ImSc$.
Moreover, Hsieh~\etal~\cite{CoAE-neurips2019} propose a subsequent squeeze-and-co-excitation method -- extending the squeeze-and-excitation of SENet~\cite{SENet-cvpr2018} -- in order to highlight correlated feature channels to detect relevant proposals and eventually the target objects.%
A similar co-attention is utilized by Hu~\etal~\cite{DCNet-cvpr2021} (DCNet).

With AIT Chen~\etal~\cite{AIT-cvpr2021} push the idea of CoAE~\cite{CoAE-neurips2019} a little further.
Instead of using a single non-local block, multi-head co-attention is utilized for aggregating query and support features before the region proposal network.
Let $\mathbf{V}$, $\mathbf{K}$ and $\mathbf{Q}$ be the \texttt{value}, \texttt{key} and \texttt{query} of a Transformer-based attention~\cite{Transformer-neurips2017}.
Similarly to the co-attention in CoAE, \texttt{query} features stem from another branch:
\begin{equation}
\footnotesize
    \mathbf{F}^\mathcal{Q} = attn \left( 
        \mathbf{V}^\mathcal{Q}, \mathbf{K}^\mathcal{Q}, \mathbf{Q}^\mathcal{S}
    \right), \quad\quad
    \mathbf{F}^\mathcal{S} = attn \left( 
        \mathbf{V}^\mathcal{S}, \mathbf{K}^\mathcal{S}, \mathbf{Q}^\mathcal{Q}
    \right)
    \label{eq:agg-AIT}
\end{equation}
where superscripts $\mathcal{Q}$ and $\mathcal{S}$ denote whether features are from the query or support branch.
The resulting features $\mathbf{F}^\mathcal{Q}$ encode related visual characteristics of both the query image $\ImQ$ and the support image $\ImS$, which helps the region proposal network to predict RoIs related to $\ImS$.
According to Chen~\etal~\cite{AIT-cvpr2021}, this improves the accuracy compared to the non-local attention block~\cite{NonLocal-cvpr2018} in CoAE~\cite{CoAE-neurips2019}.
After the RPN, AIT~\cite{AIT-cvpr2021} uses a transformer-based encoder-decoder architecture for transforming the RoIs $\RoI$ to emphasize visual features corresponding to the given support image $\ImS$.

A similar aggregation to $\mathbf{F}^\mathcal{S}$ in \autoref{eq:agg-AIT} is also used in Meta-DETR~\cite{Meta-DETR-tpami2022} and APSP~\cite{APSP-wacv2022}.
However, both Meta-DETR and APSP first enhance query or support features as we will describe in the following sections.

\TakeAway 
The spatial information of a support image $\ImS$, as well as its relation to a query image $\ImQ$, is best incorporated through transformer-based attention mechanisms~\cite{NonLocal-cvpr2018, Transformer-neurips2017}.

\subsubsectionspace{Multi-Level Aggregation}
So far, support and query features were only aggregated after feature extraction in the backbone.
However, Han~\etal~\cite{FCT-cvpr2022} (FCT) argue that multi-level feature interactions between the query and support branch could better align the features.
Therefore, they come up with a novel \underline{F}ully \underline{C}ross-\underline{T}ransformer based on the improved Pyramid Vision Transformer PVTv2~\cite{PVTv2-cvm2022}.
The FCT model consists of three interaction stages between query and support in the backbone and one additional interaction stage in the detection head.
Finally, a pairwise matching similar to the one from AttentionRPN~\cite{AttentionRPN-cvpr2020} outputs the final detections.

\TakeAway
Aggregation of low-, mid- and high-level features can boost the performance.

\begin{figure}[htb]
    \centering
    \includegraphics[width=\linewidth]{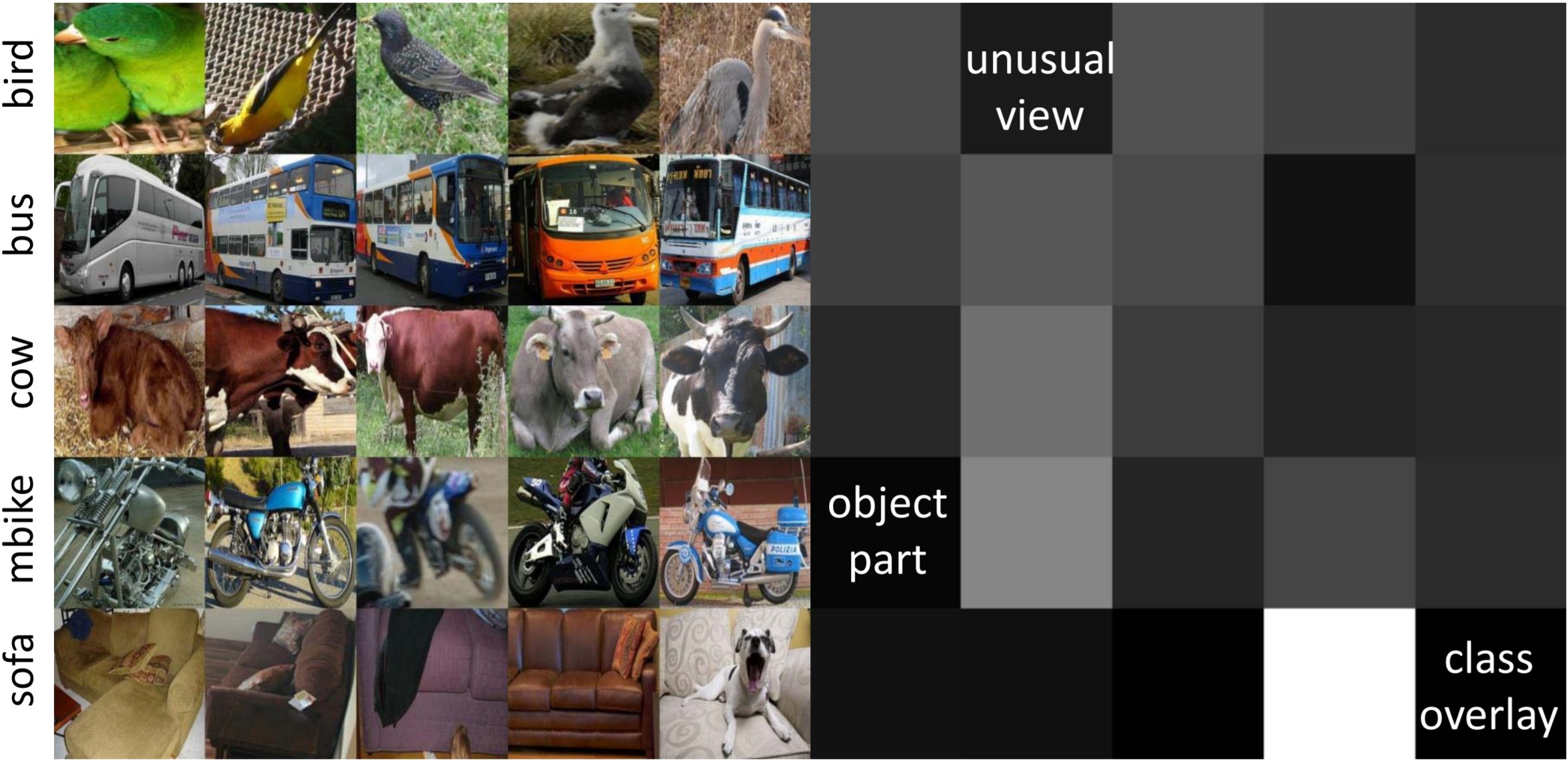}
    \caption{Different amount of information for several support images of the same category. Image from GenDet~\cite{GenDet-tnnls2021}.}
    \label{fig:aggregation-several-images}
\end{figure}

\subsubsectionspace{Aggregation of Several Support Images}
\label{sec:meta-aggregation-several-support}
In the general approach, to fuse all support images of category $c$, the mean of their features is calculated:
\begin{equation}
\footnotesize
    \left\{\ImSc_i\right\}_{i=1}^K: \quad \featSc = \frac{1}{K}\sum\limits_{i=1}^{K}{\featSc_i}
\end{equation}
However, not all support images provide the same amount of information for the respective category as shown in~\autoref{fig:aggregation-several-images}.
Unusual object views, object parts, or even occlusion by objects of other categories impair the discriminative power if support features $\featSc_i$ are simply averaged.

Therefore, a weighted average is proposed in GenDet~\cite{GenDet-tnnls2021}.
The weight $w_i$ for each support image $\ImSc_i$ is computed by the similarity between the single-shot and the mean detector and learned during training.
\begin{equation} \label{eq:weightedSumFeatSc}
\footnotesize
    \left\{\ImSc_i\right\}_{i=1}^K: \quad \featSc = \frac{1}{K}\sum\limits_{i=1}^{K}{w_i \cdot \featSc_i}
\end{equation}

Quan~\etal~\cite{CAReD-displays2022} (CAReD) follow a similar approach.
However, the weight $w_i$ is determined by the softmax over the correlation between the support features $\featSc_i$ and all other support features $\left\{\featSc_j\right\}_{j=1}^K$ of the same category $c$.
Due to the softmax, the weighting factors already sum up to 1 
and the factor $\sfrac{1}{k}$ is omitted.

DAnA~\cite{DAnA-tmm2021}, SQMG~\cite{SQMG-cpvr2021} as well as APSP~\cite{APSP-wacv2022} incorporate the similarity of query $\featQ$ and different support features $\featS$: 
In DAnA~\cite{DAnA-tmm2021} support features $\featSc_i$ of $K$ different images $\left\{\ImSc_i\right\}_{i=1}^K$  are first aggregated independently with the query features $\featQ$ based on the correlation between query and support.
As the importance of each support images $\ImSc_i$ is already incorporated, the resulting $K$ aggregated features can be simply averaged:
\begin{equation}
\footnotesize
    \left\{\ImSc_i\right\}_{i=1}^K: \quad \Aggregator(\featQ, \featSc) = \frac{1}{K}\sum\limits_{i=1}^{K}{\Aggregator(\featQ, \featSc_i)}
\end{equation}

In SQMG~\cite{SQMG-cpvr2021} the support features $\featSc_i$ of multiple support images $\ImSc_i$ are weighted according to their similarity with the query features $\featQ$ using an attention mechanism.
First, the similarity is computed with a relation network~\cite{RelationNetwork-cvpr2018}.
Afterwards, the weighting values $w_i$ for support features $\featSc_i$ are computed with a softmax on the similarity score.
The final support features are achieved by a weighted sum as in \autoref{eq:weightedSumFeatSc}.

Lee~\etal~\cite{APSP-wacv2022} (APSP) first use a multi-head attention to refine each individual support vector $\featSc_i$ by incorporating all other support vectors of the same category $c$.
Afterwards, instead of computing one single support vector, all $K$ support vectors $\left\{\featSc_i\right\}_{i=1}^K$ are utilized in a second multi-head attention for aggregation with the query features.
Thus, not all variances of different support images need to be incorporated in a single support vector and therefore leads to more robust features.

\TakeAway As not all support images provide the same amount of information, their individual relevance should be incorporated, as shown in \autoref{fig:aggregation-several-images}.

\subsection{Incorporate Relations between Categories}
\label{sec:meta-relations-categories}
Han~\etal~\cite{QAFewDet-iccv2021} (QA-FewDet) highlight the problem that many dual-branch meta-learning approaches work as a kind of single-category detector without modeling multi-category relations.
However, especially for novel categories resembling base categories, these relations can help in correctly classifying objects (e.g., a motorbike is more similar to a bicycle than to an airplane).

Therefore, in contrast to using visual features only, Kim~\etal~\cite{FewShotKnowledgeTransfer-smc2020} (FSOD-KT) additionally incorporate linguistic features.
Before aggregation, the support vectors $\featS$ are fed through a knowledge transfer module which exploits semantic correlations between different categories.
This knowledge transfer module is implemented by a graph convolutional network~\cite{GCN-iclr2017}.
The input to this graph convolutional network is a graph where each node represents one category, and the values on the edges represent the similarities between linguistic category names.
However, this is only applicable if all categories have predefined and distinct category names, and might be hard to transfer to, e.g., medical imaging.

Han~\etal~\cite{QAFewDet-iccv2021} (QA-FewDet) also utilize graph convolutions but do not rely on the linguistic category names.
In contrast, they build a heterogeneous graph which enhances support vectors $\featS$ with multi-category relations in order to better model their relations and incorporate features from similar categories.
Moreover, their heterogeneous graph also aligns support and query features:
Since the support features $\featSc$ of one category $c$ are only extracted from few support images, there might be a huge discrepancy to query RoIs $\RoIc$ that actually belong to the same category $c$.
Therefore, the heterogeneous graph also contains pairwise edges between RoIs, in order to mutually adapt features of $\featSc$ and $\RoIc$ and reduce their discrepancy.

Although not using graph convolutions, Zhang~\etal~\cite{Meta-DETR-tpami2022} (Meta-DETR) also incorporate relations between different categories by transforming their support features.
The authors introduce a correlation aggregation module, which is able to simultaneously aggregate multiple support categories in order to capture their inter-class correlation.
This helps in reducing misclassification and enhances generalization to novel categories.
First, the query features $\featQ$ are matched with multiple support features $\featS$ simultaneously by utilizing attention modules~\cite{Transformer-neurips2017}.
Afterwards, task encodings help to differentiate these support categories.

\TakeAway Incorporating the relations between different categories helps in better representing and classifying the data-sparse novel categories $\Cnovel$.

\subsection{Increase Discriminative Power}
\label{sec:dual-branch-meta-discriminative}
After aggregation, for each RoI $\RoI$, there exist N category-specific RoIs $\RoIAggc$, which are classified independently.
If the support features $\featS$ for different categories are too similar, this independent classification might lead to ambiguities.
Therefore, some approaches use an additional meta loss to enforce the support features $\featS$ to be as diverse as possible.
Most often (e.g., in~\cite{MetaRCNN-iccv2019, FSDetView-eecv2020, TIP-cvpr2021, QAFewDet-iccv2021, FewShotKnowledgeTransfer-smc2020}), the support features $\featS$ are classified, and a simple cross-entropy loss is applied.
This encourages the support vectors to fall in the category the respective object belongs to. More advanced approaches utilize techniques from metric learning to increase the discriminative power, as described in the following.

GenDet~\cite{GenDet-tnnls2021} and Meta-DETR~\cite{Meta-DETR-tpami2022} use a loss based on cosine similarity for more discriminative support vectors.
First of all, the support vectors $\featS$ are normalized.
Afterwards, for each pair of support vectors $\left( \featScnum[i], \featScnum[j] \right)$ the cosine similarity is computed which results in a similarity matrix $A \in \mathbb{R}^{N{\times}N}$, where $N$ is the number of different categories.
With an $\ell^1$ loss, the similarity matrix $A$ is constrained to be close to the identity matrix $I_N \in \mathbb{R}^{N{\times}N}$.
Intuitively speaking, this results in minimizing the similarity between different support vectors and maximizing the discriminative ability of each support vector, i.e., a high margin between different support vectors.

Wang~\etal~\cite{IFC-apin2022} (IFC), Kobayashi~\cite{SPCD-iciap2022} (SPCD) and Huang~\etal~\cite{ARRM-electronimaging2022} (ARRM) also use a cosine loss, but in ARRM an additional margin is added to further increase discrimination and reduce misclassification.

MM-FSOD~\cite{MM-FSOD-arxiv2020} uses the pearson distance for aggregating $\featS$ and $\featQ$.
Compared to cosine similarity, Pearson distance first normalizes each dimension with the mean of all dimensions, resulting in a smaller inner-class variance.
Therefore, there is no need for designing a special distance loss function, and the simple cross-entropy loss can be utilized.

Li \etal~\cite{CME-cvpr2021} (CME) propose an adversarial training procedure for min-max-margin:
Next to a loss for increasing the margin, the features of novel categories are disturbed to reduce the discriminative power of their support vectors and, thus, decrease the margin.
To be precise, the most discriminative pixels are erased in an adversarial manner by backpropagating the gradient to the input support image.
With this approach, CME~\cite{CME-cvpr2021} is capable to accurately detect more objects with fewer false positives.

For the meta learning approach, the detector is supposed to detect objects in a query image $\ImQ$ that are of the same category $c$ as the object in the support image $\ImSc$.
Due to this problem definition, meta leaning approaches tend to focus on separating foreground from background instead of distinguishing different categories, as noted by Zhang~\etal~\cite{SQMG-cpvr2021} (SQMG).
This often leads to false positives, i.e., predicted bounding boxes, even though the query image $\ImQ$ does not contain any instance of the regarded category $c$.
However, it is equally important that the detector can distinguish different categories and identify which object categories are \emph{not} present in the query image.

Therefore, in AttentionRPN~\cite{AttentionRPN-cvpr2020} a multi-relation detector as well as a two-way contrastive training strategy is proposed.
The multi-relation detector incorporates global, local and patch-based relations between support features $\featS$ and query RoIs $\RoI$ in order to measure their similarity.
The outputs of all three matching modules are summed to give the final matching score.
Many others~\cite{CAReD-displays2022, FCT-cvpr2022, PNSD-accv2020, KFSOD-cvpr2022} adopt or build upon this multi-relation detector.
The additionally proposed two-way contrastive training strategy is implemented as follows:
In addition to a positive support image $\ImSc$, a negative support image $I^{\mathcal{S}, n}$ is used from an object category $n \in \mathcal{C}\backslash \{ c \}$ that is not present in the query image $\ImQ$.
This two-way-contrastive training strategy is adapted by DAnA~\cite{DAnA-tmm2021} and, similarly, by CAReD~\cite{CAReD-displays2022}.
Zhang~\etal~\cite{SQMG-cpvr2021} (SQMG) extend the contrastive loss with an adaptive margin~\cite{AdaptiveMargin-cvpr2020} in order to separate the different categories by a proper distance.
The adaptive margin incorporates semantic similarity of the categories by word embeddings~\cite{WordEmbeddingGlove-emnlp2014}.

A second problem highlighted by Zhang~\etal~\cite{SQMG-cpvr2021} (SQMG) is the extreme imbalance of many background proposals vs. few foreground proposals, which impedes distance metric learning.
To combat the foreground-background imbalance, the authors use a focal loss~\cite{RetinaNet-iccv2017} which down-weights the easy background proposals and focuses on the hard negatives.

CoAE~\cite{CoAE-neurips2019} uses an additional margin-based loss to improve the ranking of the regions of interest in the region proposal network.
Those regions of interest with a high similarity to the object in the support image $\ImS$ should be at the top of the ranking, since only the top 128 RoIs will be further processed.
Therefore, the authors designed a margin-based metric to predict the similarities for all regions of interest.
Chen~\etal~\cite{AIT-cvpr2021} (AIT) adopt this margin-based ranking loss.

In typical episodic training, only $N$ categories are presented in each episode.
According to Liu \etal~\cite{GenDet-tnnls2021} (GenDet), this could lead to a low discriminative ability of the extracted features, as only the sampled categories are distinguished.
Thus, their approach GenDet~\cite{GenDet-tnnls2021} utilizes an additional reference detector during training, where all base categories $\Cbase$ need to be distinguished.
The index of a specific base category stays the same over all episodes.
Via an additional loss, both detectors are constrained to output similar results.
This guides the backbone to extract more discriminative features.

\TakeAway In order to increase the discriminative power and differentiate between several categories, ideas from metric learning such as similarity metrics, as well as contrastive training should be employed.

\subsection{Improve representation capability}
Kobayashi~\cite{SPCD-iciap2022} (SPCD) emphasizes that during base training all other non-base categories are treated as negative.
This leads to insufficient expressive power to identify novel categories.
Therefore, they introduce an additional self-supervised module:
With selective search~\cite{selective-search-ijcv2013} rectangular regions different to those from base categories are extracted, and the network is taught to detect the same regions before and after applying strong data augmentation in a self-supervised manner.

\subsection{Proposal-free Detectors}
Most approaches build on top of the two-stage detector Faster R-CNN~\cite{FasterRCNN-neurips2015}.
However, these approaches need to deal with possibly inaccurate region proposals and the decision of whether to aggregate support features $\featS$ and query features $\featQ$ before or after the region proposal network or both.
When utilizing proposal-free detectors, $\featS$ and $\featQ$ can simply be aggregated after feature extraction and before classification and bounding box regression.

Some approaches utilize simple one-stage detectors, such as YOLOv2~\cite{YOLOv2-cvpr2017} in MetaYOLO~\cite{MetaYOLO-iccv2019} and CME~\cite{CME-cvpr2021} or RetinaNet~\cite{RetinaNet-iccv2017} in DAnA~\cite{DAnA-tmm2021}.
Others build on top of anchor-free detectors like CenterNet~\cite{CenterNet_ObjectsAsPoints-arxiv2019} in ONCE~\cite{ONCE-cvpr2020} or FCOS~\cite{FCOS-iccv2019} in Li~\etal~\cite{OneShotWithoutFinetuning-arxiv2020} (OSWF) and GenDet~\cite{GenDet-tnnls2021}.
The transformer-based detector Deformable DETR~\cite{DeformableDETR-iclr2021} is utilized in Meta-DETR~\cite{Meta-DETR-tpami2022}.
Meta-DETR aggregates support features $\featS$ and query features $\featQ$ after the shared backbone.
Subsequently, a category-agnostic transformer architecture predicts the objects.

\TakeAway While most approaches build on top of \mbox{Faster R-CNN}, proposal-free detectors are easier to implement. Especially transformer-based architectures such as Meta-DETR~\cite{Meta-DETR-tpami2022} already surpass other approaches.

\subsection{Keep the Performance on Base Categories}
In order to better detect base categories and prevent catastrophic forgetting, Han~\etal~\cite{MetaFasterRCNN-aaai2022} (Meta Faster R-CNN) use an additional branch following the original Faster R-CNN~\cite{FasterRCNN-neurips2015} architecture.
As Meta Faster R-CNN already aggregates query features $\featQ$ and support features $\featS$ before the region proposal network, only the weights for the backbone are shared between those two branches.
After meta training on the base categories $\Cbase$, the weights of the backbone are fixed and the RPN and RoI head for the base category branch are trained.
Finally, the other branch is adapted or simply applied to novel categories with meta finetuning or meta testing, respectively (see~\autoref{sec:meta-learning-training-scheme} for terminology definitions).
As the first branch stays fixed, the performance for base categories $\Cbase$ won't drop due to meta finetuning.

For the incremental learning approaches ONCE~\cite{ONCE-cvpr2020} and Sylph~\cite{Sylph-cvpr2022} the weights for the already learned categories also stay fixed.
Instead of a softmax-based classifier, Sylph uses several independent binary sigmoid-based classifiers (one for each category), such that the categories do not influence each other.
For each novel category $c$, a hypernetwork on top of the support branch $\BranchS$ generates the weights for its classifier.
Thus, no meta finetuning is required.

\subsection{Increase the Variance of Novel Categories}
TIP~\cite{TIP-cvpr2021} expands the few training examples for novel categories with data augmentation techniques such as Gaussian noise or cutout.
However, naively adding data augmentation impairs detection performance.
Therefore, Li \etal~\cite{TIP-cvpr2021} (TIP) use an additional transformed guidance consistency loss, implemented by $\ell^2$ norm, which constrains support vectors $\featS_i, \featS_j$ generated by original image $\ImS_i$ and transformed image $\ImS_j = \phi ( \ImS_i )$ to be close to each other.
This results in more similar and representative support vectors even for different support images, thus improving detection performance of novel categories.
Moreover, during training, the query branch $\BranchQ$ also receives transformed as well as original images.
The features of the transformed query image $\ImQ$ are fed into the region proposal network to predict regions of interest (RoIs). These RoIs are then cropped from the features of the original non-transformed query image via RoI Align~\cite{MaskRCNN}.
This forces the detector to predict consistent RoIs independent of the transformation used for the query image.

\subsection{Incorporate Context Information}
Typically, by applying RoI pool or RoI align, region proposals are pooled to a specific squared size of, e.g., $7{\times}7$.
However, this might lead to information loss during training, which could be remedied with abundant training data.
With only few training examples available, this information loss could result in misleading detections.
Therefore, DCNet~\cite{DCNet-cvpr2021} uses three different resolutions and performs parallel pooling.
Similar to the pyramid pooling module in the PSPNet~\cite{PSPNet-cvpr2017} for semantic segmentation, this helps to extract context information, where larger resolutions help to focus on local details, while smaller resolutions help to capture holistic information.
In contrast to the pyramid pooling module, the branches are fused with attention-based summation.

Han~\etal~\cite{QAFewDet-iccv2021} (QA-FewDet) find that query RoIs $\RoI$ might be noisy and may not contain complete objects.
Therefore, they built a heterogeneous graph which uses graph convolutional layers~\cite{GCN-iclr2017}. 
Pairwise edges between proposal nodes incorporate both local and global context of different RoIs in order to improve classification and bounding box regression.

\subsection{Category-agnostic Bounding Box Regression}
Even though parameters for binary classification and bounding box regression are shared for all categories, most approaches compute them for each category-specific RoI independently.
In contrast, GenDet~\cite{GenDet-tnnls2021}, MM-FSOD~\cite{MM-FSOD-arxiv2020}, SQMG~\cite{SQMG-cpvr2021}, and Sylph~\cite{Sylph-cvpr2022} share the bounding box computation among different categories.
This follows the intuition that even though different categories vary in their visual appearances, regression of bounding box values has common traits.
Moreover, it saves computation overhead.

\subsection*{Summary of Best Performing Dual-Branch Meta Learning Approaches}
\label{sec:dual-branch-best-approaches}
In the following, we summarize selected dual-branch meta learning approaches that perform best on FSOD benchmark datasets (see \autoref{sec:experiments}), in order to highlight their key concepts.
\vspace{1mm}

\textbf{Meta-DETR}~\cite{Meta-DETR-tpami2022} is the first approach building on top of the transformer-based detector DETR.
Without depending on accurate region proposals, Meta-DETR circumvents the challenge to adapt these for novel categories.
Moreover, in its attention-based aggregation module, the correlation between different categories is incorporated, which reduces missclassification.
With an additional loss based on cosine similarity the learned features are more discriminative and, thus, enhance generalization.
\vspace{0.5mm}

\textbf{FCT}~\cite{FCT-cvpr2022} also uses a transformer, but instead of DETR, the ResNet backbone of Faster R-CNN is simply replaced by the improved Pyramid Vision Transformer PVTv2.
However, additionally, support and query features are aggregated at multiple levels to better align the features.
Moreover, a multi-relation detector computes similarities between support and query features to output the final detections.
\vspace{0.5mm}

\textbf{IFC}~\cite{IFC-apin2022} does not build on top of transformers, but utilizes an interactive self-attention module to capture the discriminating features from scarce novel categories.
Moreover, a novel feature aggregation mechanism is introduced, which aims at shrinking redundant information while enhancing feature sensitivity and stability for both novel and base categories.
Finally, an orthogonal cosine loss enhances foreground distinguishability.
\vspace{0.5mm}

One of the few approaches not requiring finetuning is \textbf{SQMG}~\cite{SQMG-cpvr2021}.
In SQMG both support and query features are enhanced through mutual guidance.
First, this helps to generate more category-aware region proposals.
Second, the individual relevance of multiple support images is also incorporated.
Moreover, SQMG focuses on correct classification with different training techniques:
To alleviate confusion of similar categories a two-way contrastive training strategy with an adaptive margin is employed.
To combat the imbalance between many background proposals vs. few foreground proposals an additional focal loss is incorporated.
Finally, the bounding box regression is shared among different categories in order to focus on classification.

\subsection*{Conclusion on Dual-Branch Meta Learning Approaches}
Dual-branch meta learning approaches are very common in few-shot object detection.
They enable fast adaption for novel categories or can even be applied to novel categories without finetuning but with a simple forward pass, i.e., meta testing.
This is especially useful for real-world applications.
However, they require a complex episodic training scheme, as described in \autoref{sec:meta-learning-training-scheme}.
Nevertheless, by utilizing attention-based aggregations and incorporating metric learning techniques, dual-branch meta learning approaches can achieve state-of-the-art results, as we will discuss in \autoref{sec:experiments}.

\section{Single-Branch Meta Learning}
\label{sec:meta-learning-single-branch}

\begin{figure}[htb]
    \centering
    \input{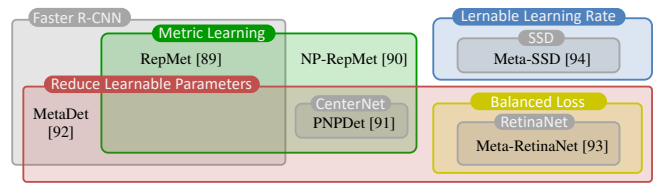}
    \vspace{-6mm}
    \caption{Categorization of single-branch meta learning approaches. Best viewed in color.}
    \label{fig:single-branch-meta-categorization}
\end{figure}

Single-branch architectures for few-shot object detection follow another approach.
Since there are no query and support branches, the general architecture resembles the architecture for generic object detectors such as Faster R-CNN~\cite{FasterRCNN-neurips2015} (see Appendix \autoref{fig:FasterRCNN}).
However, there is no single approach from which others deviate.
Still, all approaches use episodic training as described in \autoref{sec:meta-learning-training-scheme}, which is typical for meta learning.
In~\autoref{fig:single-branch-meta-categorization}, we display our categorization of single-branch meta learning approaches, which we further describe in the following.

\subsection{Metric Learning}
Similarly to dual-branch meta learning, metric learning plays a key role for single-branch approaches.

One of the first approaches for few-shot object detection -- RepMet~\cite{RepMet-cvpr2019} -- defines the FSOD task as a distance metric learning problem.
For localization, RepMet simply uses the regions of interest (RoIs) $\RoI$ from Faster R-CNN~\cite{FasterRCNN-neurips2015}.
Embedded feature vectors $f^+$ of these RoIs are compared to multiple learned representatives for each category, in order to determine the category for a RoI.
To learn an adequate feature embedding, an additional embedding loss is used, which enforces a minimum margin between the distance of the embedded vector to the closest representative of the correct category and the distance to the closest representative of the wrong category.

RepMet uses the positive region proposals of a category, but discards its negative proposals.
However, for learning the embedding space, negative -- especially hard negative -- proposals are essential.
Therefore, NP-RepMet~\cite{NP-RepMet-neurips2020} also learns negative embedded feature vectors $f^-$ and negative representative vectors per category.
The embedding spaces for the representatives are learned by utilizing a triplet loss~\cite{TripletLoss-jmlr2009}.

PNPDet~\cite{PNPDet-wacv2021} uses cosine similarity for distance metric learning of the objects' categories to allow for better generalization to novel categories.
Cosine similarity computes the similarity of the input image's features with learned prototypes of each category.

\TakeAway Metric learning helps in creating more discriminative features for better distinguishing between different categories.

\subsection{Reduce Learnable Parameters}
Since few training examples of novel categories might not be sufficient to train a deep neural network, some approaches reduce the number of learnable parameters for few-shot finetuning.

After training MetaDet~\cite{MetaDet-iccv2019} on the base dataset, category-agnostic weights (i.e. backbone and RPN of Faster R-CNN) are frozen, and an episodic training scheme is applied to learn how to predict category-specific weights first for the base categories $\Cbase$ and then for the novel categories $\Cnovel$.
For inference, the meta model can be detached, and the detector looks like the standard Faster R-CNN.

Li~\etal~\cite{MetaRetinaNet-bmvc2020} (MetaRetinaNet) reduce the number of learnable parameters by freezing all backbone layers after training on $\Dbase$ and instead learn coefficient vectors $v$ initialized to ones.
These learnable coefficient vectors $v$ are multiplied with the convolution weights $w$, resulting in a modified convolution operation: \mbox{$\mathrm{f_{out}} = \mathrm{f_{in}} \otimes \left( w \odot v \right) \oplus b$}.

Zhang~\etal~\cite{PNPDet-wacv2021} (PNPDet) freeze the whole network after training on $\Dnovel$.
For few-shot finetuning, a second small sub-network is introduced for learning to classify the novel categories $\Cnovel$.
This disentangling of novel and base categories prevents a decreasing performance on base categories.

\TakeAway When training on data-scarce $\Dnovel$, the number of learnable parameters should be reduced.

\subsection{Learnable Learning Rate}
Fu~\etal~\cite{MetaSSD-ieee2019} design their Meta-SSD such that the model's parameters can adjust fast -- with just one parameter update -- to the novel categories $\Cnovel$.
All parameters from original SSD detector~\cite{SSD-eccv2016} get an additional learnable learning rate.
During meta learning, these learning rates are learned individually by a meta learner from the distribution of the current task, resulting in neither overfitting nor underfitting.

\subsection{Balanced Loss Function}
Li~\etal~\cite{MetaRetinaNet-bmvc2020} highlight that for meta training, in each episode different training examples of different categories are sampled, and they achieve different performances.
This performance imbalance hinders stability and makes it difficult to adapt the model to novel categories.
Thus, in their MetaRetinaNet a balancing loss is introduced, which constrains the detector to achieve similar performance across episodes.

\subsection*{Conclusion on Single-Branch Meta Learning Approaches}
Single-branch meta learning approaches are much less explored in few-shot object detection.
Thus, more advanced dual-branch approaches or transfer learning approaches are able to surpass the approaches presented here.

\section{Transfer Learning}
\label{sec:transfer-learning}
\begin{figure}[!t]
    \centering
    \includegraphics[width=\linewidth]{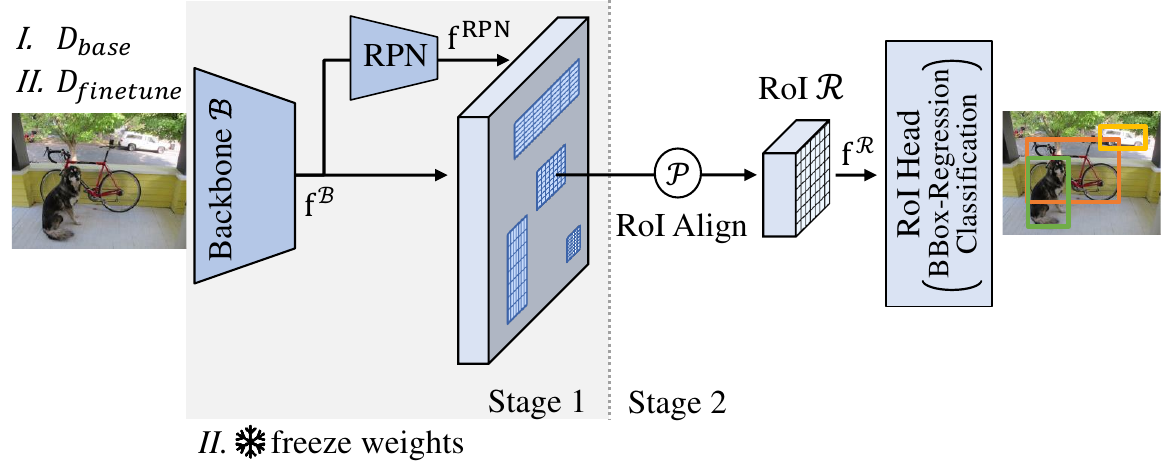}
    \caption{Realization with transfer learning
    }
    \label{fig:transfer-learning}
\end{figure}

Meta learning approaches depend on a complex episodic training.
In contrast, transfer learning approaches utilize a fairly simple two-phase approach on a single-branch architecture, most often a Faster R-CNN~\cite{FasterRCNN-neurips2015}, as first proposed by Wang \etal~\cite{TFA-icml2020} (TFA) and shown in \autoref{fig:transfer-learning}.

In the first phase, the detector is trained on the base categories $\Cbase$.
Afterwards, all detector weights are frozen except for RoI head, which is responsible for bounding box regression and classification.
In the second phase, transfer learning is performed, by finetuning the last layers on the base categories $\Cbase$ and novel categories $\Cnovel$.
For finetuning, the training set is composed of balanced subsets of base category data $\Dbase$ and novel category data $\Dnovel$ with $K$ shots for each of the base and novel categories.
The only modification to \mbox{Faster R-CNN} is the use of cosine similarity for classification, which is crucial to compensate for differences in feature norms of base categories $\Cbase$ and novel categories $\Cnovel$, as analyses in \cite{FSCE_FSOD-cvpr2021} have shown.
Wang \etal~\cite{TFA-icml2020} showed that this simple approach is sufficient to adequately learn the novel categories $\Cnovel$ and outperform earlier meta learning approaches that are more complex.

\begin{figure}[bh]
    \centering
    \input{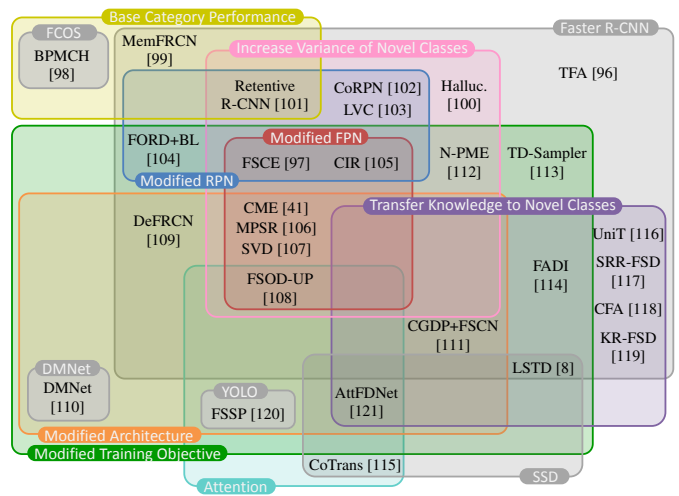}
    \vspace*{-7mm}%
    \caption{Transfer learning approaches categorized by detector architecture and types of modifications.
    Best viewed in color.}
    \label{fig:categroies-transfer-learning}
\end{figure}

Building upon this simple approach, many modifications have been proposed.
\autoref{fig:categroies-transfer-learning} shows all transfer learning approaches categorized by the architecture employed and by their modifications.
Below, we describe all the proposed modifications, grouped by the categories shown.

\subsection{Modifications of the Region Proposal Network}
\label{subsec:rpnModifications}

For the very few-shot setting, where the number of instances $K$ for novel categories $\Cnovel$ is very low, the region proposal network (RPN) was identified as a key source for errors~\cite{CooperatingRPNs-arxiv2020}.
For example, if the detector must learn to detect a category from a single example, the detector can model the categories' variation only by proposing multiple regions of interest (RoIs) that match the object's ground truth, which is similar to random cropping augmentation, as shown in \cite{FSCE_FSOD-cvpr2021} (see \autoref{subsec:tl:increase_variance}).
If the RPN misses even one of these RoIs, the performance on this novel category may drop noticeably.
Therefore, Zhang \etal~\cite{CooperatingRPNs-arxiv2020} (CoRPN) modify the RPN, by replacing the single binary foreground classifier in the RPN with $M$ binary classifiers.
The goal is that at least one classifier identifies the relevant RoI as foreground.
Vu \etal~\cite{FORD+BL-imavis2022} (FORD+BL) added an atrous spatial pyramid pooling (ASPP) \cite{ASPP-eccv2018} context module before the RPN to increase its receptive field.
This helps in identifying relevant RoI as foreground.

As in TFA~\cite{TFA-icml2020}, in the second training phase, the weights of the RPN are frozen in many transfer learning approaches.
Fan \etal~\cite{RetentiveRCNN-cvpr2021} (Retentive R-CNN) observed that the RPN suppresses RoIs of novel categories $\Cnovel$ after it was trained only on $\Dbase$ in the first phase.
They found that unfreezing the weights of the RPN's final layer that classifies whether objects are foreground or background is sufficient to improve the RPN in the second phase.
The same conclusion was drawn by Sun \etal~\cite{FSCE_FSOD-cvpr2021} (FSCE), Wang \etal~\cite{CIR-remote-sensing-2022} (CIR), and Kaul \etal~\cite{LVC-cvpr2022} (LVC) and as a result, all RPN weights were unfrozen.
Additionally, FSCE and CIR doubled the number of proposals that pass non-maximum suppression (NMS) to get more proposals for novel categories.
FSCE compensates for this by sampling only half the number of proposals in the RoI head used for loss computation, as they observed that in the second training phase the discarded half contains only backgrounds.

\TakeAway To reduce the number of missed detections, the RPN weights should be adapted during finetuning on $\Dnovel$ and the number of proposals passing NMS can be increased.

\subsection{Modifications of the Feature Pyramid Network}
Next to unfreezing the RPN, Sun \etal~\cite{FSCE_FSOD-cvpr2021} (FSCE) showed that also finetuning the feature pyramid network (FPN) in the second phase improves the performance compared to freezing its weights.
They assume that the concepts from the base categories cannot be transferred to novel categories without any finetuning.

Wu \etal~\cite{MPSR-eecv2020} (MPSR) observed that the scales of the FPN do not compensate for the sparsity of the scales of the few samples of novel categories.
Therefore, in a refinement branch, specific data augmentation is applied to sove this issue (see \autoref{subsec:tl:increase_variance}).
Wang \etal~\cite{CIR-remote-sensing-2022} (CIR) designed a context module to enlarge the receptive field of the FPN, which also addresses the problem of varying scales and, in particular, improves the detection of small objects.

\TakeAway Also FPN weights should be adapted during finetuning \cite{FSCE_FSOD-cvpr2021}.

\subsection{Increase the Variance of Novel Categories}
\label{subsec:tl:increase_variance}

If training instances for novel categories $\Cnovel$ are limited, also the variance of the data regarding these categories is limited.
Therefore, some approaches try to increase the variance of the data for novel categories.

\begin{figure}[htb]
    \centering
    \includegraphics[width=\textwidth]{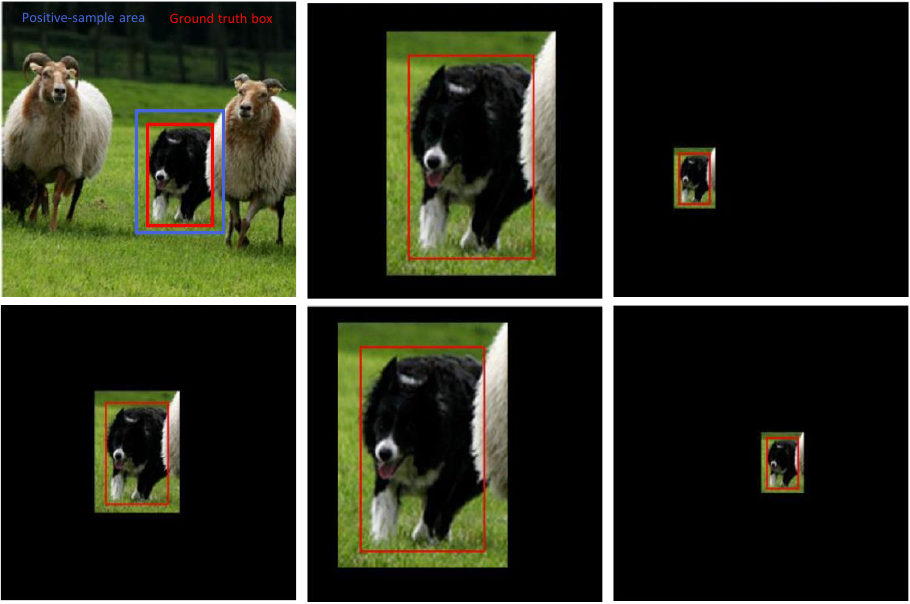}
    \caption{Augmentation for novel cateories regarding scale and translation in FSSP~\cite{FSSP-ieee2021}.}
    \label{fig:augmentation-fssp}
\end{figure}

In the refinement branch of MPSR~\cite{MPSR-eecv2020}, each object is cropped by a square window and resized to various scales.
This increases the variance regarding object sizes.
This augmentation is also employed in FSOD-UP~\cite{FSOD-UP-iccv2021} and CME~\cite{CME-cvpr2021}.
A similar approach is taken by Xu \etal~\cite{FSSP-ieee2021} (FSSP), where in an auxiliary branch the objects are augmented regarding scale and translation, as shown in~\autoref{fig:augmentation-fssp}.

Zhang \etal~\cite{Halluc.-cvpr2021} (Halluc.) introduce a hallucinator network that learns to generate additional training examples for novel categories $\Cnovel$.
To achieve this, the features in the RoI head $\featRoI$ of novel category samples are augmented by leveraging the shared within-class feature variation from base categories $\Cbase$.

Kaul \etal~\cite{LVC-cvpr2022} (LVC) show in their experiments that data augmentation, i.e., color jittering, random cropping, mosaicing, and dropout for the extracted features for each ROI, significantly improves the performance.
Sun \etal~\cite{FSCE_FSOD-cvpr2021} (FSCE) describe the similarity between the augmentation with several random image crops and multiple RoI proposals from the RPN.
Thus, increasing the number of proposed RoIs per novel category instance as described in \autoref{subsec:rpnModifications} is also increasing the variance of novel categories as it resembles random cropping augmentation.
According to \cite{Halluc.-cvpr2021}, increasing the variance of novel categories primarily benefits the extreme few-shot scenario with very few samples $K$ per novel category.

If additional unlabeled data containing novel categories are available, techniques from semi-supervised learning can be applied to increase the number of samples for novel categories.
Liu \etal~\cite{N-PME-icassp2022} (N-PME) pseudo-label the base dataset $\Dbase$ after finetuning in order to find additional samples of $\Cnovel$ in $\Dbase$.
The novel samples are then used for an additional finetuning phase with more shots by including the samples pseudo-labeled as on of the categories in $\Cnovel$.
Since the bounding boxes for the additional samples are rather imprecise, they are omitted for the regression loss.
Kaul \etal~\cite{LVC-cvpr2022} (LVC) further improve this attempt by first verifying that the searched novel samples indeed belong to $\Cnovel$ and then correct the inaccurate bounding boxes.
For verification, they apply a vision transformer (ViT) \cite{ViT-iclr2021}, which was trained in a self-supervised manner by DINO \cite{DINO-iccv2021}, to get features that can be used in a k-nearest-neighbor classifier to compare to the $K$ shots of the novel categories.
If novel samples can be verified, they are included in $\Dnovel$, otherwise, these regions are ignored during the following additional finetuning on the extended data.
Bounding boxes for verified samples are corrected in the fashion of Cascade R-CNN \cite{Cascade-R-CNN-cvpr-2018}.
Using the high-quality extended data, the detector can be finetuned end-to-end without the need for freezing any components of the detector.
While this seems to improve the performance significantly, it should be noted that for the FSOD benchmark datasets Microsoft COCO and PASCAL VOC, searching for novel objects in images of $\Dbase$ is sufficient, but that for real-world few-shot applications, such as medical applications or the detection of rare species, additional (unlabeled) data is needed, which could prove problematic.

\TakeAway Increasing the variance of training examples from novel categories~-- e.g., by data augmentation or pseudo-labeling additional data~-- improves detection accuracy, especially when the number of training examples $K$ is very low.

\subsection{Transfer Knowledge Between Base and Novel Categories}
In LSTD~\cite{LSTD-aaai2018}, using a soft assignment of similar base categories weights of components for novel categories are initialized by base category weights to transfer base knowledge.
Chen \etal~\cite{BottomUpTopDownAttention-arxiv2020} (AttFDNet) initialize the parameters of the novel object detector using parameters from the base object detector and an imprinting initialization method \cite{imprinting-cvpr2018, imprinting-tip2020}.
Also Li \etal~\cite{CGDP+FSCN-cvpr2021} (CGDP+FSCN) use imprinting for initialization \cite{imprinting-cvpr2018}.

By learning and leveraging visual and semantic lingual similarities between the novel and base categories, in the second training phase, Khandelwal \etal~\cite{UniT-cvpr2021} (UniT) transfer weights for bounding box regression and classification from base categories to novel categories.
Zhu \etal~\cite{SRR-FSD-cvpr2021} (SRR-FSD) represent each category concept by a semantic word embedding learned from a large corpus of text.
The image representations of objects are projected into this embedding space to learn $\Cnovel$ from both the visual information and the semantic relation.
Cao~\etal~\cite{FADI-neurips2021} (FADI) also incorporate the categories' semantic meaning:
After training on $\Dbase$, they measure the semantic similarity of base and novel categories via WordNet~\cite{WordNet-acm1995}.
The authors argue, that in the second finetuning phase, associating novel categories to multiple base categories leads to scattered intra-class structures for the novel categories.
Thus, each novel category is associated to exactly one base category with the highest similarity.
Afterwards, each novel category is assigned a pseudo label of the associated base category.
Then, the whole network is frozen -- except for the second fully connected layer in RoI Head -- and the network is trained such that it learns to align the feature distribution of the novel category to the associated base category.
This leads to low intra-class variation of the novel category but inevitably to confusion between $\Cbase$ and $\Cnovel$.
Thus, in a subsequent discrimination step, the classification branches for $\Cbase$ and $\Cnovel$ are disentangled to learn a good discrimination.
In \cite{KR-FSOD-electronics2022} (KR-FSOD), a semantic knowledge graph based on word embeddings is used to describe a scene and relations between objects.
This helps to improve knowledge propagation between novel and related categories.

\TakeAway For initializing the weights of components for each novel category, knowledge from the semantically most similar base category should be transferred \cite{FADI-neurips2021}.

\subsection{Keep the Performance on Base Categories}
\label{sec:tl:base_categories}
Many approaches suffer from catastrophic forgetting when trained on $\Cnovel$.
Although the model can be trained on $\Cbase$ as well in the finetuning phase, the performance still drops compared to before finetuning.
Therefore, Fan \etal~\cite{RetentiveRCNN-cvpr2021} (Retentive R-CNN) propose to duplicate the RPN and the classification heads for RoI proposal and classification of $\Cbase$ and $\Cnovel$ separately.
During finetuning of the $\Cnovel$ head, a cosine classifier is used to balance the variations in feature norms of $\Cbase$ and $\Cnovel$.
The frozen RPN and RoI head for $\Cbase$ shall keep the performance on the base categories.
Feng \etal~\cite{BPMCH-prl2022} (BPMCH) combat catastrophic forgetting for base categories $\Cbase$ during the finetuning phase mainly by fixing the backbone $\mathcal{B}_{base}$ and the classification head for these categories, and use an additional backbone $\mathcal{B}_{novel}$ as a feature extractor for novel categories $\Cnovel$.

In MemFRCN~\cite{MemFRCN-tfeccs2022}, additional to the softmax-based classifier in the RoI head, epresentative feature vectors $\mathrm{f}^{\RoI, c_i}$ for each category $c_i$ are learned and stored to remember the base categories $\Cbase$ after the RoI head is modified during the finetuning phase.
During inference, extracted features $\featRoI$ can be compared to these category representatives by cosine similarity.
This is similar to support vectors in dual-branch meta learning.

Guirguis \etal~\cite{CFA-cvpr2022} (CFA) build on the continual learning approaches GEM \cite{GEM-neurips2017} and A-GEM \cite{AGEM-iclr2019}, which observed that catastrophic forgetting occurs when the angle between loss gradient vectors of previous tasks and the gradient update of the current task is obtuse.
Therefore, CFA stores K shots of the base categories in an episodic memory, analogous to A-GEM, in order to be able to compute gradients on $\Dbase$.
During the finetuning phase, the episodic memory is static, meaning that no further samples are added.
The finetuning is then conducted as follows:
The base category gradient $g_{base}$ is calculated on a mini-batch drawn from the episodic memory and the novel category gradient $g_{novel}$ is calculated on a mini-batch from $\Dnovel$.
If the angle between $g_{base}$ and $g_{novel}$ is acute, $g_{novel}$ is back-propagated as it is.
Otherwise, a new gradient update rule is derived, which averages the base gradients $g_{base}$ and novel gradients $g_{novel}$.
It also adaptively reweights them in case the novel gradients $g_{novel}$ point towards a direction that could lead to forgetting.

\TakeAway To prevent catastrophic forgetting and keep the performance on base categories, the angle between gradients of novel and base categories must be taken into account \cite{CFA-cvpr2022}.

\subsection{Modify the Training Objective}
\label{subsec:tl_loss}

A modified loss, which updates the training objective, can guide the detector towards focusing on foreground regions or specific aspects, may improve the consistency in multiple branches, and may also help to improve the inner- and inter-class variance of features for object classification.
Furthermore, restricting the gradient flow in the detector or slightly modifying the training scheme can improve the training of the different components of the detector.

\subsubsectionspace{Additional Loss Terms}
Chen \etal~\cite{LSTD-aaai2018} (LSTD) use additional background-depression and transfer-knowledge regularization terms in the loss function to help the detector focusing on target objects and incorporating source-domain knowledge.
Li \etal~\cite{CGDP+FSCN-cvpr2021} (CGDP+FSCN) identified unlabeled instances of novel categories $\Cnovel$ in the base dataset $\Dbase$ as problematic.
They introduce an additional semi-supervised loss term to also utilize these unlabeled instances.
Chen \etal~\cite{BottomUpTopDownAttention-arxiv2020} (AttFDNet) propose two loss terms to maximizes the cosine similarity between instances of the same category and to tackle the problem of unlabeled instances in the dataset, respectively.
Cao \etal~\cite{FADI-neurips2021} (FADI) introduce an additional set-specialized margin loss to enlarge inter-class separability.
In contrast to previous margin losses such as ArcFace~\cite{ArcFace-cvpr2019}, they use scaling factors for different margins, where the scaling factor for $\Cnovel$ is higher than for $\Dbase$, as novel categories are much more challenging.
Liu \etal~\cite{N-PME-icassp2022} (N-PME) use a margin loss to exploit error-prone pseudo-labels by evaluating the uncertainty scores of both correct and incorrect pseudo-labels for novel categories on additional data.

\subsubsectionspace{Loss for Auxiliary Branches}
Similar to the meta-learning approach TIP~\cite{TIP-cvpr2021}, Wu \etal~\cite{FSOD-UP-iccv2021} (FSOD-UP) use a consistency loss to force features of two branches to be similar.
They apply the KL-Divergence loss between these features.
The context module of CIR~\cite{CIR-remote-sensing-2022} is trained in a supervised manner by an auxiliary classification branch that predicts a binary foreground-background segmentation map.
The two branches of MPSR~\cite{MPSR-eecv2020} are loosely coupled via shared weights and contributions of both branches to the loss function.
CME~\cite{CME-cvpr2021} builds on top of MPSR, but introduces an additional adversarial training as we described in~\autoref{sec:dual-branch-meta-discriminative}.
Also Xu \etal~\cite{FSSP-ieee2021} (FSSP) introduce an auxiliary branch.
It includes a full replication of the detection network used for data augmentation.
A modified classification loss combines the decisions in the original branch and this auxiliary branch that processes only one object with most of the background removed.

Sun \etal~\cite{FSCE_FSOD-cvpr2021} (FSCE) introduce a new branch in the RoI head.
In addition to the standard RoI head they apply a single fully connected layer as contrastive branch to be able to measure similarity scores between learned object proposal representations.
On the contrastive branch, they use a contrastive proposal encoding loss for training that enables increasing the cosine similarity of representation from the same category and reduce the similarity of proposals from different categories.
Lu \etal~\cite{DMNet-tcyb2022} (DMNet) follow a similar approach.
They use an auxiliary classification branch in which they compare extracted features to representatives for each category by euclidean distance.
The feature embedding and the category representatives are learned by triplet-loss based metric learning.

\subsubsectionspace{Modified Gradient Flow}
Qiao \etal~\cite{DeFRCN-iccv2021} (DeFRCN) additionally want to update the backbone in both training phases, but they identified contradictions in training as problematic.
The goals of RPN and ROI head are contrary since the RPN tries to learn class-agnostic region proposals whereas the ROI head tries to distinguish categories.
Their extensive experiments showed that it is key to stop the gradient flow from the RPN to the backbone and scale the gradient from the ROI head to the backbone.
During training on $\Dbase$ in the first phase, they scale the gradient from the ROI head by $0.75$, so that the backbone learns a little less than the rest of the detector.
During training on $\Dbase \cup \Dnovel$ in the second phase, it proved necessary to scale the gradient by $0.01$, which is in the direction of freezing the backbone.
Stopping the gradient from the RPN and scaling the gradient from the ROI head significantly boosts the performance, especially in the second phase.
The authors observed, that this gradient scaling also benefits Faster R-CNN as generic object detector, when trained with sufficient data as well.

Guirguis \etal~\cite{CFA-cvpr2022} (CFA) derived a new gradient update rule that takes the angle between gradients for samples of $\Dbase$ and samples of $\Dnovel$ into account in order to combat catastrophic forgetting for base categories $\Cbase$ during the finetuning phase, as already described in \autoref{sec:tl:base_categories}.
While this gradient update rule primarily intends to preserve the performance on base categories $\Cbase$, it also has a positive impact on the performance regarding novel categories $\Cnovel$.

\subsubsectionspace{Modified Training Scheme}
Inspired by infants beginning to learn from a single observation, in \cite{FORD+BL-imavis2022} (FORD+BL) it is shown that instead of finetuning with $K$ shots immediately, the performance can be improved by first finetuning with a single shot per category and only then finetuning with all $K$ shots.
Wu \etal~\cite{TD-Sampler-icccbda2022} (TD-Sampler) introduced a batch sampling strategy for the finetuning phase that enables to use all samples of $\Dbase$ instead of $K$ shots per base category $\Cbase$ and to use more samples of novel categories $\Cnovel$ per training batch.
This is achieved by selecting batches that contain a large number of novel category samples and are unlikely to significantly change the detector activation pattern, judged by an estimated training difficulty (TD).

\TakeAway The loss should be modified regarding optimized gradient flow and inter-class separability.
In an auxiliary branch, a contrastive loss can help to improve the discriminative power of features, like in two-branch meta learning.

\subsection{Use Attention}

Attention blocks help to enhance features.
In this sense, Wu \etal~\cite{FSOD-UP-iccv2021} (FSOD-UP) use soft attention between learned prototypes (see \autoref{subsec:tl:architecture}) and RPN outputs to enhance features in an extra branch.
Yang \etal~\cite{ContextTransformer-aaai2020} (CoTrans) use the affinity between an anchor box and its contextual field as a relational attention to integrate contexts into the representation of the anchor box.
Xu \etal~\cite{FSSP-ieee2021} (FSSP) first process the image by a self-attention module and then process the attention-enriched input by a one-stage detector.
Therefore, the detector can focus on important parts of the input image.
Chen \etal~\cite{BottomUpTopDownAttention-arxiv2020} (AttFDNet) combine top-down and bottom-up attention.
Top-down attention is learned in supervised fashion in a simplified non-local block and a squeeze-and-excitation block.
Bottom-up attention is computed by a saliency prediction model (BMS \cite{saliency-bms} or SAM \cite{saliency-sam}).

\subsection{Modify Architecture}
\label{subsec:tl:architecture}

\subsubsectionspace{Architectures based on Faster R-CNN}
The majority of transfer learning approaches are based on the Faster R-CNN detector as shown in \autoref{fig:transfer-learning}.
Benchmark results confirm the superiority of this two-stage detector for transfer learning approaches.
Only few approaches deviate from this architecture.

Li \etal~\cite{CGDP+FSCN-cvpr2021} (CGDP+FSCN) observed that the performance degradation for novel categories in Faster R-CNN is mainly caused by false positive classifications, i.e., by category confusion.
Therefore, they refine the classification in an additional discriminability enhancement branch, which is trained with misclassified false positive samples.
It directly processes the cropped image of the object to be classified.
Then, the classification result is fused with the one of the original Faster R-CNN branch.
Qiao \etal~\cite{DeFRCN-iccv2021} (DeFRCN) also observed many low classification scores for novel categories.
As Li \etal, they conclude that contrary requirements of translation invariant features for classification and translation covariant features for localization are problematic. 
To tackle this issue, they propose a prototypical calibration block, which performs score refinement to eliminate high-scored false positive classifications.

Wu \etal~\cite{MPSR-eecv2020} (MPSR) use an auxiliary refinement branch for data augmentation during training that is excluded during inference.
SVD~\cite{SVD-Dictionary-neurips2021} builds upon MPSR~\cite{MPSR-eecv2020}.
With a singular value decomposition (SVD), they decompose the backbone features $\featBackbone$ into eigenvectors with their relevance quantified by the corresponding singular values.
The eigenvectors corresponding to the largest singular values are incorporated for localization, since they are able to suppress certain variations.
In contrast, the eigenvectors corresponding to the smaller singular values are incorporated for category discrimination since they encode category-related information.
This discrimination space is further refined by utilizing dictionary learning~\cite{dictionary-learning-cvpr2017} to facilitate classification.
Wu \etal~\cite{FSOD-UP-iccv2021} (FSOD-UP) adapt the few-shot learning idea of prototypes \cite{PrototypicalNetworks-NeurIPS2017, prototype-iccv2019, prototype-eccv2020} that reflect category information.
In contrast to category-specific prototypes in dual-branch meta learning, they learn universal prototypes based on all categories in an extra branch that processes backbone features.
These universal prototypes are invariant under different visual changes and, thus, enhance the original features from the backbone.
After processing original and enhanced features in the RPN, this processing in an auxiliary branch is repeated for RPN features to compute the input for the ROI head.

\subsubsectionspace{Incorporating One-Stage Detectors}
One of the earliest FSOD approaches, LSDT~\cite{LSTD-aaai2018}, combines bounding box regression following the SSD~\cite{SSD-eccv2016} approach and Faster R-CNN~\cite{FasterRCNN-neurips2015} concepts for object classification.

Yang \etal~\cite{ContextTransformer-aaai2020} (CoTrans) use SSD~\cite{SSD-eccv2016} as a one-stage detector.
They argue that the multi-scale spatial receptive fields in this architecture provide rich contexts, which are important for knowledge transfer.
Chen \etal~\cite{BottomUpTopDownAttention-arxiv2020} (AttFDNet) also use the SSD detector, but add two attention branches, to help the detector to focus on the important parts of the image, and six prediction heads that predict bounding boxes and categories for objects at different scales.

Lu \etal~\cite{DMNet-tcyb2022} (DMNet) proposed a one-stage detector that follows the design principles of SSD and YOLO, but uses two decoupled branches for localization and classification.
It is argued that this decoupling facilitates adaptation with only a few examples.

Xu \etal~\cite{FSSP-ieee2021} (FSSP) show, how a  fast one-stage detector, namely YOLOv3~\cite{YOLOv3-arxiv2018}, can be made competitive with the slower two-stage detector Faster R-CNN in the vanilla setup as described in~\cite{TFA-icml2020}.
This is possible only by putting in a lot of effort, namely incorporating a self-attention module, using an additional auxiliary branch that contains a full replication of the detection network, augmenting the input data of the auxiliary branch, and applying an additional loss.
However, due to these modifications, the fast one-stage detector of Xu \etal~\cite{FSSP-ieee2021} is especially performing better than TFA~\cite{TFA-icml2020} for the extremely low-shot scenario.

\TakeAway In detectors based on Faster R-CNN, a score refinement can help to reduce false positive classifications \cite{DeFRCN-iccv2021}.
Single-stage detectors can profit from an auxiliary branch in order to enable data augmentation \cite{FSSP-ieee2021}.

\subsection*{Summary of Best Performing Transfer Learning Approaches}
\label{sec:transfer-best-approaches}

In the following, we summarize selected transfer learning approaches with distinct concepts regarding the categories described above that perform best on FSOD benchmark datasets (see \autoref{sec:experiments}).

In \textbf{DeFRCN} \cite{DeFRCN-iccv2021} it was found that the class-agnostic localization task in the RPN and the class-distinguishing task of the ROI head are contrary.
Therefore, it is key to stop the gradient flow from the RPN to the backbone and scale the gradient from the ROI head to the backbone.
Then, it is possible to train all components of the detector, including the backbone, in both training phases, which significantly boosts the performance, especially in the finetuning phase.
Additionally, a prototypical calibration block performs score refinement in the ROI head to eliminate high-scored false positive classifications, which are a result of contrary requirements of translation invariant features for classification and translation covariant features for localization.

\textbf{CFA} \cite{CFA-cvpr2022} can be applied on top of DeFRCN to tackle catastrophic forgetting of base categories during finetuning, which occurs when angles between base category gradients and novel category gradients are obtuse.
Therefore, CFA stores K shots of the base categories in an episodic memory for computing gradients on $\Dbase$ during finetuning.
If the angle between base and novel category gradients is obtuse, both gradients are averaged and adaptively reweighted, otherwise the novel category gradient can be back-propagated without the risk for catastrophic forgetting.
This gradient update rule also benefits the performance regarding novel categories.

Before finetuning, \textbf{FADI} \cite{FADI-neurips2021} associates each novel category to exactly one base category by measuring their semantic similarity via WordNet.
Then, the network is trained to align the feature distribution of the novel category to the associated base category.
This leads to low intra-class variation of the novel category but inevitably to confusion between $\Cbase$ and $\Cnovel$.
Thus, in a subsequent discrimination step, the classification branches for $\Cbase$ and $\Cnovel$ are disentangled to learn a good discrimination.
Additionally, a set-specialized margin loss is employed to enlarge inter-class separability.

If additional unlabeled data containing novel categories are available, techniques from semi-supervised learning can be applied.
On FSOD benchmark datasets $\Dbase$ can be used for this purpose.
\textbf{LVC} \cite{LVC-cvpr2022} pseudo-labels this data with the detector that was finetuned on $\Dnovel$ in the second training phase.
First, the searched novel samples are verified to belong to $\Cnovel$ by using features of a vision transformer to compare with the shots of $\Dnovel$ in a nearest-neighbor fashion.
Then, the inaccurate bounding boxes of verified samples are corrected similar to Cascade R-CNN.
Using the high-quality extended data for novel categories, all components of the detector can be trained in an additional finetuning phase.

\subsection*{Conclusion on Transfer Learning Approaches}
Transfer learning approaches have a much simpler training pipeline, as they do not require complex episodic training as in meta learning.
By incorporating specific techniques to be able to finetune as much components of the detector as possible --~e.g., modifying the training objective or transferring knowledge between base and novel categories~-- transfer learning approaches are able to reach state of the art performance.

\section{Comparison between Meta Learning and Transfer Learning}
\label{sec:comparison}

After elaborating on different approaches for meta learning as well as transfer learning, we now want to draw a comparison.
Since single-branch meta learning is less explored in recent works and also falls behind in terms of performance, we discard it in the comparison.
In \autoref{tab:comparison-meta-transfer}, we compare dual-branch meta learning and transfer learning according to several important aspects.
Both seem promising for future work and either could benefit by also incorporating ideas from the other training scheme.

\begin{table}[ht]
    \footnotesize
    \resizebox{\textwidth}{!}{%
\begin{tabular}{p{0.48\linewidth}p{0.57\linewidth}}
\toprule
\textbf{\dualbranch~Dual-Branch Meta Learning} & \textbf{\transfer~Transfer Learning} \\ 
\midrule
\multicolumn{2}{c}{\textsc{Main Focus}} \\[0.5mm]
How to aggregate information of the support and query branch &
How to freeze less components of the detector without performance decline \\ 
\midrule
\rowcolor[HTML]{EFEFEF}
\multicolumn{2}{c}{\textsc{Training Scheme}} \\[0.5mm]
\rowcolor[HTML]{EFEFEF}
Complex episodic training scheme & 
Simple training protocol \\ 
\midrule
\multicolumn{2}{c}{\textsc{Adaptation to $\Cnovel$}} \\[0.5mm]
Fast adaptation to $\Cnovel$ (some approaches can even be applied to $\Cnovel$ without finetuning, e.g., SQMG \cite{SQMG-cpvr2021}) & 
Requires finetuning on $\Cnovel$ \\ 
\midrule
\rowcolor[HTML]{EFEFEF}
\multicolumn{2}{c}{\textsc{Combat Too Few Proposals for $\Cnovel$}} \\
\bottomrule
    \end{tabular}
    }
    \caption{
    Comparison between dual-branch meta learning and transfer learning
    }%
    \label{tab:comparison-meta-transfer}
    \vspace{-2mm}
\end{table}
\section{Datasets, Evaluation Protocols, and Benchmark Results}
\label{sec:experiments}

\begin{table*}[!b]
    \centering
    \scriptsize
    \setlength\tabcolsep{3pt} %
    \resizebox{\textwidth}{!}{%
\begin{tabular}{@{}clrlcccccccccccccccccc@{}}
\toprule
\multicolumn{1}{l}{\textbf{}}                                                              & \textbf{}                                                                                      &                                                          & \multicolumn{1}{l|}{\textbf{}}                                                        & \multicolumn{1}{c|}{\textbf{Novel}}                                      & \multicolumn{5}{c|}{\textbf{Novel Set 1}}                                                                                                                                                                                                                                                       & \multicolumn{5}{c|}{\textbf{Novel Set 2}}                                                                                                                                                                                                                                                       & \multicolumn{5}{c|}{\textbf{Novel Set 3}}                                                                                                                                                                                                                                                       & \multicolumn{2}{c}{\textbf{Base Set 1}}                                                                          \\
\bottomrule
\end{tabular}%
}
    \caption{%
    $AP_{50}$ published results on the PASCAL VOC benchmark for all three sets and different number of shots $K$.
    We sort the approaches by the mean over all novel sets and shots.
    ---:~no result reported in paper.
    $\available$: results only reported for different shots or sets, therefore these results are not included here.
    *:~Deviating evaluation protocol preventing fair comparison as described in \autoref{sec:experiments-deviating-evaluation}.
    \dualbranch:~dual-branch meta learning. \singlebranch:~single-branch meta learning. \transfer:~transfer learning.
    }%
    \label{tab:voc}
\end{table*}

Evaluation of few-shot object detectors requires tailored datasets that distinguish between base and novel categories.
Therefore, most approaches use specific splits of the common object detection datasets PASCAL VOC~\cite{VOC} and Microsoft COCO~\cite{COCO}.
Only rarely, other datasets, like FSOD, ImageNet-LOC, or LVIS, are applied.
Further explanations on these datasets can therefore be found in the appendix in \autoref{sec:appendix-further-results}.
Generally, few-shot object detectors are evaluated in the $K$-shot-$N$-way manner, i.e., $\Dnovel$ consists of $K$ labeled examples for $N$ novel categories.

\vspace{-1mm}
\subsection{PASCAL VOC Dataset}
The PASCAL VOC dataset~\cite{VOC} contains annotations for 20 categories.
Commonly, the combination of VOC\,07+12 trainval sets is used for training, and VOC\,07 test set for testing.
For evaluating few-shot object detectors, most often three category splits are used; each with 15 base categories and five novel categories ($N=5$):
\begin{itemize}
    \item Set 1: $C_{novel} = \{\textit{bird, bus, cow, motorbike, sofa}\}$
    \item Set 2: $C_{novel} = \{\textit{aeroplane, bottle, cow, horse, sofa}\}$
    \item Set 3: $C_{novel} = \{\textit{boat, cat, motorbike, sheep, sofa}\}$
\end{itemize}
The number of shots $K$ for novel categories is set to $1$, $2$, $3$, $5$, and $10$.
As evaluation metric, the mean average precision at an intersection over union (IoU) threshold of 0.5 is used ($AP_{50}$).
Unfortunately, the specific $K$-shot object instances are not fixed, which leads to varying instances used in different approaches.
As stated by Wang~\etal~\cite{TFA-icml2020}, this high variance in training samples makes it difficult to compare approaches against each other, as approach-based performance differences may be insignificant compared to differences based on different instances.
Therefore, Wang \etal~\cite{TFA-icml2020} propose a revised evaluation protocol, where results are averaged over 30 runs with different random samples of training shots.
Moreover, they also report the performance on base categories since ignoring the performance for base categories might hide a potential performance drop and is, therefore, not suitable for evaluating the overall performance of a model.
Currently, approaches focusing on this topic also report results for generalized FSOD (G-FSOD) performance, which refers to the mean over novel and base categories.

\vspace{4mm}
In~\autoref{tab:voc}, we list benchmark results of the described approaches for the PASCAL VOC dataset.
We split the table according to whether the results are given for a singe run or averaged over multiple runs as proposed by Wang~\etal~\cite{TFA-icml2020}.
Approaches that give results for both evaluation protocols are marked with "$\longleftarrow$" to act as anchors for comparison.
The performance gap between the two evaluation protocols is not negligible and shows that the averaged results are more reliable.
Furthermore, we also denote whether finetuning on $\Dnovel$ is required or if the results are achieved by simply meta-testing.
Each approach is characterized by a small symbol at the front.
In general, both transfer learning and dual-branch meta learning approaches can achieve similar results.
The main characteristics of the best performing approaches are summarized at the end of \autoref{sec:meta-learning-dual-branch} for dual-branch meta learning and \autoref{sec:transfer-learning} for transfer learning.

Although it is very commonly used, according to Michaelis~\etal~\cite{ClosingGeneralizationGap-arxiv2020}, the PASCAL VOC dataset is too easy:
With a dual-branch meta learning approach and uninformative all-black support images, they are still able to locate the novel objects and reach a $mAP_{50}$ of 33.2.
However, we want to highlight that in general objects do not simply need to be \emph{located} but also \emph{classified}, i.e., detectors need to also determine which category is present in the image.

\subsection{Microsoft COCO Dataset}
In comparison to PASCAL VOC, the Microsoft COCO dataset~\cite{COCO} is more challenging and contains annotations for 80 categories, including the 20 VOC categories.
For few-shot object detection, most often the 20 VOC categories are used as novel categories, leaving the remaining 60 categories as base categories.
Typically, the number of shots $K$ is set to 10 and 30.
However, some approaches focus on the extremely low-shot regime and use only 1--3 shots per category.

For evaluation, the standard COCO metrics are used:
The primary metric is $AP_{50:95}$: the mean of 10 average precision values with IoU thresholds in range $[0.5, 0.95]$.
Moreover, $AP_{50}$ is reported, which corresponds to the Pascal VOC metric.
$AP_{75}$ is more strict, as detections only count as positive when their IoU with a ground truth object is larger than 0.75.
Some approaches also report average recall $AR_1$, $AR_{100}$ and $AR_{1000}$ for 1, 100 or 1000 detections per image, respectively.
Average precision and average recall are also available for small, medium-sized and large objects ($AP_S$, $AP_M$, $AP_L$, $AR_S$, $AR_M$, $AR_L$).

We list published benchmark results for $K=10$ in \autoref{tab:coco-10-shot}.
Since the relative order for $K=30$ is similar, the corresponding table can be found in the appendix in \autoref{tab:coco-30-shot}.
Results for the extreme 1-shot scenario are shown in the appendix in~\autoref{tab:coco-1-shot}.
Similar to PASCAL VOC, the evaluation of few-shot detectors on Microsoft COCO also suffers from varying $K$-shot instances between approaches.
Therefore, we again split the tables according to whether the revised evaluation protocol proposed by Wang \etal~\cite{TFA-icml2020} was used, where results are averaged over 10 runs with different random samples.

\begin{table*}[!b]
    \centering
    \scriptsize
    \setlength\tabcolsep{3pt} %
    \resizebox{\textwidth}{!}{%
\begin{tabular}{@{}clrl|cccccccccccc|cc@{}}
\toprule
\multicolumn{1}{l}{}                                                                &                                                                                                & \textbf{}                                                & \multicolumn{1}{r|}{}                                            & \multicolumn{12}{c|}{\textbf{Novel Categories}}                                                                                                                                                                                                                                                                                                                                                                                                                                                                                                                                                                                                                                   & \multicolumn{2}{c}{\textbf{Base   Cat.}}                                                                \\
\bottomrule
\end{tabular}%
}
    \caption{
    Benchmark results for 10-shot on the Microsoft COCO dataset sorted by novel $AP_{50:95}$. 
    ---: no result reported in paper. 
     *:~Deviating evaluation protocol preventing fair comparison as described in \autoref{sec:experiments-deviating-evaluation}.
    \dualbranch:~dual-branch meta learning. \singlebranch:~single-branch meta learning.~\transfer: transfer learning.
    }
    \label{tab:coco-10-shot}
\end{table*}

\vspace{2mm}
\subsection{Deviating Evaluation Protocols}
\label{sec:experiments-deviating-evaluation}
In general, object detection is the joint task of localizing and classifying object instances.
However, some few-shot approaches deviate from this setting and create a simpler task:
For example, some approaches that explicitly focus on the one-shot scenario (CoAE~\cite{CoAE-neurips2019}, AIT~\cite{AIT-cvpr2021}, OSWF~\cite{OneShotWithoutFinetuning-arxiv2020} CGG~\cite{ClosingGeneralizationGap-arxiv2020}) assume each query image $\ImQ$ to have at least one instance of the object category $c$ from the support image $\ImSc$.
Implicitly, this removes the classification task and only requires localization.
The same applies to the 1-way training and evaluation setting of
DAnA~\cite{DAnA-tmm2021}, where the detector only needs to estimate whether the current query image $\ImQ$ contains the object and where it is located, but the difficulty of correct classification is eliminated.
In order to report comparable results, we therefore strongly recommend to always evaluate with the $N$-way setting.

In contrast, PNSD~\cite{PNSD-accv2020} and KFSOD~\cite{KFSOD-cvpr2022} simplify the generalization to novel categories by already utilizing a ResNet pretrained on COCO, such that the novel categories are not really novel anymore.

\subsection{Problems of Common Evaluation Protocols}

\subsubsectionspace{High Variance for Different Samples}
As pointed out by Wang~\etal~\cite{TFA-icml2020}, the use of different samples for novel categories can lead to a high variance in performance and, therefore, makes comparison difficult.
Hence, we highly recommend to always report the average of the results over multiple random runs.

\vspace*{4mm}
\subsubsection{ImageNet Pretraining and Choice of Novel Categories}
\phantom{.}\\[-3mm]
Most approaches use an ImageNet-pretrained backbone.
While this is common for generic object detection, it has a negative side-effect for FSOD:
The novel categories are not truly novel anymore, as the model has probably already seen images of this category.
However, omitting ImageNet pretraining altogether results in worse performance even for the base categories.
To alleviate this problem, there are two options:

First, the ImageNet categories which correspond to the novel categories can be excluded from ImageNet-pretraining as done in CoAE~\cite{CoAE-neurips2019}, SRR-FSD~\cite{SRR-FSD-cvpr2021}, and AIT~\cite{AIT-cvpr2021}.
CoAE~\cite{CoAE-neurips2019} and AIT~\cite{AIT-cvpr2021} even remove all COCO-related categories from ImageNet, which results in 275 categories beeing removed (see \autoref{tab:voc-imagenet} in the appendix for benchmark results).
However, as Zhu \etal~\cite{SRR-FSD-cvpr2021} (SRR-FSD) argue, removing all COCO-related categories is not realistic, as these categories are very common in the natural world and removing 275 categories may affect the representational power learned through pretraining.
Therefore, Zhu \etal~\cite{SRR-FSD-cvpr2021} only remove categories corresponding to novel categories for PASCAL VOC, resulting in 50 categories being removed on average.
Yet, this requires an additional pretraining for every different set of novel categories.

The second option for preventing foreseeing novel categories is using a dataset with novel categories that do not occur in ImageNet.
Such a dataset would also be more realistic.
Using categories such as cats as novel categories is absurd as there is loads of annotated data.
Therefore, a more realistic approach would be to select novel categories that are indeed rare.
For example, the LVIS~\cite{LVIS-cvpr2019} dataset provides a natural long tail with more and less frequent categories.
A maximum of 10 training instances are available for rare categories. Therefore, they can be used as novel categories.
However, Huang~\etal~\cite{FSOD-self-supervised-survey-arxiv2021} point out that some rare categories in the training set do not appear in the validation set at all, which hinders the performance evaluation and requires further refinement of balanced splits and evaluation sets.
\clearpage
\section{Current trends}

\vspace*{1.5mm}%
\hspace*{-3.5mm}\textbf{Improvement of Techniques}:
Currently, dual-branch meta learning approaches improve much by using attention for aggregating features of both branches \cite{APSP-wacv2022, IFC-apin2022, Meta-DETR-tpami2022}.
By aggregating before the RPN \cite{AttentionRPN-cvpr2020} or using a proposal-free transformer as detector \cite{Meta-DETR-tpami2022}, the issue of missing proposals for novel categories is effectively solved.
Transfer learning approaches currently improve much by guiding the gradient flow to be able to train as much components of the detector as possible \cite{DeFRCN-iccv2021}.
In both types of approaches a current trend is the use of metric learning concepts by modifying the loss function to enable better category separation \cite{IFC-apin2022}.
More trends are highlighted as part of the comparison in \autoref{sec:comparison}.

\vspace*{1.5mm}%
\hspace*{-3.5mm}\textbf{Extension to Related Research Areas}:
Besides these trends towards improving FSOD techniques, the extension of FSOD concepts to further research areas such as a weakly-supervised setting \cite{StarNet-aaai2021}, self-supervised learning \cite{FSOD-self-supervised-survey-arxiv2021}, or to few-shot instance segmentation \cite{FGN-few-shot-instance-cvpr2020, FAPIS-few-shot-instance-cvpr2021} is also a current trend.

\vspace*{1.5mm}%
\hspace*{-3.5mm}\textbf{Open Challenges}:
Many approaches focus on either improving meta learning or transfer learning, but often neglect that concepts between both types of approaches are exchangeable as pointed out in \autoref{tab:comparison-meta-transfer}, which leaves potential for improvements in future work.
Since the mainly used FSOD benchmarks PASCAL VOC and Microsoft COCO do not contain realistic novel categories that represent rare objects, we would like to encourage future research to additionally evaluate on more realistic datasets such as LVIS or FSOD, as already done in \cite{AttentionRPN-cvpr2020, MM-FSOD-arxiv2020, TFA-icml2020, LearningSegmentTail-cvpr2020}.
Additionally, in \cite{FSOD-toolchain-mmm2022} a framework for creating a customized FSOD dataset is provided.
When employing FSOD in a realistic setting, including really rare categories, likely, a domain shift will occur.
Therefore, concepts from cross-domain detection \cite{cai2019exploring, xu2020cross, chen2020harmonizing} should be further explored in future work.

\section{Conclusion}
In this survey, we provided a comprehensive overview of the state of the art for few-shot object detection.
We categorized the approaches according to their training scheme and architectural layout into single-branch and dual-branch meta learning and transfer learning.
Meta learning approaches use episodic training to improve the subsequent learning with few object instances per novel category.
Dual-branch meta leaning approaches utilize a separate support branch receiving the image of a designated object, to learn how to represent the objects' category and where to attend in the query image.
Transfer learning approaches use a more simplified training scheme, by simply finetuning on the novel categories.

After introducing the main concepts, we elaborated on how specific approaches differ from the general realization and gave short takeaways in order to highlight key insights for well performing methods.
Based on an analysis of benchmark results on the most widely used datasets PASCAL VOC and Microsoft COCO, we identified current trends in the best performing dual-branch meta learning and transfer learning approaches.
It remains an open question which of these two concepts will prevail.

\ifCLASSOPTIONcaptionsoff
  \newpage
\fi

\bibliographystyle{IEEEtran}
\bibliography{Few_Shot_Object_Detection.bib}

\begin{thebibliography}{100}
\providecommand{\url}[1]{#1}
\csname url@samestyle\endcsname
\providecommand{\newblock}{\relax}
\providecommand{\bibinfo}[2]{#2}
\providecommand{\BIBentrySTDinterwordspacing}{\spaceskip=0pt\relax}
\providecommand{\BIBentryALTinterwordstretchfactor}{4}
\providecommand{\BIBentryALTinterwordspacing}{\spaceskip=\fontdimen2\font plus
\BIBentryALTinterwordstretchfactor\fontdimen3\font minus
  \fontdimen4\font\relax}
\providecommand{\BIBforeignlanguage}[2]{{%
\expandafter\ifx\csname l@#1\endcsname\relax
\typeout{** WARNING: IEEEtran.bst: No hyphenation pattern has been}%
\typeout{** loaded for the language `#1'. Using the pattern for}%
\typeout{** the default language instead.}%
\else
\language=\csname l@#1\endcsname
\fi
#2}}
\providecommand{\BIBdecl}{\relax}
\BIBdecl

\bibitem{ObjectDetectionReview-TNNLS2019}
Z.~Q. Zhao, P.~Zheng, S.~T. Xu, and X.~Wu, ``{Object detection with deep
  learning: A review},'' \emph{IEEE Transactions on Neural Networks and
  Learning Systems (TNNLS)}, vol.~30, no.~11, pp. 3212--3232, 2019.

\bibitem{ObjectDetectionSurvey-IJCV2020}
L.~Liu, W.~Ouyang, {\textperiodcentered}.~X. Wang, P.~Fieguth,
  {\textperiodcentered}.~J. Chen, {\textperiodcentered}.~X. Liu, and
  M.~Pietik{\"{a}}inen, ``{Deep Learning for Generic Object Detection: A
  Survey},'' \emph{International Journal of Computer Vision (IJCV)}, vol. 128,
  pp. 261--318, 2020.

\bibitem{medical-katzmann-neurocomputing2021}
A.~Katzmann, O.~Taubmann, S.~Ahmad, A.~M{\"{u}}hlberg, M.~S{\"{u}}hling, and
  H.-M. Gro{\ss}, ``{Explaining clinical decision support systems in medical
  imaging using cycle-consistent activation maximization},''
  \emph{Neurocomputing}, vol. 458, pp. 141--156, 2021.

\bibitem{DetectRareSpecies-ConsBio2021}
L.~Mannocci, S.~Villon, M.~Chaumont, N.~Guellati, N.~Mouquet, C.~Iovan,
  L.~Vigliola, and D.~Mouillot, ``Leveraging social media and deep learning to
  detect rare megafauna in video surveys,'' \emph{Conservation Biology}, 2021.

\bibitem{ObjectNameLearningHumans-Psycho-2002}
L.~B. Smith, S.~S. Jones, B.~Landau, L.~Gershkoff-Stowe, and L.~Samuelson,
  ``Object name learning provides on-the-job training for attention,''
  \emph{Psychological Science}, vol.~13, no.~1, pp. 13--19, 2002.

\bibitem{HumanObjectNaming-Development-2005}
L.~K. Samuelson and L.~B. Smith, ``They call it like they see it: Spontaneous
  naming and attention to shape,'' \emph{Developmental Science}, vol.~8, no.~2,
  pp. 182--198, 2005.

\bibitem{HumansWordMeaning-phd-2009}
L.~A. Schmidt, ``Meaning and compositionality as statistical induction of
  categories and constraints,'' Ph.D. dissertation, Massachusetts Institute of
  Technology, 2009.

\bibitem{LSTD-aaai2018}
H.~Chen, Y.~Wang, G.~Wang, and Y.~Qiao, ``{LSTD: A Low-Shot Transfer Detector
  for Object Detection},'' in \emph{AAAI Conference on Artificial Intelligence
  (AAAI)}, 2018, pp. 2836--2843.

\bibitem{MetaRCNN-iccv2019}
X.~Yan, Z.~Chen, A.~Xu, X.~Wang, X.~Liang, and L.~Lin, ``{Meta R-CNN: Towards
  general solver for instance-level low-shot learning},'' in
  \emph{International Conference on Computer Vision (ICCV)}, 2019, pp.
  9576--9585.

\bibitem{MatchingNets-NeurIPS2016}
O.~Vinyals, G.~Deepmind, C.~Blundell, T.~Lillicrap, K.~Kavukcuoglu, and
  D.~Wierstra, ``{Matching Networks for One Shot Learning},'' in \emph{Neural
  Information Processing Systems (NeurIPS)}, 2016, pp. 3630--3638.

\bibitem{PrototypicalNetworks-NeurIPS2017}
J.~Snell, K.~Swersky, T.~R. Zemel, and R.~Zemel, ``{Prototypical Networks for
  Few-shot Learning},'' in \emph{Conference on Neural Information Processing
  Systems (NeurIPS)}, vol.~30, 2017.

\bibitem{Survey-Few-Shot-Learning-CSUR2020}
Y.~Wang, J.~T. Kwok, L.~M. Ni, H.~Kong, and Q.~Yao, ``{Generalizing from a Few
  Examples: A Survey on Few-shot Learning},'' \emph{ACM Computing Surveys
  (CSUR)}, vol.~53, pp. 1--34, 2020.

\bibitem{Meta-Learning-Survey-TPAMI2021}
T.~M. Hospedales, A.~Antoniou, P.~Micaelli, and A.~J. Storkey, ``{Meta-Learning
  in Neural Networks: A Survey},'' \emph{IEEE Transactions on Pattern Analysis
  and Machine Intelligence (TPAMI)}, 2021.

\bibitem{Semi-Supervised-Book-2010}
O.~Chapelle, B.~Scholkopf, and A.~Zien, ``Semi-supervised learning,''
  \emph{IEEE Transactions on Neural Networks (TNN)}, vol.~20, no.~3, pp.
  542--542, 2010.

\bibitem{semi-Supervised-Survey-ML2020}
J.~E. Van~Engelen and H.~H. Hoos, ``A survey on semi-supervised learning,''
  \emph{Machine Learning (ML)}, vol. 109, no.~2, pp. 373--440, 2020.

\bibitem{Semi-Supervised-Detection-WACV2021}
P.~Tang, C.~Ramaiah, Y.~Wang, R.~Xu, C.~Xiong, and S.~Research, ``{Proposal
  Learning for Semi-Supervised Object Detection},'' in \emph{IEEE Winter
  Conference on Applications of Computer Vision (WACV)}, 2021, pp. 2291--2301.

\bibitem{incremental-algorithms-applications-esann2016}
A.~Gepperth and B.~Hammer, ``Incremental learning algorithms and
  applications,'' in \emph{European symposium on artificial neural networks
  (ESANN)}, 2016.

\bibitem{incremental-end-to-end-eccv2018}
F.~M. Castro, M.~J. Mar{\'\i}n-Jim{\'e}nez, N.~Guil, C.~Schmid, and K.~Alahari,
  ``End-to-end incremental learning,'' in \emph{European conference on computer
  vision (ECCV)}, 2018, pp. 233--248.

\bibitem{incremental-large-scale-cvpr2019}
Y.~Wu, Y.~Chen, L.~Wang, Y.~Ye, Z.~Liu, Y.~Guo, and Y.~Fu, ``Large scale
  incremental learning,'' in \emph{IEEE Conference on Computer Vision and
  Pattern Recognition (CVPR)}, 2019, pp. 374--382.

\bibitem{cai2019exploring}
Q.~Cai, Y.~Pan, C.-W. Ngo, X.~Tian, L.~Duan, and T.~Yao, ``Exploring object
  relation in mean teacher for cross-domain detection,'' in \emph{IEEE
  Conference on Computer Vision and Pattern Recognition (CVPR)}, 2019, pp.
  11\,457--11\,466.

\bibitem{xu2020cross}
M.~Xu, H.~Wang, B.~Ni, Q.~Tian, and W.~Zhang, ``Cross-domain detection via
  graph-induced prototype alignment,'' in \emph{IEEE Conference on Computer
  Vision and Pattern Recognition (CVPR)}, 2020, pp. 12\,355--12\,364.

\bibitem{chen2020harmonizing}
C.~Chen, Z.~Zheng, X.~Ding, Y.~Huang, and Q.~Dou, ``Harmonizing transferability
  and discriminability for adapting object detectors,'' in \emph{IEEE
  Conference on Computer Vision and Pattern Recognition (CVPR)}, 2020, pp.
  8869--8878.

\bibitem{zero-shot-object-detection-eccv2018}
A.~Bansal, K.~Sikka, G.~Sharma, R.~Chellappa, and A.~Divakaran, ``{Zero-Shot
  Object Detection},'' in \emph{European Conference on Computer Vision (ECCV)},
  2018, pp. 384--400.

\bibitem{ZSOD-accv2018}
S.~Rahman, S.~Khan, and F.~Porikli, ``{Zero-Shot Object Detection: Learning to
  Simultaneously Recognize and Localize Novel Concepts},'' in \emph{Asian
  Conference on Computer Vision (ACCV)}, 2018, pp. 547--563.

\bibitem{weaklySupervisedSurvey-tpami2021}
D.~Zhang, J.~Han, G.~Cheng, and M.-H. Yang, ``Weakly supervised object
  localization and detection: A survey,'' \emph{IEEE Transactions on Pattern
  Analysis and Machine Intelligence (TPAMI)}, 2021.

\bibitem{weaklySupervisedSurvey-arxiv2021}
F.~Shao, L.~Chen, J.~Shao, W.~Ji, S.~Xiao, L.~Ye, Y.~Zhuang, and J.~Xiao,
  ``Deep learning for weakly-supervised object detection and object
  localization: A survey,'' \emph{arXiv preprint arXiv:2105.12694}, 2021.

\bibitem{Deep-Metric-Learning-Survey-2019}
M.~Kaya and H.~{\c{S}}. Bilge, ``Deep metric learning: A survey,''
  \emph{Symmetry}, vol.~11, no.~9, p. 1066, 2019.

\bibitem{TripletLoss-jmlr2009}
K.~Q. Weinberger and L.~K. Saul, ``{Distance Metric Learning for Large Margin
  Nearest Neighbor Classification},'' \emph{Journal of Machine Learning
  Research (JMLR)}, vol.~10, pp. 207--244, 2009.

\bibitem{Aganian-ReID-icann2021}
D.~Aganian, M.~Eisenbach, J.~Wagner, D.~Seichter, and H.-M. Gross, ``Revisiting
  loss functions for person re-identification,'' in \emph{International
  Conference on Artificial Neural Networks (ICANN)}.\hskip 1em plus 0.5em minus
  0.4em\relax Springer, 2021, pp. 30--42.

\bibitem{FSOD-Survey-acm2022}
S.~Antonelli, D.~Avola, L.~Cinque, D.~Crisostomi, G.~L. Foresti, F.~Galasso,
  M.~R. Marini, A.~Mecca, and D.~Pannone, ``Few-shot object detection: A
  survey,'' \emph{ACM Computing Surveys}, 2 2022.

\bibitem{FSOD-comparative-review-arxiv2021}
L.~Jiaxu, C.~Taiyue, G.~Xinbo, Y.~Yongtao, W.~Ye, G.~Feng, and W.~Yue, ``A
  comparative review of recent few-shot object detection algorithms,''
  \emph{arXiv preprint arXiv:2111.00201}, 2021.

\bibitem{FSOD-empirical-study-arxiv2022}
T.~Liu, L.~Zhang, Y.~Wang, J.~Guan, Y.~Fu, and S.~Zhou, ``An empirical study
  and comparison of recent few-shot object detection algorithms,'' \emph{arXiv
  preprint arXiv:2203.14205}, 2022.

\bibitem{Low-Shot-Detection-survey-arxiv2021}
Q.~Huang, H.~Zhang, J.~Song, and M.~Song, ``A survey of deep learning for
  low-shot object detection,'' \emph{arXiv preprint arXiv:2112.02814}, 2021.

\bibitem{FSOD-self-supervised-survey-arxiv2021}
G.~Huang, I.~Laradji, D.~Vázquez, S.~Lacoste-Julien, and P.~Rodríguez, ``A
  survey of self-supervised and few-shot object detection,'' \emph{arXiv
  preprint arXiv:2110.14711}, 2021.

\bibitem{FSOD-survey-chinese}
C.~Liu, C.~Tianen, W.~Cong, J.~Shuwen, and C.~Dong, ``A survey of few-shot
  object detection,'' \emph{Journal of Frontiers of Computer Science and
  Technology}, 2022.

\bibitem{SQMG-cpvr2021}
L.~Zhang, S.~Zhou, J.~Guan, and J.~Zhang, ``{Accurate Few-shot Object Detection
  with Support-Query Mutual Guidance and Hybrid Loss},'' in \emph{IEEE
  Conference on Computer Vision and Pattern Recognition (CVPR)}, 2021, pp.
  14\,424--14\,432.

\bibitem{MaskRCNN}
K.~He, G.~Gkioxari, P.~Doll{\'{a}}r, and R.~Girshick, ``{Mask R-CNN},''
  \emph{International Conference on Computer Vision (ICCV)}, pp. 2961--2969,
  2017.

\bibitem{FSDetView-eecv2020}
Y.~Xiao and R.~Marlet, ``{Few-Shot Object Detection and Viewpoint Estimation
  for Objects in the Wild},'' in \emph{European Conference on Computer Vision
  (ECCV)}, 2020, pp. 192--210.

\bibitem{TIP-cvpr2021}
A.~Li and Z.~Li, ``{Transformation Invariant Few-Shot Object Detection},'' in
  \emph{IEEE Conference on Computer Vision and Pattern Recognition (CVPR)},
  2021, pp. 3094--3102.

\bibitem{MetaYOLO-iccv2019}
B.~Kang, Z.~Liu, X.~Wang, F.~Yu, J.~Feng, and T.~Darrell, ``{Few-shot object
  detection via feature reweighting},'' in \emph{International Conference on
  Computer Vision (ICCV)}, 2019, pp. 8419--8428.

\bibitem{CME-cvpr2021}
B.~Li, B.~Yang, C.~Liu, F.~Liu, R.~Ji, and Q.~Ye, ``{Beyond Max-Margin: Class
  Margin Equilibrium for Few-shot Object Detection},'' in \emph{IEEE Conference
  on Computer Vision and Pattern Recognition (CVPR)}, 2021.

\bibitem{SPCD-iciap2022}
D.~Kobayashi, ``Self-supervised prototype conditional few-shot object
  detection,'' in \emph{Image Analysis and Processing (ICIAP)}.\hskip 1em plus
  0.5em minus 0.4em\relax Springer International Publishing, 2022, pp.
  681--692.

\bibitem{IFC-apin2022}
M.~Wang, H.~Ning, and H.~Liu, ``Object detection based on few-shot learning via
  instance-level feature correlation and aggregation,'' \emph{Applied
  Intelligence}, pp. 1--18, 2022.

\bibitem{ARRM-electronimaging2022}
L.~Huang, Z.~He, and X.~Feng, ``Few-shot object detection with affinity
  relation reasoning,'' \emph{Journal of Electronic Imaging}, vol.~31, p.
  033016, 2022.

\bibitem{MM-FSOD-arxiv2020}
Y.~Li, W.~Feng, S.~Lyu, Q.~Zhao, and X.~Li, ``{MM-FSOD: Meta and metric
  integrated few-shot object detection},'' \emph{arXiv preprint
  arXiv:2012.15159}, pp. 1--30, 2020.

\bibitem{GenDet-tnnls2021}
L.~Liu, B.~Wang, Z.~Kuang, J.~H. Xue, Y.~Chen, W.~Yang, Q.~Liao, and W.~Zhang,
  ``{GenDet: Meta Learning to Generate Detectors From Few Shots},'' \emph{IEEE
  Transactions on Neural Networks and Learning Systems (TNNLS)}, pp. 1--13,
  2021.

\bibitem{CAReD-displays2022}
J.~Quan, B.~Ge, and L.~Chen, ``Cross attention redistribution with contrastive
  learning for few shot object detection,'' \emph{Displays}, vol.~72, 2022.

\bibitem{DAnA-tmm2021}
T.-I. Chen, Y.-C. Liu, H.-T. Su, Y.-C. Chang, Y.-H. Lin, J.-F. Yeh, W.-C. Chen,
  and W.~H. Hsu, ``{Dual-awareness Attention for Few-Shot Object Detection},''
  \emph{IEEE Transactions on Multimedia (TMM)}, vol.~23, 2021.

\bibitem{APSP-wacv2022}
H.~Lee, M.~Lee, and N.~Kwak, ``Few-shot object detection by attending to
  per-sample-prototype,'' in \emph{IEEE Winter Conference on Applications of
  Computer Vision (WACV)}, 2022, pp. 2445--2454.

\bibitem{AttentionRPN-cvpr2020}
Q.~Fan, W.~Zhuo, C.-K. Tang, and Y.-W. Tai, ``{Few-Shot Object Detection with
  Attention-RPN and Multi-Relation Detector},'' in \emph{IEEE Conference on
  Computer Vision and Pattern Recognition (CVPR)}, 2020, pp. 4013--4022.

\bibitem{AIT-cvpr2021}
D.-J. Chen, H.-Y. Hsieh, and T.-L. Liu, ``{Adaptive Image Transformer for
  One-Shot Object Detection},'' in \emph{IEEE Conference on Computer Vision and
  Pattern Recognition (CVPR)}, 2021, pp. 12\,247--12\,256.

\bibitem{CoAE-neurips2019}
T.-i. Hsieh, Y.-c. Lo, H.-t. Chen, and T.-l. Liu, ``{One-Shot Object Detection
  with Co-Attention and Co-Excitation},'' in \emph{Conference on Neural
  Information Processing Systems (NeurIPS)}, 2019.

\bibitem{ClosingGeneralizationGap-arxiv2020}
C.~Michaelis, M.~Bethge, and A.~S. Ecker, ``{Closing the Generalization Gap in
  One-Shot Object Detection},'' \emph{arXiv preprint arXiv:2011.04267}, pp.
  1--13, 2020.

\bibitem{MetaFasterRCNN-aaai2022}
G.~Han, S.~Huang, J.~Ma, Y.~He, and S.-F. Chang, ``Meta faster r-cnn: Towards
  accurate few-shot object detection with attentive feature alignment,'' in
  \emph{AAAI Conference on Artificial Intelligence (AAAI)}, 2022, pp. 780--789.

\bibitem{OneShotWithoutFinetuning-arxiv2020}
X.~Li, L.~Zhang, Y.~P. Chen, Y.-W. Tai, and C.-K. Tang, ``{One-Shot Object
  Detection without Fine-Tuning},'' \emph{arXiv preprint arXiv:2005.03819},
  2020.

\bibitem{QAFewDet-iccv2021}
G.~Han, Y.~He, S.~Huang, J.~Ma, and S.-F. Chang, ``Query adaptive few-shot
  object detection with heterogeneous graph convolutional networks,'' in
  \emph{International Conference on Computer Vision (ICCV)}, 2021, pp.
  3263--3272.

\bibitem{DCNet-cvpr2021}
H.~Hu, S.~Bai, A.~Li, J.~Cui, and L.~Wang, ``{Dense Relation Distillation with
  Context-aware Aggregation for Few-Shot Object Detection},'' in \emph{IEEE
  Conference on Computer Vision and Pattern Recognition (CVPR)}, 2021.

\bibitem{FewShotKnowledgeTransfer-smc2020}
G.~Kim, H.-G. Jung, and S.-W. Lee, ``{Few-Shot Object Detection via Knowledge
  Transfer},'' in \emph{IEEE International Conference on Systems, Man, and
  Cybernetics (SMC)}, 2020.

\bibitem{Meta-DETR-tpami2022}
G.~Zhang, Z.~Luo, K.~Cui, S.~Lu, and E.~Xing, ``Meta-detr: Image-level few-shot
  detection with inter-class correlation exploitation,'' \emph{IEEE
  Transactions on Pattern Analysis and Machine Intelligence (TPAMI)}, 2022.

\bibitem{ONCE-cvpr2020}
X.~Zhu, T.~Hospedales, and T.~Xiang, ``{Incremental Few-Shot Object
  Detection},'' in \emph{IEEE Conference on Computer Vision and Pattern
  Recognition (CVPR)}, 2020.

\bibitem{FCT-cvpr2022}
G.~Han, J.~Ma, S.~Huang, L.~Chen, and S.-F. Chang, ``Few-shot object detection
  with fully cross-transformer,'' in \emph{IEEE Conference on Computer Vision
  and Pattern Recognition (CVPR)}, 2022, pp. 5321--5330.

\bibitem{Sylph-cvpr2022}
L.~Yin, J.~M. Perez-Rua, and L.~K. J, ``Sylph: A hypernetwork framework for
  incremental few-shot object detection,'' in \emph{IEEE Conference on Computer
  Vision and Pattern Recognition (CVPR)}, 2022, pp. 9035--9045.

\bibitem{PNSD-accv2020}
S.~Zhang, D.~Luo, L.~Wang, and P.~Koniusz, ``Few-shot object detection by
  second-order pooling,'' in \emph{Asian Conference on Computer Vision (ACCV)},
  2020.

\bibitem{KFSOD-cvpr2022}
S.~Zhang, L.~Wang, N.~Murray, and P.~Koniusz, ``Kernelized few-shot object
  detection with efficient integral aggregation,'' in \emph{IEEE Conference on
  Computer Vision and Pattern Recognition (CVPR)}, 2022, pp. 19\,207--19\,216.

\bibitem{FasterRCNN-neurips2015}
S.~Ren, K.~He, R.~Girshick, and J.~Sun, ``Faster {R-CNN}: Towards real-time
  object detection with region proposal networks,'' in \emph{Conference on
  Neural Information Processing Systems (NeurIPS)}, 2015.

\bibitem{ResNet-cvpr2016}
K.~He, X.~Zhang, S.~Ren, and J.~Sun, ``{Deep Residual Learning for Image
  recognition},'' in \emph{IEEE Conference on Computer Vision and Pattern
  Recognition (CVPR)}, 2016, pp. 770--778.

\bibitem{power-normalization-wacv2019}
H.~Zhang and P.~Koniusz, ``Power normalizing second-order similarity network
  for few-shot learning,'' in \emph{IEEE Winter Conference on Applications of
  Computer Vision (WACV)}, 2019, pp. 1185--1193.

\bibitem{kernelized-covariance-ijcv2021}
J.~Zhang, L.~Wang, L.~Zhou, and W.~Li, ``Beyond covariance: Sice and kernel
  based visual feature representation,'' \emph{International Journal of
  Computer Vision (IJCV)}, vol. 129, no.~2, pp. 300--320, 2021.

\bibitem{kernels-regularization-springer2003}
A.~J. Smola and R.~Kondor, ``Kernels and regularization on graphs,'' in
  \emph{Learning theory and kernel machines}.\hskip 1em plus 0.5em minus
  0.4em\relax Springer, 2003, pp. 144--158.

\bibitem{OSIS-arxiv2018}
C.~Michaelis, I.~Ustyuzhaninov, M.~Bethge, and A.~S. Ecker, ``{One-Shot
  Instance Segmentation},'' \emph{arXiv preprint arXiv:1811.11507}, 2018.

\bibitem{DynamicConvolution-cvpr2020}
Y.~Chen, X.~Dai, M.~Liu, D.~Chen, L.~Yuan, and Z.~Liu, ``Dynamic convolution:
  Attention over convolution kernels,'' in \emph{IEEE Conference on Computer
  Vision and Pattern Recognition (CVPR)}, 2020, pp. 11\,030--11\,039.

\bibitem{survey-vision-transformer-tpami2022}
K.~Han, Y.~Wang, H.~Chen, X.~Chen, J.~Guo, Z.~Liu, Y.~Tang, A.~Xiao, C.~Xu,
  Y.~Xu \emph{et~al.}, ``A survey on vision transformer,'' \emph{IEEE
  Transactions on Pattern Analysis and Machine Intelligence (TPAMI)}, 2022.

\bibitem{NonLocal-cvpr2018}
X.~Wang, R.~Girshick, A.~Gupta, and K.~He, ``{Non-local Neural Networks},'' in
  \emph{IEEE Conference on Computer Vision and Pattern Recognition (CVPR)},
  2018, pp. 7794--7803.

\bibitem{Transformer-neurips2017}
A.~Vaswani, N.~Shazeer, N.~Parmar, J.~Uszkoreit, L.~Jones, A.~N. Gomez,
  {\L}.~Kaiser, and I.~Polosukhin, ``{Attention Is All You Need},'' in
  \emph{Conference on Neural Information Processing Systems (NeurIPS)}, 2017.

\bibitem{soft-threshold-shrinkage-trans-ind-inform2019}
M.~Zhao, S.~Zhong, X.~Fu, B.~Tang, and M.~Pecht, ``Deep residual shrinkage
  networks for fault diagnosis,'' \emph{IEEE Transactions on Industrial
  Informatics}, vol.~16, no.~7, pp. 4681--4690, 2019.

\bibitem{SENet-cvpr2018}
J.~Hu, L.~Shen, and G.~Sun, ``{Squeeze-and-excitation networks},'' in
  \emph{IEEE Conference on Computer Vision and Pattern Recognition (CVPR)},
  2018, pp. 7132--7141.

\bibitem{PVTv2-cvm2022}
W.~Wang, E.~Xie, X.~Li, D.-P. Fan, K.~Song, D.~Liang, T.~Lu, P.~Luo, and
  L.~Shao, ``Pvt v2: Improved baselines with pyramid vision transformer,''
  \emph{Computational Visual Media}, vol.~8, no.~3, pp. 415--424, 2022.

\bibitem{RelationNetwork-cvpr2018}
F.~Sung, Y.~Yang, L.~Zhang, T.~Xiang, P.~H. Torr, and T.~M. Hospedales,
  ``Learning to compare: Relation network for few-shot learning,'' in
  \emph{IEEE Conference on Computer Vision and Pattern Recognition (CVPR)},
  2018, pp. 1199--1208.

\bibitem{GCN-iclr2017}
T.~N. Kipf and M.~Welling, ``Semi-supervised classification with graph
  convolutional networks,'' in \emph{International Conference on Learning
  Representations (ICLR)}, 2017.

\bibitem{AdaptiveMargin-cvpr2020}
A.~Li, W.~Huang, X.~Lan, J.~Feng, Z.~Li, and L.~Wang, ``Boosting few-shot
  learning with adaptive margin loss,'' in \emph{IEEE Conference on Computer
  Vision and Pattern Recognition (CVPR)}, 2020, pp. 12\,576--12\,584.

\bibitem{WordEmbeddingGlove-emnlp2014}
J.~Pennington, R.~Socher, and C.~D. Manning, ``Glove: Global vectors for word
  representation,'' in \emph{Conference on Empirical Methods in Natural
  Language Processing (EMNLP)}, 2014, pp. 1532--1543.

\bibitem{RetinaNet-iccv2017}
T.~Y. Lin, P.~Goyal, R.~Girshick, K.~He, and P.~Dollar, ``{Focal Loss for Dense
  Object Detection},'' in \emph{International Conference on Computer Vision
  (ICCV)}, vol.~42, 2017, pp. 318--327.

\bibitem{selective-search-ijcv2013}
J.~R. Uijlings, K.~E. Van De~Sande, T.~Gevers, and A.~W. Smeulders, ``Selective
  search for object recognition,'' \emph{International Journal of Computer
  Vision (IJCV)}, vol. 104, no.~2, pp. 154--171, 2013.

\bibitem{YOLOv2-cvpr2017}
J.~Redmon and A.~Farhadi, ``{YOLO9000: Better, faster, stronger},'' in
  \emph{IEEE Conference on Computer Vision and Pattern Recognition (CVPR)},
  2017, pp. 6517--6525.

\bibitem{CenterNet_ObjectsAsPoints-arxiv2019}
X.~Zhou, D.~Wang, and P.~Kr{\"{a}}henb{\"{u}}hl, ``{Objects as Points},''
  \emph{arXiv preprint arXiv:1904.07850}, 2019.

\bibitem{FCOS-iccv2019}
Z.~Tian, C.~Shen, H.~Chen, and T.~He, ``{FCOS: Fully convolutional one-stage
  object detection},'' in \emph{International Conference on Computer Vision
  (ICCV)}, 2019, pp. 9627--9636.

\bibitem{DeformableDETR-iclr2021}
X.~Zhu, W.~Su, L.~Lu, B.~Li, X.~Wang, J.~Dai, and S.~Research, ``{Deformable
  DETR: Deformable Transformers for End-to-End Object Detection},'' in
  \emph{International Conference on Learning Representations (ICLR)}, 2021.

\bibitem{PSPNet-cvpr2017}
H.~Zhao, J.~Shi, X.~Qi, X.~Wang, and J.~Jia, ``{Pyramid Scene Parsing
  Network},'' in \emph{IEEE Conference on Computer Vision and Pattern
  Recognition (CVPR)}, 2017.

\bibitem{RepMet-cvpr2019}
L.~Karlinsky, J.~Shtok, S.~Harary, E.~Schwartz, A.~Aides, R.~Feris, R.~Giryes,
  and A.~M. Bronstein, ``{RepMet: Representative-based metric learning for
  classification and few-shot object detection},'' in \emph{IEEE Conference on
  Computer Vision and Pattern Recognition (CVPR)}, 2019, pp. 5192--5201.

\bibitem{NP-RepMet-neurips2020}
Y.~Yang, F.~Wei, M.~Shi, and G.~Li, ``{Restoring negative information in
  few-shot object detection},'' in \emph{Conference on Neural Information
  Processing Systems (NeurIPS)}, 2020.

\bibitem{PNPDet-wacv2021}
G.~Zhang, K.~Cui, R.~Wu, S.~Lu, and Y.~Tian, ``{PNPDet : Efficient Few-shot
  Detection without Forgetting},'' in \emph{Winter Conference on Applications
  of Computer Vision (WACV)}, 2021, pp. 3823--3832.

\bibitem{MetaDet-iccv2019}
Y.~X. Wang, D.~Ramanan, and M.~Hebert, ``{Meta-Learning to Detect Rare
  Objects},'' in \emph{International Conference on Computer Vision (ICCV)},
  2019, pp. 9924--9933.

\bibitem{MetaRetinaNet-bmvc2020}
S.~Li, W.~Song, S.~Li, A.~Hao, and H.~Quin, ``{Meta-RetinaNet for Few-shot
  Object Detection},'' in \emph{British Machine Vision Conference (BMVC)},
  2020.

\bibitem{MetaSSD-ieee2019}
K.~Fu, T.~Zhang, Y.~Zhang, M.~Yan, Z.~Chang, Z.~Zhang, and X.~Sun, ``{Meta-SSD:
  Towards Fast Adaptation for Few-Shot Object Detection with Meta-Learning},''
  \emph{IEEE Access}, vol.~7, pp. 77\,597--77\,606, 2019.

\bibitem{SSD-eccv2016}
W.~Liu, D.~Anguelov, D.~Erhan, C.~Szegedy, S.~Reed, C.~Y. Fu, and A.~C. Berg,
  ``{SSD: Single shot multibox detector},'' in \emph{European Conference on
  Computer Vision (ECCV)}, 2016, pp. 21--37.

\bibitem{TFA-icml2020}
X.~Wang, T.~E. Huang, T.~Darrell, J.~E. Gonzalez, and F.~Yu, ``{Frustratingly
  Simple Few-Shot Object Detection},'' in \emph{International Conference on
  Machine Learning (ICML)}, 2020.

\bibitem{FSCE_FSOD-cvpr2021}
B.~Sun, B.~Li, S.~Cai, Y.~Yuan, and C.~Zhang, ``{FSCE: Few-Shot Object
  Detection via Contrastive Proposal Encoding},'' in \emph{IEEE Conference on
  Computer Vision and Pattern Recognition (CVPR)}, 2021.

\bibitem{BPMCH-prl2022}
H.~Feng, L.~Zhang, X.~Yang, Z.~Liu, and J.~Lu, ``Incremental few-shot object
  detection via knowledge transfer,'' \emph{Pattern Recognition Letters}, vol.
  156, pp. 67--73, 2022.

\bibitem{MemFRCN-tfeccs2022}
T.~Lu, S.~Jia, and H.~Zhang, ``Memfrcn: Few shot object detection with
  memorable faster-rcnn,'' \emph{IEICE Transactions on Fundamentals of
  Electronics, Communications and Computer Sciences (TFECCS)}, 2022.

\bibitem{Halluc.-cvpr2021}
W.~Zhang and Y.-X. Wang, ``{Hallucination Improves Few-Shot Object
  Detection},'' in \emph{IEEE Conference on Computer Vision and Pattern
  Recognition (CVPR)}, 2021.

\bibitem{RetentiveRCNN-cvpr2021}
Z.~Fan, Y.~Ma, Z.~Li, and J.~Sun, ``{Generalized Few-Shot Object Detection
  without Forgetting},'' in \emph{IEEE Conference on Computer Vision and
  Pattern Recognition (CVPR)}, 2021.

\bibitem{CooperatingRPNs-arxiv2020}
W.~Zhang, Y.-X. Wang, and D.~A. Forsyth, ``{Cooperating RPN's Improve Few-Shot
  Object Detection},'' \emph{arXiv preprint arXiv:2011.10142}, pp. 1--10, 2020.

\bibitem{LVC-cvpr2022}
P.~Kaul, W.~Xie, and A.~Zisserman, ``Label, verify, correct: A simple few shot
  object detection method,'' in \emph{IEEE Conference on Computer Vision and
  Pattern Recognition (CVPR)}, 2022, pp. 14\,237--14\,247.

\bibitem{FORD+BL-imavis2022}
A.~K.~N. Vu, N.~D. Nguyen, K.~D. Nguyen, V.~T. Nguyen, T.~D. Ngo, T.~T. Do, and
  T.~V. Nguyen, ``Few-shot object detection via baby learning,'' \emph{Image
  and Vision Computing}, vol. 120, 2022.

\bibitem{CIR-remote-sensing-2022}
Y.~Wang, C.~Xu, C.~Liu, and Z.~Li, ``Context information refinement for
  few-shot object detection in remote sensing images,'' \emph{Remote Sensing},
  vol.~14, 2022.

\bibitem{MPSR-eecv2020}
J.~Wu, S.~Liu, D.~Huang, and Y.~Wang, ``{Multi-Scale Positive Sample Refinement
  for Few-Shot Object Detection},'' in \emph{European Conference on Computer
  Vision (ECCV)}, 2020, pp. 456--472.

\bibitem{SVD-Dictionary-neurips2021}
A.~Wu, S.~Zhao, C.~Deng, and W.~Liu, ``{Generalized and Discriminative Few-Shot
  Object Detection via SVD-Dictionary Enhancement},'' in \emph{Neural
  Information Processing Systems (NeurIPS)}, 2021.

\bibitem{FSOD-UP-iccv2021}
A.~Wu, Y.~Han, L.~Zhu, and Y.~Yang, ``Universal-prototype enhancing for
  few-shot object detection,'' in \emph{International Conference on Computer
  Vision (ICCV)}, 2021, pp. 9567--9576.

\bibitem{DeFRCN-iccv2021}
L.~Qiao, Y.~Zhao, Z.~Li, X.~Qiu, J.~Wu, and C.~Zhang, ``Defrcn: Decoupled
  faster r-cnn for few-shot object detection,'' in \emph{International
  Conference on Computer Vision (ICCV)}, 2021, pp. 8681--8690.

\bibitem{DMNet-tcyb2022}
Y.~Lu, X.~Chen, Z.~Wu, and J.~Yu, ``Decoupled metric network for single-stage
  few-shot object detection,'' \emph{IEEE Transactions on Cybernetics}, 2022.

\bibitem{CGDP+FSCN-cvpr2021}
Y.~Li, H.~Zhu, Y.~Cheng, and W.~Wang, ``{Few-Shot Object Detection via
  Classification Refinement and Distractor Retreatment},'' in \emph{IEEE
  Conference on Computer Vision and Pattern Recognition (CVPR)}, 2021, pp.
  15\,395--15\,403.

\bibitem{N-PME-icassp2022}
W.~Liu, C.~Wang, S.~Yu, C.~Tao, J.~Wang, and J.~Wu, ``Novel instance mining
  with pseudo-margin evaluation for few-shot object detection,'' in \emph{IEEE
  International Conference on Acoustics, Speech and Signal Processing
  (ICASSP)}, 2022, pp. 2250--2254.

\bibitem{TD-Sampler-icccbda2022}
C.~Wu, B.~Wang, S.~Liu, X.~Liu, and P.~Wu, ``Td-sampler: Learning a training
  difficulty based sampling strategy for few-shot object detection,'' in
  \emph{International Conference on Cloud Computing and Big Data Analytics
  (ICCCBDA)}, 2022, pp. 275--279.

\bibitem{FADI-neurips2021}
Y.~Cao, J.~Wang, Y.~Jin, T.~Wu, K.~Chen, Z.~Liu, and D.~Lin, ``{Few-Shot Object
  Detection via Association and DIscrimination},'' in \emph{Conference on
  Neural Information Processing Systems (NeurIPS)}, 2021.

\bibitem{ContextTransformer-aaai2020}
Z.~Yang, Y.~Wang, X.~Chen, J.~Liu, and Y.~Qiao, ``{Context-Transformer:
  Tackling Object Confusion for Few-Shot Detection},'' in \emph{AAAI Conference
  on Artificial Intelligence (AAAI)}, 2020.

\bibitem{UniT-cvpr2021}
S.~Khandelwal, R.~Goyal, and L.~Sigal, ``{UniT: Unified Knowledge Transfer for
  Any-shot Object Detection and Segmentation},'' in \emph{IEEE Conference on
  Computer Vision and Pattern Recognition (CVPR)}, 2021.

\bibitem{SRR-FSD-cvpr2021}
C.~Zhu, F.~Chen, U.~Ahmed, Z.~Shen, and M.~Savvides, ``{Semantic Relation
  Reasoning for Shot-Stable Few-Shot Object Detection},'' in \emph{IEEE
  Conference on Computer Vision and Pattern Recognition (CVPR)}, 2021.

\bibitem{CFA-cvpr2022}
K.~Guirguis, A.~Hendawy, G.~Eskandar, M.~Abdelsamad, M.~Kayser, and J.~Beyerer,
  ``Cfa: Constraint-based finetuning approach for generalized few-shot object
  detection,'' in \emph{IEEE Conference on Computer Vision and Pattern
  Recognition (CVPR)}, 2022, pp. 4039--4049.

\bibitem{KR-FSOD-electronics2022}
J.~Wang and D.~Chen, ``Few-shot object detection method based on knowledge
  reasoning,'' \emph{Electronics}, vol.~11, 2022.

\bibitem{FSSP-ieee2021}
H.~Xu, X.~Wang, F.~Shao, B.~Duan, and P.~Zhang, ``{Few-Shot Object Detection
  via Sample Processing},'' \emph{IEEE Access}, pp. 29\,207--29\,221, 2021.

\bibitem{BottomUpTopDownAttention-arxiv2020}
X.~Chen, M.~Jiang, and Q.~Zhao, ``{Leveraging Bottom-Up and Top-Down Attention
  for Few-Shot Object Detection},'' \emph{arXiv preprint arXiv:2007.12104}, pp.
  1--12, 2020.

\bibitem{ASPP-eccv2018}
L.-C. Chen, Y.~Zhu, G.~Papandreou, F.~Schroff, and H.~Adam, ``{Encoder-Decoder
  with Atrous Separable Convolution for Semantic Image Segmentation},'' in
  \emph{European Conference on Computer Vision (ECCV)}, 2018, pp. 801--818.

\bibitem{ViT-iclr2021}
A.~Dosovitskiy, L.~Beyer, A.~Kolesnikov, D.~Weissenborn, X.~Zhai,
  T.~Unterthiner, M.~Dehghani, M.~Minderer, G.~Heigold, S.~Gelly \emph{et~al.},
  ``{An Image is Worth 16x16 Words: Transformers for Image Recognition at
  Scale},'' in \emph{International Conference on Learning Representations
  (ICLR)}, 2021.

\bibitem{DINO-iccv2021}
M.~Caron, H.~Touvron, I.~Misra, H.~J{\'e}gou, J.~Mairal, P.~Bojanowski, and
  A.~Joulin, ``{Emerging Properties in Self-Supervised Vision Transformers},''
  in \emph{International Conference on Computer Vision (ICCV)}, 2021, pp.
  9650--9660.

\bibitem{Cascade-R-CNN-cvpr-2018}
Z.~Cai and N.~Vasconcelos, ``{Cascade R-CNN: Delving Into High Quality Object
  Detection},'' in \emph{IEEE Conference on Computer Vision and Pattern
  Recognition (CVPR)}, 2018, pp. 6154--6162.

\bibitem{imprinting-cvpr2018}
H.~Qi, M.~Brown, and D.~G. Lowe, ``Low-shot learning with imprinted weights,''
  in \emph{IEEE Conference on Computer Vision and Pattern Recognition (CVPR)},
  2018, pp. 5822--5830.

\bibitem{imprinting-tip2020}
X.~Chen, Y.~Wang, J.~Liu, and Y.~Qiao, ``Did:
  Disentangling-imprinting-distilling for continuous low-shot detection,''
  \emph{IEEE Transactions on Image Processing (TIP)}, vol.~29, pp. 7765--7778,
  2020.

\bibitem{WordNet-acm1995}
G.~A. Miller, ``Wordnet: a lexical database for english,'' \emph{Communications
  of the ACM}, vol.~38, no.~11, pp. 39--41, 1995.

\bibitem{GEM-neurips2017}
D.~Lopez-Paz and M.~Ranzato, ``{Gradient Episodic Memory for Continual
  Learning},'' in \emph{Advances in Neural Information Processing Systems
  (NeurIPS)}, 2017, p. 6470––6479.

\bibitem{AGEM-iclr2019}
A.~Chaudhry, M.~Ranzato, M.~Rohrbach, and M.~Elhoseiny, ``{Efficient Lifelong
  Learning with A-GEM},'' in \emph{International Conference on Learning
  Representations (ICLR)}, 2019.

\bibitem{ArcFace-cvpr2019}
J.~Deng, J.~Guo, N.~Xue, and S.~Zafeiriou, ``Arcface: Additive angular margin
  loss for deep face recognition,'' in \emph{IEEE Conference on Computer Vision
  and Pattern Recognition (CVPR)}, 2019, pp. 4690--4699.

\bibitem{saliency-bms}
J.~Zhang and S.~Sclaroff, ``Saliency detection: A boolean map approach,'' in
  \emph{International Conference on Computer Vision (ICCV)}, 2013, pp.
  153--160.

\bibitem{saliency-sam}
M.~Cornia, L.~Baraldi, G.~Serra, and R.~Cucchiara, ``Predicting human eye
  fixations via an lstm-based saliency attentive model,'' \emph{IEEE
  Transactions on Image Processing (TIP)}, vol.~27, no.~10, pp. 5142--5154,
  2018.

\bibitem{dictionary-learning-cvpr2017}
H.~Zhang, J.~Xue, and K.~Dana, ``Deep ten: Texture encoding network,'' in
  \emph{IEEE Conference on Computer Vision and Pattern Recognition (CVPR)},
  2017, pp. 708--717.

\bibitem{prototype-iccv2019}
K.~Wang, J.~H. Liew, Y.~Zou, D.~Zhou, and J.~Feng, ``Panet: Few-shot image
  semantic segmentation with prototype alignment,'' in \emph{International
  Conference on Computer Vision (ICCV)}, 2019, pp. 9197--9206.

\bibitem{prototype-eccv2020}
J.~Liu, L.~Song, and Y.~Qin, ``Prototype rectification for few-shot learning,''
  in \emph{European Conference on Computer Vision (ECCV)}, 2020, pp. 741--756.

\bibitem{YOLOv3-arxiv2018}
J.~Redmon and A.~Farhadi, ``Yolo3: An incremental improvement,'' \emph{arXiv
  preprint arXiv:1804.02767}, 2018.

\bibitem{VOC}
M.~Everingham, L.~{Van Gool}, C.~K. Williams, J.~Winn, and A.~Zisserman, ``{The
  Pascal Visual Object Classes (VOC) Challenge},'' \emph{International Journal
  of Computer Vision (IJCV)}, vol.~88, no.~2, pp. 303--338, 2010.

\bibitem{COCO}
T.-Y. Lin, M.~Maire, S.~Belongie, J.~Hays, P.~Perona, D.~Ramanan,
  P.~Doll{\'a}r, and C.~L. Zitnick, ``{Microsoft COCO: Common Objects in
  Context},'' in \emph{European Conference on Computer Vision (ECCV)}, 2014,
  pp. 740--755.

\bibitem{LVIS-cvpr2019}
A.~Gupta, P.~Dollar, and R.~Girshick, ``{Lvis: A Dataset for Large Vocabulary
  Instance Segmentation},'' in \emph{IEEE Conference on Computer Vision and
  Pattern Recognition (CVPR)}, 2019, pp. 5351--5359.

\bibitem{StarNet-aaai2021}
L.~Karlinsky, J.~Shtok, A.~Alfassy, M.~Lichtenstein, S.~Harary, E.~Schwartz,
  S.~Doveh, P.~Sattigeri, R.~Feris, A.~Bronstein, and R.~Giryes, ``{StarNet:
  towards Weakly Supervised Few-Shot Object Detection},'' in \emph{AAAI
  Conference on Artificial Intelligence (AAAI)}, 2021, pp. 1743--1753.

\bibitem{FGN-few-shot-instance-cvpr2020}
Z.~Fan, J.-G. Yu, Z.~Liang, J.~Ou, C.~Gao, G.-S. Xia, and Y.~Li, ``Fgn: Fully
  guided network for few-shot instance segmentation,'' in \emph{IEEE Conference
  on Computer Vision and Pattern Recognition (CVPR)}, 2020, pp. 9172--9181.

\bibitem{FAPIS-few-shot-instance-cvpr2021}
K.~Nguyen and S.~Todorovic, ``Fapis: A few-shot anchor-free part-based instance
  segmenter,'' in \emph{IEEE Conference on Computer Vision and Pattern
  Recognition (CVPR)}, 2021, pp. 11\,099--11\,108.

\bibitem{LearningSegmentTail-cvpr2020}
X.~Hu, Y.~Jiang, K.~Tang, J.~Chen, C.~Miao, and H.~Zhang, ``{Learning to
  Segment the Tail},'' in \emph{IEEE Conference on Computer Vision and Pattern
  Recognition (CVPR)}, 2020, pp. 14\,042--14\,051.

\bibitem{FSOD-toolchain-mmm2022}
W.~Bailer, ``Making few-shot object detection simpler and less frustrating,''
  in \emph{International Conference on Multimedia Modeling (MMM)}, 2022, pp.
  445--451.

\bibitem{FPN-cvpr2017}
T.-Y. Lin, P.~Doll{\'a}r, R.~Girshick, K.~He, B.~Hariharan, and S.~Belongie,
  ``{Feature Pyramid Networks for Object Detection},'' in \emph{IEEE Conference
  on Computer Vision and Pattern Recognition (CVPR)}, 2017, pp. 2117--2125.

\bibitem{ImageNet}
O.~Russakovsky, J.~Deng, H.~Su, J.~Krause, S.~Satheesh, S.~Ma, Z.~Huang,
  A.~Karpathy, A.~Khosla, M.~Bernstein, A.~C. Berg, and L.~Fei-Fei, ``{ImageNet
  Large Scale Visual Recognition Challenge},'' \emph{International Journal of
  Computer Vision (IJCV)}, pp. 211--252, 2015.

\bibitem{OpenImages}
A.~Kuznetsova, H.~Rom, N.~Alldrin, J.~Uijlings, I.~Krasin, J.~Pont-Tuset,
  S.~Kamali, S.~Popov, M.~Malloci, A.~Kolesnikov \emph{et~al.}, ``The open
  images dataset v4,'' \emph{International Journal of Computer Vision (IJCV)},
  vol. 128, no.~7, pp. 1956--1981, 2020.

\end{thebibliography}

\newpage
\begin{appendix}
\section{Appendix}
\label{appendix}

\subsection{Generic Object Detection}
\label{sec:appendix-generic-object-detection}
Generic object detectors can be categorized into one-stage and two-stage detectors.
Note that, one-stage detectors are also sometimes referred to as single-shot detectors.
However, in this context, \emph{shot} still refers to just one processing stage and not to the number of training examples.

One-stage detectors such as SSD~\cite{SSD-eccv2016}, YOLO \cite{YOLOv2-cvpr2017, YOLOv3-arxiv2018} or RetinaNet~\cite{RetinaNet-iccv2017} use a backbone network for feature extraction and do classification as well as bounding box regression directly on the extracted feature maps.
In contrast, two-stage detectors such as 
Faster R-CNN~\cite{FasterRCNN-neurips2015} first find regions of interest (RoIs) that can contain any object.
Afterwards, these RoIs are classified and the bounding box parameters are adapted.
A lot of approaches for few-shot object detection build upon Faster R-CNN~\cite{FasterRCNN-neurips2015}, most often with a ResNet~\cite{ResNet-cvpr2016} as backbone.
For higher detection rates, the backbone is often extended with a feature pyramid network (FPN)~\cite{FPN-cvpr2017}.

As a lot of approaches for few-shot object detection build upon Faster R-CNN, its architecture is depicted in Figure~\ref{fig:FasterRCNN} and further described in the following:
\begin{figure}[h]
    \centering
    \includegraphics[width=\linewidth]{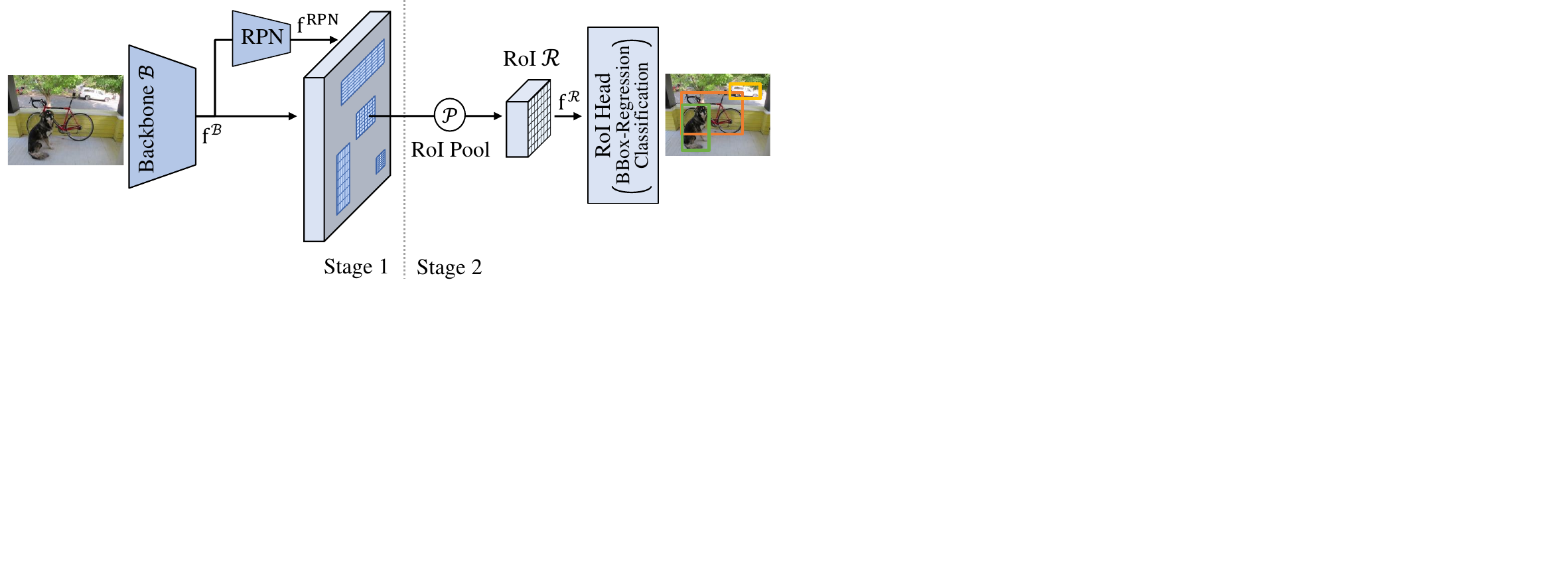}
    \caption{Architecture of Faster R-CNN~\cite{FasterRCNN-neurips2015}}
    \label{fig:FasterRCNN}
\end{figure}
First, features are extracted in a backbone network $\Backbone$ such as a ResNet~\cite{ResNet-cvpr2016}.
Afterwards, a region proposal network (RPN) predicts bounding box parameters for region proposals.
Therefore, anchor boxes of different aspect ratios are used which are placed all over the feature maps and encode all possible object locations.
The RPN's binary classifier predicts whether an anchor box is a foreground object, or whether it will be discarded as background.
Moreover, the RPN also regresses anchor deltas to better adjust the anchor boxes to the objects.
The foreground objects are then considered as region proposals and can contain arbitrary objects.
In the second stage, RoI Pooling uses the predicted bounding box parameters for cropping regions of the backbone's output feature maps.
These pooled regions are referred to as regions of interests (RoIs).
RoI Pooling as well as its advanced method RoI Align~\cite{MaskRCNN} are configured such that all RoIs have the same spatial resolution (typically $7{\times}7$).
Finally, the box head classifies the RoIs into the object categories (+ one additional background category) and the bounding box parameters are further refined. 

\begin{figure}[ht]
    \centering
    \includegraphics[scale=0.5]{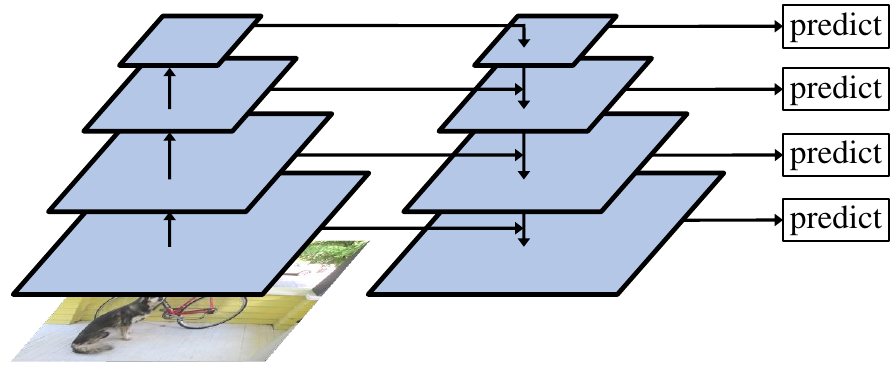}
    \caption{Feature pyramid network (FPN) \cite{FPN-cvpr2017}}
    \label{fig:FPN}
\end{figure}

For higher detection rates, the backbone is often extended with a feature pyramid network (FPN)~\cite{FPN-cvpr2017} as illustrated in~\autoref{fig:FPN}.
The FPN outputs feature maps of different resolutions which helps the detector in detecting objects of different sizes.
When a FPN is used, the RPN is applied once for each feature scale.
Depending on the size of the region proposal a corresponding feature scale is used for RoI Pooling.

\subsection{Mathematical notation}

The following listing of mathematical identifiers is intended to serve as a reference.

\vspace*{1.5mm}%
\begin{itemize}[itemsep=0.5ex, leftmargin=8ex]
    \item[$\Cbase$] base categories
    \item[$\Cnovel$] novel categories
    \item[$\Dbase$] base dataset
    \item[$\Dnovel$] novel dataset
    \item[$\featBackbone$] backbone features
    \item[$\featRPN$] features after RPN
    \item[$\featRoI$] features of the RoI
    \item[$\RoI$] RoI
    \item[$\ImQ$] query image
    \item[$\ImS$] support image
    \item[$\ImSc$] support image of category c
    \item[$\BranchQ$] query branch
    \item[$\BranchS$] support branch
    \item[$\featQ$] query features
    \item[$\featS$] support features
    \item[$\featSc$] support features of category c
    \item[$\Backbone$] backbone
    \item[$\featBackboneQ$] features after query backbone
    \item[$\featBackboneS$] features after support backbone
    \item[$\RoIc$] RoI of category c
    \item[$\Aggregator$] aggregation
    \item[$\RoIAgg$] RoI after aggregation
    \item[$\RoIAggc$] RoI after aggregation of category c
    \item[$K$] number of shots
    \item[$N$] number of novel categories
    \item[$\ModelInit$] initial model (pretrained on imagenet)
    \item[$\ModelBase$] model trained on $\Dbase$
    \item[$\ModelFinal$] model trained on $\Dnovel$ and maybe $\Dbase$
\end{itemize}

\subsection{Scope of this survey}

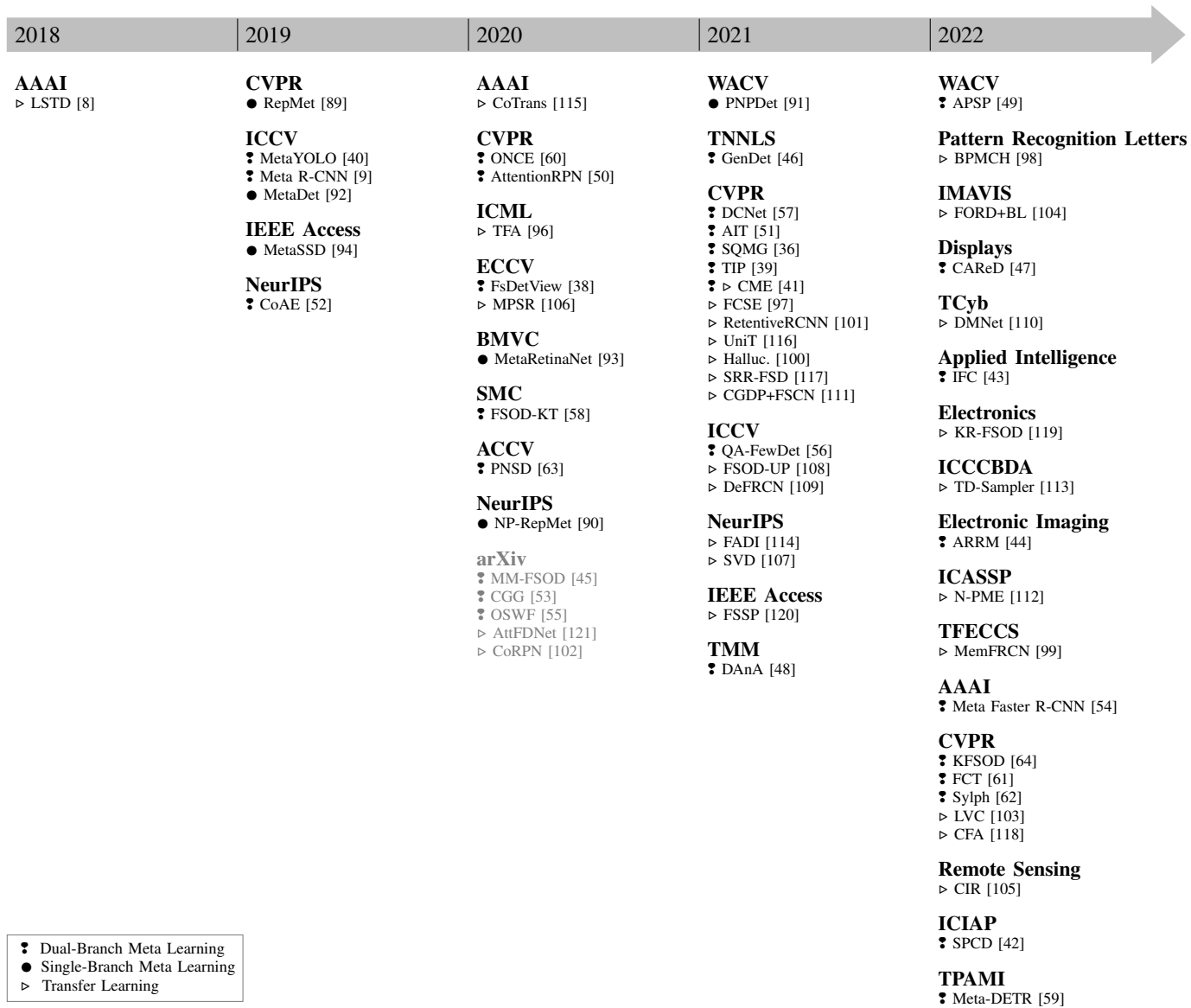
\begin{figure*}[!ht]
    \centering
    \begin{tikzpicture}
\newcommand{\combineXY}[2]{(#1 |- 0, 0 |- #2)}
\newcommand{\drawVertical}[2]{\draw[line width=1, color=lightgray] (#1) -- \combineXY{#1}{#2}}
\def\arrowwidth{15};
\def\ywidth{2.6};
\def\mwidth{\ywidth/12}; %
\def\distance{0.2}; %
\def\height{0.6};
\newcommand{\capt}[1]{\\\small{\textbf{#1}}}

\draw[-{Triangle[width=25,length=\arrowwidth]}, line width=\arrowwidth, color=lightgray](0,0) -- (\linewidth, 0);
\node[right] (2018) at (0,0) {2018};
\node[right=\ywidth of 2018, anchor=west] (2019) {2019};
\draw (2019.south west) -- (2019.north west);
\node[right=\ywidth of 2019, anchor=west] (2020) {2020};
\draw (2020.south west) -- (2020.north west);
\node[right=\ywidth of 2020, anchor=west] (2021) {2021};
\draw (2021.south west) -- (2021.north west);
\node[right=\ywidth of 2021, anchor=west] (2022) {2022};
\draw (2022.south west) -- (2022.north west);
\node[below left=\distance and 0 of 2018, anchor=north west, align=left, font=\scriptsize] (2018conf){
    \capt{AAAI} \\ %
    \transfer~LSTD~\cite{LSTD-aaai2018} \\
};
\node[below left=\distance and 0 of 2019, anchor=north west, align=left, font=\scriptsize] (2019conf) {
    \capt{CVPR} \\
    \singlebranch~RepMet~\cite{RepMet-cvpr2019} \\
    \capt{ICCV} \\
    \dualbranch~MetaYOLO~\cite{MetaYOLO-iccv2019} \\
    \dualbranch~Meta R-CNN~\cite{MetaRCNN-iccv2019} \\
    \singlebranch~MetaDet~\cite{MetaDet-iccv2019} \\
    \capt{IEEE Access} \\
    \singlebranch~MetaSSD~\cite{MetaSSD-ieee2019} \\
    \capt{NeurIPS} \\
    \dualbranch~CoAE~\cite{CoAE-neurips2019} \\
    
};
\node[below left=\distance and 0 of 2020, anchor=north west, align=left, font=\scriptsize] (2020conf) {
    \capt{AAAI} \\  %
    \transfer~CoTrans~\cite{ContextTransformer-aaai2020} \\
    \capt{CVPR} \\ %
    \dualbranch~ONCE~\cite{ONCE-cvpr2020} \\
    \dualbranch~AttentionRPN~\cite{AttentionRPN-cvpr2020} \\
    \capt{ICML} \\ %
    \transfer~TFA~\cite{TFA-icml2020} \\
    \capt{ECCV} \\ %
    \dualbranch~FsDetView~\cite{FSDetView-eecv2020} \\
    \transfer~MPSR~\cite{MPSR-eecv2020} \\
    \capt{BMVC} \\ %
    \singlebranch~MetaRetinaNet~\cite{MetaRetinaNet-bmvc2020} \\
    \capt{SMC} \\ %
    \dualbranch~FSOD-KT~\cite{FewShotKnowledgeTransfer-smc2020} \\
    \capt{ACCV} \\ %
    \dualbranch~PNSD~\cite{PNSD-accv2020} \\
    \capt{NeurIPS} \\ %
    \singlebranch~NP-RepMet~\cite{NP-RepMet-neurips2020} \\
    \capt{\graytext{arXiv}} \\
    \graytext{\dualbranch~MM-FSOD~\cite{MM-FSOD-arxiv2020}} \\

    \graytext{\dualbranch~CGG~\cite{ClosingGeneralizationGap-arxiv2020}} \\
    \graytext{\dualbranch~OSWF~\cite{OneShotWithoutFinetuning-arxiv2020}} \\
    \graytext{\transfer~AttFDNet~\cite{BottomUpTopDownAttention-arxiv2020}} \\
    \graytext{\transfer~CoRPN~\cite{CooperatingRPNs-arxiv2020}} \\
};
\node[below left=\distance and 0 of 2021, anchor=north west, align=left, font=\scriptsize] (2021conf) {
    \capt{WACV} \\ %
    \singlebranch~PNPDet~\cite{PNPDet-wacv2021} \\
    \capt{TNNLS} \\
    \dualbranch~GenDet~\cite{GenDet-tnnls2021} \\
    \capt{CVPR} \\ %
    \dualbranch~DCNet~\cite{DCNet-cvpr2021} \\
    \dualbranch~AIT~\cite{AIT-cvpr2021} \\
    \dualbranch~SQMG~\cite{SQMG-cpvr2021} \\
    \dualbranch~TIP~\cite{TIP-cvpr2021} \\
    \dualbranch~\transfer~CME~\cite{CME-cvpr2021} \\
    \transfer~FCSE~\cite{FSCE_FSOD-cvpr2021} \\
    \transfer~RetentiveRCNN~\cite{RetentiveRCNN-cvpr2021} \\
    \transfer~UniT~\cite{UniT-cvpr2021} \\
    \transfer~Halluc.~\cite{Halluc.-cvpr2021} \\
    \transfer~SRR-FSD~\cite{SRR-FSD-cvpr2021} \\
    \transfer~CGDP+FSCN~\cite{CGDP+FSCN-cvpr2021} \\
    \capt{ICCV} \\
    \dualbranch~QA-FewDet~\cite{QAFewDet-iccv2021} \\
    \transfer~FSOD-UP~\cite{FSOD-UP-iccv2021} \\
    \transfer~DeFRCN~\cite{DeFRCN-iccv2021} \\
    \capt{NeurIPS} \\
    \transfer~FADI~\cite{FADI-neurips2021} \\
    \transfer~SVD~\cite{SVD-Dictionary-neurips2021} \\
    \capt{IEEE Access} \\
    \transfer~FSSP~\cite{FSSP-ieee2021} \\
    \capt{TMM} \\
    \dualbranch~DAnA~\cite{DAnA-tmm2021} \\
};
\node[below left=\distance and 0 of 2022, anchor=north west, align=left, font=\scriptsize] (2022conf) {
    \capt{WACV} \\ %
    \dualbranch~APSP~\cite{APSP-wacv2022} \\
    \capt{Pattern Recognition Letters} \\ %
    \transfer~BPMCH~\cite{BPMCH-prl2022} \\
    \capt{IMAVIS} \\ %
    \transfer~FORD+BL~\cite{FORD+BL-imavis2022} \\
    \capt{Displays} \\ %
    \dualbranch~CAReD~\cite{CAReD-displays2022} \\
    \capt{TCyb} \\ %
    \transfer~DMNet~\cite{DMNet-tcyb2022} \\
    \capt{Applied Intelligence} \\ %
    \dualbranch~IFC~\cite{IFC-apin2022} \\
    \capt{Electronics} \\ %
    \transfer~KR-FSOD~\cite{KR-FSOD-electronics2022} \\
    \capt{ICCCBDA} \\ %
    \transfer~TD-Sampler~\cite{TD-Sampler-icccbda2022} \\
    \capt{Electronic Imaging} \\ %
    \dualbranch~ARRM~\cite{ARRM-electronimaging2022} \\    
    \capt{ICASSP} \\ %
    \transfer~N-PME~\cite{N-PME-icassp2022} \\    
    \capt{TFECCS} \\
    \transfer~MemFRCN~\cite{MemFRCN-tfeccs2022} \\
    \capt{AAAI} \\
    \dualbranch~Meta Faster R-CNN~\cite{MetaFasterRCNN-aaai2022} \\
    \capt{CVPR} \\ %
    \dualbranch~KFSOD~\cite{KFSOD-cvpr2022} \\
    \dualbranch~FCT~\cite{FCT-cvpr2022} \\
    \dualbranch~Sylph~\cite{Sylph-cvpr2022} \\
    \transfer~LVC~\cite{LVC-cvpr2022} \\
    \transfer~CFA~\cite{CFA-cvpr2022} \\
    \capt{Remote Sensing} \\ %
    \transfer~CIR~\cite{CIR-remote-sensing-2022} \\    
    \capt{ICIAP} \\ %
    \dualbranch~SPCD~\cite{SPCD-iciap2022} \\
    \capt{TPAMI} \\ %
    \dualbranch~Meta-DETR~\cite{Meta-DETR-tpami2022} \\
};
\node[below left=13.6cm and 0 of 2018, anchor=north west, align=left, font=\scriptsize, draw=gray] (legend){
    ~\dualbranch~~Dual-Branch Meta Learning\\
    ~\singlebranch\,~Single-Branch Meta Learning\\
    ~\transfer~~Transfer Learning
};
\end{tikzpicture}
    \vspace*{-8mm}%
    \caption{Scope of this survey. We categorize approaches into dual- and single-branch meta learning and transfer learning.}
    \label{fig:timeline}
    \vspace*{5mm}
\end{figure*}

In~\autoref{fig:timeline} we list the approaches, that we cover in this survey.
First of all, we can see that few-shot object detection is a rather young but emerging research field as most approaches have been published only within the last three years.
Most approaches use transfer learning or dual-branch meta learning.

\newpage
\newpage
\subsection{Further benchmark results}
\label{sec:appendix-further-results}

\subsubsectionspace{FSOD dataset}
In contrast to VOC and COCO, the FSOD dataset~\cite{AttentionRPN-cvpr2020} is specifically designed for few-shot object detection, but rarely used so far.
It is comprised of images from other datasets such as ImageNet~\cite{ImageNet} and OpenImages~\cite{OpenImages}.
In total, there are 1000 categories of which 800 are used as base categories, and the remaining 200 categories as novel categories.
Fan~\etal~\cite{AttentionRPN-cvpr2020} constructed the dataset such that the 200 novel categories are the most distinct to the remaining base categories.
Moreover, the novel categories do not contain any COCO categories, which allows pretraining with COCO.
Despite the many categories, the dataset remains compact, as the most frequent category still has no more than 208 images.
With around 66K images, the FSOD dataset contains only about half as many images as the COCO dataset and therefore allows for faster training.

Fan~\etal~\cite{AttentionRPN-cvpr2020} (AttentionRPN) propose a 2-way-5-shot evaluation protocol, where models are only pretrained on $\Dbase$ and not finetuned on $\Dnovel$.
Counterintuitively, approaches do not need to differentiate between the 200 novel categories.
Only four approaches have been benchmarked on the FSOD dataset so far.
The benchmarking results are shown in \autoref{tab:fsod-results}.

\begin{table}[b!]
    \centering
    \scriptsize
    \setlength\tabcolsep{3pt} %
    \resizebox{\textwidth}{!}{%
\begin{tabular}{@{}clrl|cc@{}}
\toprule
\multicolumn{1}{l}{}                                                           & \textbf{Approach}                                                      & \textbf{Publication}               & \textbf{Detector}                         & \textbf{AP50}                & \textbf{AP75}                \\ \midrule
\multicolumn{1}{c|}{}                                                          & \dualbranch~AttentionRPN   \cite{AttentionRPN-cvpr2020}                & CVPR 2020                          & Faster R-CNN R-50                         & 27.5                         & 19.4                         \\
\multicolumn{1}{c|}{}                                                          & \cellcolor[HTML]{EFEFEF}\dualbranch~PNSD~\cite{PNSD-accv2020}          & \cellcolor[HTML]{EFEFEF}ACCV 2020  & \cellcolor[HTML]{EFEFEF}Faster R-CNN R-50 & \cellcolor[HTML]{EFEFEF}29.8 & \cellcolor[HTML]{EFEFEF}22.6 \\
\multicolumn{1}{c|}{}                                                          & \dualbranch~KFSOD   \cite{KFSOD-cvpr2022}                              & CVPR 2022                          & Faster R-CNN R-50                         & 33.4                         & 29.6                         \\
\multicolumn{1}{c|}{\multirow{-4}{*}{\rotatebox[origin=c]{90}{no   finetun.}}} & \cellcolor[HTML]{EFEFEF}\dualbranch~MM-FSOD   \cite{MM-FSOD-arxiv2020} & \cellcolor[HTML]{EFEFEF}arXiv 2020 & \cellcolor[HTML]{EFEFEF}Faster R-CNN R-34 & \cellcolor[HTML]{EFEFEF}51.7 & \cellcolor[HTML]{EFEFEF}31.1 \\ \bottomrule
\end{tabular}%
}
    \caption{Results for 2-way-5-shot evaluation on the FSOD~\cite{AttentionRPN-cvpr2020} dataset.}
    \label{tab:fsod-results}
\end{table}

\begin{table*}[!b]
    \centering
    \scriptsize
    \setlength\tabcolsep{3pt} %
    \resizebox{0.7\textwidth}{!}{%
    \begin{tabular}{@{}clrlcccccc@{}}
\toprule
\multicolumn{1}{l}{}                                                              & \textbf{}                                                                             & \textbf{}                                               & \multicolumn{1}{l|}{\textbf{}}                                                        & \multicolumn{3}{c|}{\textbf{Novel Categories}}                                                                                                                                       & \multicolumn{3}{c}{\textbf{Base   Categories}}                                                                                                                  \\ \midrule
\multicolumn{1}{l}{\textbf{}}                                                     & \textbf{Approach}                                                                     & \multicolumn{1}{c}{\textbf{Publication}}                & \multicolumn{1}{l|}{\textbf{Detector}}                                                & \textbf{AP}                                         & \textbf{AP50}                                       & \multicolumn{1}{c|}{\textbf{AP75}}                                       & \textbf{AP}                                         & \textbf{AP50}                                       & \textbf{AP75}                                       \\ \midrule
\multicolumn{1}{l}{}                                                              & \multicolumn{9}{l}{\textbf{Results over a single   run:}}                                                                                                                                                                                                                                                                                                                                                                                                                                                                                                                                        \\ \midrule
\multicolumn{1}{c|}{}                                                             & \dualbranch~Meta Faster-RCNN   \cite{MetaFasterRCNN-aaai2022}                         & AAAI 2022                                               & \multicolumn{1}{l|}{Faster R-CNN R-101}                                               & 5.0                                                 & 10.2                                                & \multicolumn{1}{c|}{4.6}                                                 & ---                                                 & ---                                                 & ---                                                 \\
\multicolumn{1}{c|}{\multirow{-2}{*}{\rotatebox[origin=c]{90}{no ft.}}}           & \cellcolor[HTML]{EFEFEF}{\color[HTML]{999999} \dualbranch~DAnA * \cite{DAnA-tmm2021}} & \cellcolor[HTML]{EFEFEF}{\color[HTML]{999999} TMM 2021} & \multicolumn{1}{l|}{\cellcolor[HTML]{EFEFEF}{\color[HTML]{999999} Faster R-CNN R-50}} & \cellcolor[HTML]{EFEFEF}{\color[HTML]{999999} 11.9} & \cellcolor[HTML]{EFEFEF}{\color[HTML]{999999} 25.6} & \multicolumn{1}{c|}{\cellcolor[HTML]{EFEFEF}{\color[HTML]{999999} 10.4}} & \cellcolor[HTML]{EFEFEF}{\color[HTML]{999999} 27.8} & \cellcolor[HTML]{EFEFEF}{\color[HTML]{999999} 46.3} & \cellcolor[HTML]{EFEFEF}{\color[HTML]{999999} 27.7} \\ \midrule
\multicolumn{1}{c|}{}                                                             & \transfer~BPMCH   \cite{BPMCH-prl2022}                                                & PRL 2022                                                & \multicolumn{1}{l|}{FCOS R-50}                                                        & 2.4                                                 & ---                                                 & \multicolumn{1}{c|}{---}                                                 & 29.4                                                & ---                                                 & ---                                                 \\
\multicolumn{1}{c|}{}                                                             & \cellcolor[HTML]{EFEFEF}\transfer~FORD   \cite{FORD+BL-imavis2022}                    & \cellcolor[HTML]{EFEFEF}IMAVIS 2022                     & \multicolumn{1}{l|}{\cellcolor[HTML]{EFEFEF}Faster R-CNN R-101}                       & \cellcolor[HTML]{EFEFEF}3.6                         & \cellcolor[HTML]{EFEFEF}7.1                         & \multicolumn{1}{c|}{\cellcolor[HTML]{EFEFEF}3.5}                         & \cellcolor[HTML]{EFEFEF}---                         & \cellcolor[HTML]{EFEFEF}---                         & \cellcolor[HTML]{EFEFEF}---                         \\
\multicolumn{1}{c|}{}                                                             & \transfer~CoRPN   w/ cos \cite{CooperatingRPNs-arxiv2020}                             & arXiv 2020                                              & \multicolumn{1}{l|}{Faster R-CNN R-50}                                                & 3.7                                                 & 6.8                                                 & \multicolumn{1}{c|}{3.8}                                                 & 32.1                                                & 52.9                                                & 34.4                                                \\
\multicolumn{1}{c|}{}                                                             & \cellcolor[HTML]{EFEFEF}\transfer~Halluc.   (TFA w/cos) \cite{Halluc.-cvpr2021}       & \cellcolor[HTML]{EFEFEF}CVPR 2021                       & \multicolumn{1}{l|}{\cellcolor[HTML]{EFEFEF}Faster R-CNN R-101}                       & \cellcolor[HTML]{EFEFEF}3.8                         & \cellcolor[HTML]{EFEFEF}6.5                         & \multicolumn{1}{c|}{\cellcolor[HTML]{EFEFEF}4.3}                         & \cellcolor[HTML]{EFEFEF}31.5                        & \cellcolor[HTML]{EFEFEF}50.8                        & \cellcolor[HTML]{EFEFEF}33.9                        \\
\multicolumn{1}{c|}{}                                                             & \transfer~CoRPN   w/ cos \cite{CooperatingRPNs-arxiv2020}                             & arXiv 2020                                              & \multicolumn{1}{l|}{Faster R-CNN R-101}                                               & 4.1                                                 & 7.2                                                 & \multicolumn{1}{c|}{4.4}                                                 & 34.1                                                & 55.1                                                & 36.5                                                \\
\multicolumn{1}{c|}{}                                                             & \cellcolor[HTML]{EFEFEF}\transfer~Halluc.   (CoRPN w/cos) \cite{Halluc.-cvpr2021}     & \cellcolor[HTML]{EFEFEF}CVPR 2021                       & \multicolumn{1}{l|}{\cellcolor[HTML]{EFEFEF}Faster R-CNN R-101}                       & \cellcolor[HTML]{EFEFEF}4.4                         & \cellcolor[HTML]{EFEFEF}7.5                         & \multicolumn{1}{c|}{\cellcolor[HTML]{EFEFEF}4.9}                         & \cellcolor[HTML]{EFEFEF}32.3                        & \cellcolor[HTML]{EFEFEF}52.4                        & \cellcolor[HTML]{EFEFEF}34.4                        \\
\multicolumn{1}{c|}{}                                                             & \dualbranch~Meta   Faster-RCNN \cite{MetaFasterRCNN-aaai2022}                         & AAAI 2022                                               & \multicolumn{1}{l|}{Faster R-CNN R-101}                                               & 5.1                                                 & 10.7                                                & \multicolumn{1}{c|}{4.3}                                                 & ---                                                 & ---                                                 & ---                                                 \\
\multicolumn{1}{c|}{}                                                             & \cellcolor[HTML]{EFEFEF}\dualbranch~FCT   \cite{FCT-cvpr2022} $\longleftarrow$        & \cellcolor[HTML]{EFEFEF}CVPR 2022                       & \multicolumn{1}{l|}{\cellcolor[HTML]{EFEFEF}Faster R-CNN PVTv2-B2-Li}                 & \cellcolor[HTML]{EFEFEF}5.6                         & \cellcolor[HTML]{EFEFEF}---                         & \multicolumn{1}{c|}{\cellcolor[HTML]{EFEFEF}---}                         & \cellcolor[HTML]{EFEFEF}---                         & \cellcolor[HTML]{EFEFEF}---                         & \cellcolor[HTML]{EFEFEF}---                         \\
\multicolumn{1}{c|}{}                                                             & \transfer~FADI~\cite{FADI-neurips2021}                                                & NeurIPS 2021                                            & \multicolumn{1}{l|}{Faster R-CNN R-101}                                               & 5.7                                                 & 10.4                                                & \multicolumn{1}{c|}{6.0}                                                 & ---                                                 & ---                                                 & ---                                                 \\
\multicolumn{1}{c|}{\multirow{-10}{*}{\rotatebox[origin=c]{90}{with finetuning}}} & \cellcolor[HTML]{EFEFEF}\transfer~DeFRCN   \cite{DeFRCN-iccv2021} $\longleftarrow$    & \cellcolor[HTML]{EFEFEF}ICCV 2021                       & \multicolumn{1}{l|}{\cellcolor[HTML]{EFEFEF}Faster R-CNN R-101}                       & \cellcolor[HTML]{EFEFEF}9.3                         & \cellcolor[HTML]{EFEFEF}---                         & \multicolumn{1}{c|}{\cellcolor[HTML]{EFEFEF}---}                         & \cellcolor[HTML]{EFEFEF}---                         & \cellcolor[HTML]{EFEFEF}---                         & \cellcolor[HTML]{EFEFEF}---                         \\ \midrule
\multicolumn{1}{l}{}                                                              & \multicolumn{9}{l}{\textbf{Results averaged over   multiple random runs:}}                                                                                                                                                                                                                                                                                                                                                                                                                                                                                                                       \\ \midrule
\multicolumn{1}{l|}{\rotatebox[origin=c]{90}{\parbox{0.25cm}{no   ft.}}}          & \dualbranch~QA-FewDet   \cite{QAFewDet-iccv2021}                                      & ICCV 2021                                               & \multicolumn{1}{l|}{Faster R-CNN R-101}                                               & 5.1                                                 & 10.5                                                & \multicolumn{1}{c|}{4.5}                                                 & ---                                                 & ---                                                 & ---                                                 \\ \midrule
\multicolumn{1}{c|}{}                                                             & \cellcolor[HTML]{EFEFEF}\transfer~TFA w/cos   \cite{TFA-icml2020}                     & \cellcolor[HTML]{EFEFEF}ICML 2020                       & \multicolumn{1}{l|}{\cellcolor[HTML]{EFEFEF}Faster R-CNN R-101}                       & \cellcolor[HTML]{EFEFEF}1.9                         & \cellcolor[HTML]{EFEFEF}3.8                         & \multicolumn{1}{c|}{\cellcolor[HTML]{EFEFEF}1.7}                         & \cellcolor[HTML]{EFEFEF}31.9                        & \cellcolor[HTML]{EFEFEF}51.8                        & \cellcolor[HTML]{EFEFEF}34.3                        \\
\multicolumn{1}{c|}{}                                                             & \transfer~DeFRCN   \cite{DeFRCN-iccv2021} $\longleftarrow$                            & ICCV 2021                                               & \multicolumn{1}{l|}{Faster R-CNN R-101}                                               & 4.8                                                 & ---                                                 & \multicolumn{1}{c|}{---}                                                 & ---                                                 & ---                                                 & ---                                                 \\
\multicolumn{1}{c|}{}                                                             & \cellcolor[HTML]{EFEFEF}\dualbranch~QA-FewDet   \cite{QAFewDet-iccv2021}              & \cellcolor[HTML]{EFEFEF}ICCV 2021                       & \multicolumn{1}{l|}{\cellcolor[HTML]{EFEFEF}Faster R-CNN R-101}                       & \cellcolor[HTML]{EFEFEF}4.9                         & \cellcolor[HTML]{EFEFEF}10.3                        & \multicolumn{1}{c|}{\cellcolor[HTML]{EFEFEF}4.4}                         & \cellcolor[HTML]{EFEFEF}---                         & \cellcolor[HTML]{EFEFEF}---                         & \cellcolor[HTML]{EFEFEF}---                         \\
\multicolumn{1}{c|}{}                                                             & \dualbranch~FCT   \cite{FCT-cvpr2022} $\longleftarrow$                                & CVPR 2022                                               & \multicolumn{1}{l|}{Faster R-CNN PVTv2-B2-Li}                                         & 5.1                                                 & ---                                                 & \multicolumn{1}{c|}{---}                                                 & ---                                                 & ---                                                 & ---                                                 \\
\multicolumn{1}{c|}{}                                                             & \cellcolor[HTML]{EFEFEF}\transfer~MemFRCN (DeFRCN) \cite{MemFRCN-tfeccs2022}          & \cellcolor[HTML]{EFEFEF}TFECCS 2022                     & \multicolumn{1}{l|}{\cellcolor[HTML]{EFEFEF}Faster R-CNN R-101}                       & \cellcolor[HTML]{EFEFEF}5.2                         & \cellcolor[HTML]{EFEFEF}---                         & \multicolumn{1}{c|}{\cellcolor[HTML]{EFEFEF}---}                         & \cellcolor[HTML]{EFEFEF}---                         & \cellcolor[HTML]{EFEFEF}---                         & \cellcolor[HTML]{EFEFEF}---                         \\
\multicolumn{1}{c|}{\multirow{-6}{*}{\rotatebox[origin=c]{90}{with finetuning}}}  & \dualbranch~Meta-DETR   \cite{Meta-DETR-tpami2022}                                    & TPAMI 2022                                              & \multicolumn{1}{l|}{Deformable DETR R-101}                                            & 7.5                                                 & 12.5                                                & \multicolumn{1}{c|}{7.7}                                                 & ---                                                 & ---                                                 & ---                                                 \\ \bottomrule
\end{tabular}%

    }
    \caption{%
    Benchmark results on the 1-shot Microsoft COCO dataset sorted by novel AP.
     ---:~no result reported in paper. 
     *:~Deviating evaluation protocol preventing fair comparison as described in \autoref{sec:experiments-deviating-evaluation}.
     \dualbranch:~dual-branch meta learning. \singlebranch:~single-branch meta learning. \transfer:~transfer learning.
     }%
    \label{tab:coco-1-shot}
\end{table*}

\subsubsectionspace{PASCAL VOC with pretraining on reduced ImageNet}
In~\autoref{tab:voc-imagenet}, we show results on VOC when pretraining with all 1000 ImageNet categories vs. when reducing the number of categories to 725 in order to remove all COCO-related categories.
All these approaches use a different split for base and novel categories than the one described above, with just $N=4$ novel categories:
$\Cnovel = \{\textit{cow, sheep, cat, aeroplane}\}$.
We refer to this as novel set 4.
We can see that for CoAE~\cite{CoAE-neurips2019} the performance drop is rather high, whereas AIT~\cite{AIT-cvpr2021} is able to better handle pretraining with reduced categories.

\begin{table}[ht!]
    \centering
    \scriptsize
    \setlength\tabcolsep{3pt} %
    \begin{tabular}{@{}lrlc|ll@{}}
\toprule
\textbf{Approach} & \multicolumn{1}{c}{\textbf{Publication}} & \textbf{Detector} & \multicolumn{1}{c|}{\textbf{ImageNet}} & \multicolumn{1}{c}{\textbf{Novel}} & \multicolumn{1}{c}{\textbf{Base}} \\ \midrule
\dualbranch~CoAE~\cite{CoAE-neurips2019} & NeurIPS 2019 & Faster R-CNN R-50 & 725 & 63.8 & 55.1 \\
\rowcolor[HTML]{EFEFEF}%
\dualbranch~CoAE~\cite{CoAE-neurips2019} & NeurIPS 2019 & Faster R-CNN R-50 & 1000 & 68.2 & 60.1 \\
\dualbranch~AIT~\cite{AIT-cvpr2021} & CVPR 2021 & Faster R-CNN R-50 & 725 & 72.2 & 67.2 \\
\rowcolor[HTML]{EFEFEF}%
\dualbranch~AIT~\cite{AIT-cvpr2021} & CVPR 2021 & Faster R-CNN R-50 & 1000 & 73.1 & 69.2 \\ \bottomrule
\end{tabular}
    \caption{Effect of removing COCO-related categories from ImageNet to prevent foreseeing novel categories. $AP_{50}$ results on PASCAL VOC novel set 4 are reported. All approaches are one-shot approaches and do not finetune on $\Dnovel$.
    }
    \label{tab:voc-imagenet}
\end{table}

\subsubsectionspace{Microsoft COCO}
The benchmark results on Microsoft COCO for $K=1$ are shown in \autoref{tab:coco-1-shot} and for $K=30$ in \autoref{tab:coco-30-shot}.

\subsubsectionspace{Other datasets}
Some early approaches for few-shot object detection \cite{LSTD-aaai2018}, \cite{RepMet-cvpr2019}, \cite{NP-RepMet-neurips2020} use the ImageNet-LOC detection dataset.
Karlinsky~\etal~\cite{RepMet-cvpr2019} (RepMet) design a benchmark by training on the first 100 categories (mostly animals) and evaluating on 214 unseen animal categories in a 5-way manner.

Despite being rarely used so far, the long-tail LVIS dataset~\cite{LVIS-cvpr2019} could also be utilized for few-shot object detection:
For the LVIS dataset the COCO images were re-annotated with 1203 categories.
The categories can be split in 405 frequent ($>$100 training examples), 461 common (\mbox{11--100} training examples) and 337 rare (\mbox{1--10} training examples) categories, containing 71 categories with just one training example.
For few-shot object detection, the rare categories with at most 10 training examples can be utilized as novel categories $\Cnovel$ and the others as base categories $\Cbase$.
However, it should be noted, that the rare categories are also rare in the validation set and, thus, prone to high variance in evaluation.
So far, the LVIS dataset with this split has only been used by TFA~\cite{TFA-icml2020} and LST~\cite{LearningSegmentTail-cvpr2020}.

\subsection{Relation to other surveys}
\label{sec:appendix-other-surveys}
Although few-shot object detection is a young research field, it is growing rapidly.
We therefore carefully searched for related surveys and at this time (September 2022) came up with the following surveys dealing with a similar topic: \cite{FSOD-Survey-acm2022, FSOD-self-supervised-survey-arxiv2021, FSOD-comparative-review-arxiv2021, Low-Shot-Detection-survey-arxiv2021, FSOD-empirical-study-arxiv2022, FSOD-survey-chinese}.
At the time of the initial submission of this paper (December 2021), only \cite{FSOD-self-supervised-survey-arxiv2021, FSOD-comparative-review-arxiv2021, Low-Shot-Detection-survey-arxiv2021} were available as preprints (see \autoref{tab:surveys-comparison} for detailed dates). 
However, none of them were peer-reviewed and only \cite{FSOD-Survey-acm2022, FSOD-survey-chinese} are officially published up to now (September 2022).
However, we still want to differentiate our work from all others.

\mydate
\begin{table*}[!t]
\footnotesize
\begin{tabular}{@{}clrrcll@{}}
\toprule
\textbf{Paper}                               & \textbf{Venue}                                          & \multicolumn{1}{l}{\textbf{First preprint}} & \multicolumn{1}{l}{\textbf{Latest preprint}} & \multicolumn{1}{l}{\textbf{Publication}} & \textbf{Language} & \textbf{\# FSOD Papers} \\ \midrule
Huang \etal~\cite{FSOD-self-supervised-survey-arxiv2021} & TPAMI preprint                                          & 27 Oct 2021                                & 23 Aug 2022                                   & ---                                      & English                & 19                     \\
Jiaxu \etal~\cite{FSOD-comparative-review-arxiv2021}     & arxiv preprint                                          & 30 Oct 2021                                & \multicolumn{1}{c}{---}                       & ---                                      & English                & 34                     \\
Huang \etal~\cite{Low-Shot-Detection-survey-arxiv2021}   & arxiv preprint                                          & 06 Dec 2021                                & 22 Jan 2022                                   & ---                                      & English                & 32                     \\
\rowcolor[HTML]{EFEFEF} 
\textbf{ours}                                         & arxiv preprint                                          & 22 Dec 2021                                & \today                                        & ---                                      & English                & 62                     \\
Antonelli \etal~\cite{FSOD-Survey-acm2022}                   & ACM CSUR                                                & \multicolumn{1}{c}{---}                    & \multicolumn{1}{c}{---}                       & \multicolumn{1}{r}{24 Feb 2022}         & English                & 15                     \\
Liu \etal~\cite{FSOD-empirical-study-arxiv2022}        & arxiv preprint                                          & 27 Mar 2022                                & \multicolumn{1}{c}{---}                       & ---                                      & English                & 32                     \\
Liu \etal~\cite{FSOD-survey-chinese}                   & Journal FCST & \multicolumn{1}{c}{---}                  & \multicolumn{1}{c}{---}                    & \multicolumn{1}{r}{Aug 2022}                  & Chinese           & 46                     \\ \bottomrule
\end{tabular}%
\caption{Comparison between different surveys related to FSOD.}
\label{tab:surveys-comparison}
\end{table*}

In the following, we like to highlight several aspects in which our survey can be distinguished from the surveys listed above:

\begin{itemize}
    \item \textbf{Focus}:
\cite{FSOD-self-supervised-survey-arxiv2021, FSOD-comparative-review-arxiv2021, Low-Shot-Detection-survey-arxiv2021} are broader surveys, also covering self-supervised, weakly-supervised and/or zero-shot learning.
In contrast, we focus on FSOD in order to give more detailed insights in current techniques and trends for FSOD.

    \item \textbf{Paper Structure}:
\cite{FSOD-Survey-acm2022, FSOD-self-supervised-survey-arxiv2021} sort the approaches into different categories and then describe the approaches in each category in a linear way.
While this works for a limited number of approaches, it becomes less useful when the number of papers grows.
Therefore, we first describe the general realization for different categories and then further elaborate on multiple techniques where the approaches differ from the general realization and how they try to improve.
We structure the sections accordingly.
This enables us to derive trends regarding several important aspects more easily.

    \item \textbf{Language}:
Most papers are written in English.
However, as~\cite{FSOD-survey-chinese} is only available in Chinese, it is not not accessible for the whole research community.

    \item \textbf{Comprehensiveness}:
\cite{FSOD-Survey-acm2022, FSOD-comparative-review-arxiv2021} only cover earlier work on FSOD and hence are somewhat outdated since at least some of the currently best performing approaches on common benchmarks are missing.
Furthermore, we provide a much more comprehensive survey by covering a lot more FSOD papers compared to all other related surveys \cite{FSOD-Survey-acm2022, FSOD-self-supervised-survey-arxiv2021, FSOD-comparative-review-arxiv2021, Low-Shot-Detection-survey-arxiv2021, FSOD-empirical-study-arxiv2022, FSOD-survey-chinese}, even including approaches from the current year. For a listing of the number of covered FSOD papers in each survey, we refer to \autoref{tab:surveys-comparison}.

    \item \textbf{Benchmark results}:
\cite{FSOD-comparative-review-arxiv2021} only elaborates on different approaches, but does not provide tables with benchmark results, which makes it difficult to see which approaches work best on different datasets.
\cite{FSOD-empirical-study-arxiv2022} present results, but group them by taxonomic aspects and do not sort them by performance.
This makes it extremely hard to compare approaches, especially from different categories.
In contrast, we provide a better guidance on benchmark results by highlighting differences in evaluation protocols and grouping approaches with comparable evaluations.
We provide the taxonomy as small symbols next to the name of the approaches, which enables a fast visual assignment, while also allowing a fast inter-category comparison.
\end{itemize}

Overall, our survey is most related to~\cite{FSOD-empirical-study-arxiv2022}, as it also elaborates on several core concepts and groups approaches according to this concepts.
However, with the visual taxonomy in \autoref{fig:dual-branch-meta-categories}, \autoref{fig:single-branch-meta-categorization}, \autoref{fig:categroies-transfer-learning}, we enable the reader to faster grasp which approaches follow similar concepts and what concepts seem to complement each other well.
We also provide a better guidance on benchmark results as described above.
Furthermore, we provide a much more comprehensive survey by covering nearly twice as many FSOD papers as \cite{FSOD-empirical-study-arxiv2022} did.

\begin{table*}[t!]
    \centering
    \scriptsize
    \setlength\tabcolsep{3pt} %
    \resizebox{\textwidth}{!}{%
\begin{tabular}{@{}clrlcccccccccccccc@{}}
\toprule
\multicolumn{1}{l}{\textbf{}}                                                                & \textbf{}                                                                                      &                                                          & \multicolumn{1}{l|}{}                                                                 & \multicolumn{12}{c|}{\textbf{Novel Categories}}                                                                                                                                                                                                                                                                                                                                                                                                                                                                                                                                                                                                                                                                & \multicolumn{2}{c}{\textbf{Base   Cat.}}                                                                \\
\bottomrule
\end{tabular}%
}
    \caption{
    Benchmark results for 30-shot on the Microsoft COCO dataset sorted by novel $AP_{50:95}$. 
    ---: no result reported in paper. 
    *:~Deviating evaluation protocol preventing fair comparison as described in \autoref{sec:experiments-deviating-evaluation}.
    \dualbranch: dual-branch meta learning. \singlebranch: single-branch meta learning. \transfer: transfer learning.
    }%
    \label{tab:coco-30-shot}
\end{table*}
\end{appendix}

\end{document}